\documentclass[twoside]{article}

\usepackage[accepted]{aistats2023}

\usepackage{graphicx,subcaption}
\usepackage{epsfig}
\usepackage{microtype}
\usepackage{graphicx}
\usepackage{booktabs} 
\usepackage{multirow}
\usepackage{multicol}
\usepackage{xcolor}
\usepackage{amsmath}
\usepackage{amsfonts}
\usepackage{soul}
\usepackage{algorithm}
\usepackage{algpseudocode}
\usepackage{comment}
\usepackage{appendix}
\usepackage[export]{adjustbox}
\usepackage{breakcites}
\usepackage[bottom]{footmisc}
\usepackage[normalem]{ulem}
\useunder{\uline}{\ul}{}
\usepackage{soul}
\usepackage{wrapfig, blindtext}
\usepackage{caption}
\usepackage{xspace}
\usepackage{bm}
\usepackage{url}
\usepackage{colortbl}

\usepackage{amsmath}
\usepackage{amssymb}
\usepackage{mathtools}
\usepackage{amsthm}

\usepackage[capitalize,noabbrev]{cleveref}

\theoremstyle{plain}
\newtheorem{theorem}{Theorem}[section]
\newtheorem{proposition}[theorem]{Proposition}

\newtheorem{corollary}[theorem]{Corollary}
\theoremstyle{definition}
\newtheorem{definition}[theorem]{Definition}

\theoremstyle{remark}
\newtheorem{remark}[theorem]{Remark}
\usepackage[textsize=tiny]{todonotes}

\usepackage[round]{natbib}

\begin{document}
\definecolor{Gray}{gray}{0.9}

%

%
\runningauthor{Seungjae Shin, Heesun Bae, Donghyeok Shin, Weonyoung Joo and Il-Chul Moon}

\twocolumn[

\aistatstitle{Loss-Curvature Matching for Dataset Selection and Condensation}

\aistatsauthor{ Seungjae Shin$^{\textbf{*}}$ \And Heesun Bae$^{\textbf{*}}$ \And  Donghyeok Shin \And  Weonyoung Joo \And  Il-Chul Moon$^{\dagger}$}

\aistatsaddress{ KAIST \And KAIST \And KAIST \And Ewha Womans University \And KAIST }]

\begin{abstract}
Training neural networks on a large dataset requires substantial computational costs. Dataset reduction selects or synthesizes data instances based on the large dataset, while minimizing the degradation in generalization performance from the full dataset. Existing methods utilize the neural network during the dataset reduction procedure, so the model parameter becomes important factor in preserving the performance after reduction. By depending upon the importance of parameters, this paper introduces a new reduction objective, coined LCMat, which Matches the Loss Curvatures of the original dataset and reduced dataset over the model parameter space, more than the parameter point. This new objective induces a better adaptation of the reduced dataset on the perturbed parameter region than the exact point matching. Particularly, we identify the worst case of the loss curvature gap from the local parameter region, and we derive the implementable upper bound of such worst-case with theoretical analyses. Our experiments on both coreset selection and condensation benchmarks illustrate that LCMat shows better generalization performances than existing baselines.
\end{abstract}

\section{INTRODUCTION}
Although we live in the world of big data, utilizing such big data induces a considerable amount of time and space complexity in the learning process \citep{Craig,EDC,carbon}. Accordingly, researchers introduced a concept of \textit{dataset selection} and \textit{dataset condensation}, etc \citep{glister,Data-Diet}. These concepts state that a dataset with smaller cardinality may yield similar performance in machine learning compared to a big dataset, if the smaller dataset delivers all task-relevant information as the original dataset. Dataset reduction provides tangible benefits because the reduced dataset will consume less time in training and less space in memory \citep{adacore}. Moreover, such benefits are the desiderata of some well-known tasks, i.e. continual learning with memory replay \citep{clmemorybase, clcoreset1}.

As we reviewed, there exist two approaches in reducing the cardinality of dataset: the selection-based method (a.k.a. dataset selection) and the condensation-based method (a.k.a. dataset condensation). While these are similar concepts in terms of reducing data cardinality without performance degradation, both approaches have been treated and researched in different papers. Hence, this paper will refer to these approaches by a unifying term of \textit{dataset reduction}. 1) Selection-based method optimally selects a small set of data instances out of the full dataset with an expectation on the identical task-relevant information of the small and the full datasets \citep{contextualdiversity,Kcenter,Herding}. In contrast, 2) condensation-based method synthesizes the data instances by directly passing the learning gradient to the data input \citep{DM, kip}. 

To identify the examples which contribute the most to learning, both lines of work mainly utilize the gradient matching between the original dataset and reduced dataset \citep{Craig, gradmatchcoreset, DC}, which provides theoretical analyses unlike other methods \citep{Uncertainty, DM}. However, gradient matching is conducted at a specific model parameter, so this implementation would fundamentally be biased by the model parameter at hand. Therefore, the generalization over the perturbed parameter point could be potentially beneficial. From the perspective of generalization over the model parameter region, the gradient matching can be generally extended to the local curvature matching in the response surface. Recently, Sharpness-Aware Minimization (SAM) \citep{SAM} has made breakthroughs which ensure the generalization of the model by regularizing the flat minima over the local parameter region, not the point estimate of the parameter. This opens a new possibility of applying the spirit of SAM to the dataset reduction field.

This paper introduces a new objective for dataset reduction, coined \underline{\textbf{L}}oss-\underline{\textbf{C}}urvature  \underline{\textbf{Mat}}ching (LCMat), which matches the loss curvature of the original dataset and the resulting reduced dataset on the target parameter region. This matching could be also interpreted as the sharpness of the loss difference between two datasets. This notion enables LCMat as the first work of sharpness-aware dataset reduction. This merge of dataset reduction and sharpness-aware minimization induces two contributions. First, SAM only provides the optimization based on the model parameter, whereas the optimization of dataset reduction is conducted based on the input data variable. To enable the input-based optimization on the defined sharpness, this paper derives an implementable upper bound of the sharpness, which results in an objective of LCMat. Second, we adaptively transform the objective into the function of either selection or condensation objective, so LCMat becomes the fundamentally applicable mechanism for dataset reduction overarching the dataset selection as well as the dataset condensation. We conduct experiments over the evaluation scenarios with different benchmark datasets, and we confirm that LCMat shows clear merit when the reduction ratio becomes significant and when the evaluation scenario becomes dynamic and complex, e.g. continual learning.  

\section{PRELIMINARY}
\subsection{Notations}
This paper focuses on dataset reduction for classification tasks, which is a widely studied scenario in the community of dataset reduction \citep{Craig,Herding,DC}. Assuming a classification into $c$ classes, let $\mathcal{X} \in \mathbb{R}^{d}$ and $\mathcal{Y} = \{1,2,...,c\}$ be input variable space and a label candidate set, respectively. Given $\mathcal{X}$ and $\mathcal{Y}$, our training dataset is $T = \{(x_{i},y_{i})\}^{n}_{i=1} \subseteq \mathcal{X} \times \mathcal{Y}$. We assume that each training instance $(x,y)$ is drawn i.i.d from the population distribution $\mathbb{D}$. 

Let a classifier $f_{\theta} : \mathbb{R}^{d} \rightarrow \mathbb{R}^{c}$ be parameterized by $\theta \in \Theta$. Under this definition, the training loss on $T$ and the population loss on $\mathbb{D}$ are denoted as $\mathcal{L}(T;\theta)\!\!=\!\!\frac{1}{n}\sum^{n}_{i=1}\ell(x_{i},y_{i};\theta)$  and $\mathcal{L}(\mathbb{D};\theta)\!\!=\!\!\mathbb{E}_{(x,y) \sim \mathbb{D}}[\ell(x,y;\theta)]$, respectively. \!Here, $\ell$ means a value of loss function for a pair of $x$ and $y$\footnote{This paper utilizes cross-entropy as a loss function.}.

\subsection{Previous Researches on Dataset Reduction}
\label{2.1}
This paper focuses on \textit{dataset reduction}, whose purpose is to generate a cardinality-reduced dataset $S$ from the training dataset $T$, as such $|S| \ll |T|$, while maximally preserving the task-relevant information from $T$. 
\paragraph{Selection-based Methods}
Selection-based methods \citep{Herding, Kcenter} find a data subset $S \subset T$ that satisfies the cardinality constraint while maximizing the objective defined by the informativeness of $S$. The approximated objectives are defined by utilizing either 1) gradient \citep{Data-Diet, Craig, gradmatchcoreset}, 2) loss \citep{Forgetting}, 3) uncertainty \citep{Uncertainty}, and 4) decision boundary \citep{DeepFool, CAL}. This section surveys existing methods with emphasis on gradient-based objectives because our method is primarily relevant to them. Gradient-based methods minimize the distance between the gradients from the training dataset $T$; and the (weighted) gradients from $S$ as follows:
\begin{align}
\label{gradmatch}
\min_{\mathbf{w}, S}  \mathcal{D}\Big{(} \sum\limits_{(x, y)\in T}&\frac{\nabla_{\theta} \ell(x,y; \theta)}{|T|}, \sum\limits_{(x, y)\in S} \frac{w_{x}\nabla_{\theta} \ell(x, y; \theta)}{\|\mathbf{w}\|_{1}} \Big{)} \\
&\text{s.t.} \quad  S \subset T, \;  w_{x}\geq0 \nonumber
\end{align}
Here, $\mathbf{w}$ is the vector of learnable weights for the data instances in subset $S$; $\|\mathbf{w}\|_{1}$ is l1 norm of $\mathbf{w}$; and $\mathcal{D}$ measures the distance between two gradients. 

To solve the selection problem, \citet{Craig} converts Eq \eqref{gradmatch} into the submodular maximization problem, and this research utilizes the greedy approach to optimize Eq \eqref{gradmatch}. Compared to \citet{Craig}, \citet{gradmatchcoreset} utilizes orthogonal matching pursuit algorithm \citep{omp} and $L_{2}$ regularization term over $\mathbf{w}$ to stabilize the optimization. \citet{adacore} replaces $\nabla_\theta l(x,y;\theta)$ in Eq \eqref{gradmatch} with a preconditioned gradient with the Hessian matrix, which leverages the second-order information for optimization. Having said that, the optimization of Eq \eqref{gradmatch} is highly dependent on the given $\theta$, so the gradient matching could be potentially biased by the single snapshot $\theta$ because the small-sized $S$ would be vulnerable to selection bias to summarize $T$.
\begin{figure*}\centering
\hfill
\begin{subfigure}{0.5\textwidth}
\includegraphics[width=0.49\linewidth]{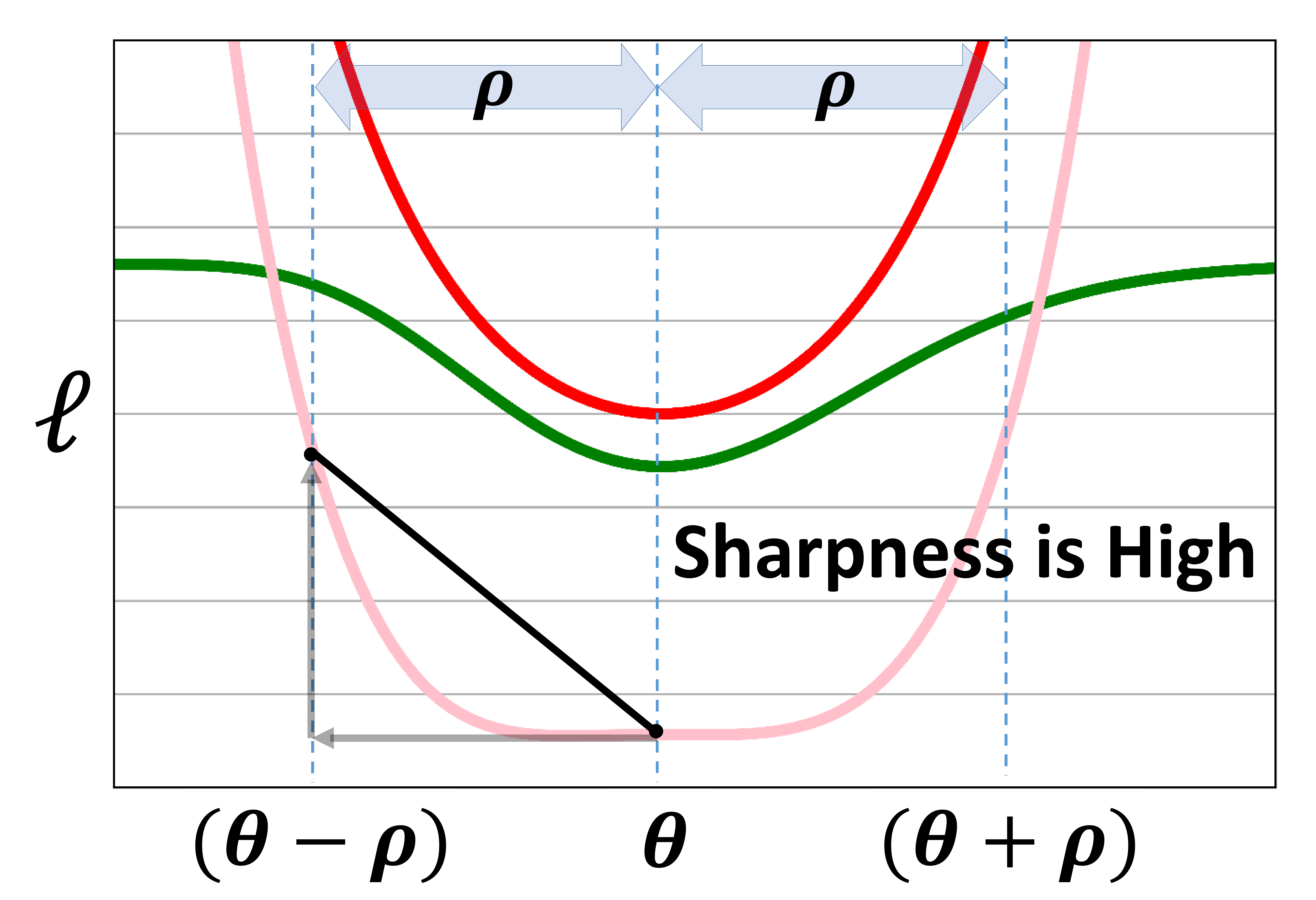}
\includegraphics[width=0.49\linewidth]{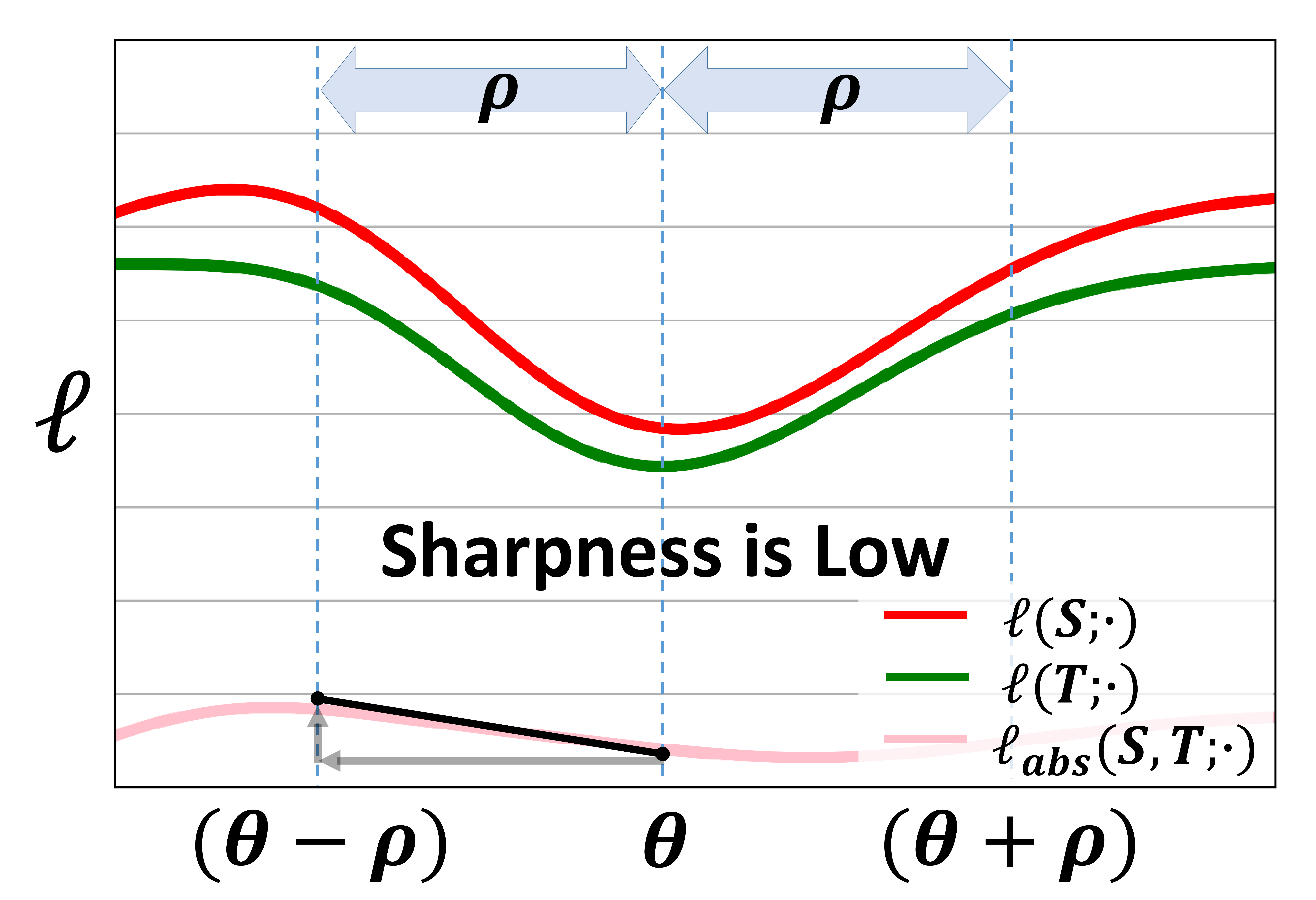}
\caption{Illustration of the sharpness on the loss difference. We measure the loss-curvature difference via the sharpness defined in Eq \eqref{intro_objective}.}\label{fig:1(a)}
\end{subfigure}%
\hfill
\begin{subfigure}{0.47\textwidth}
\includegraphics[width=0.49\linewidth]{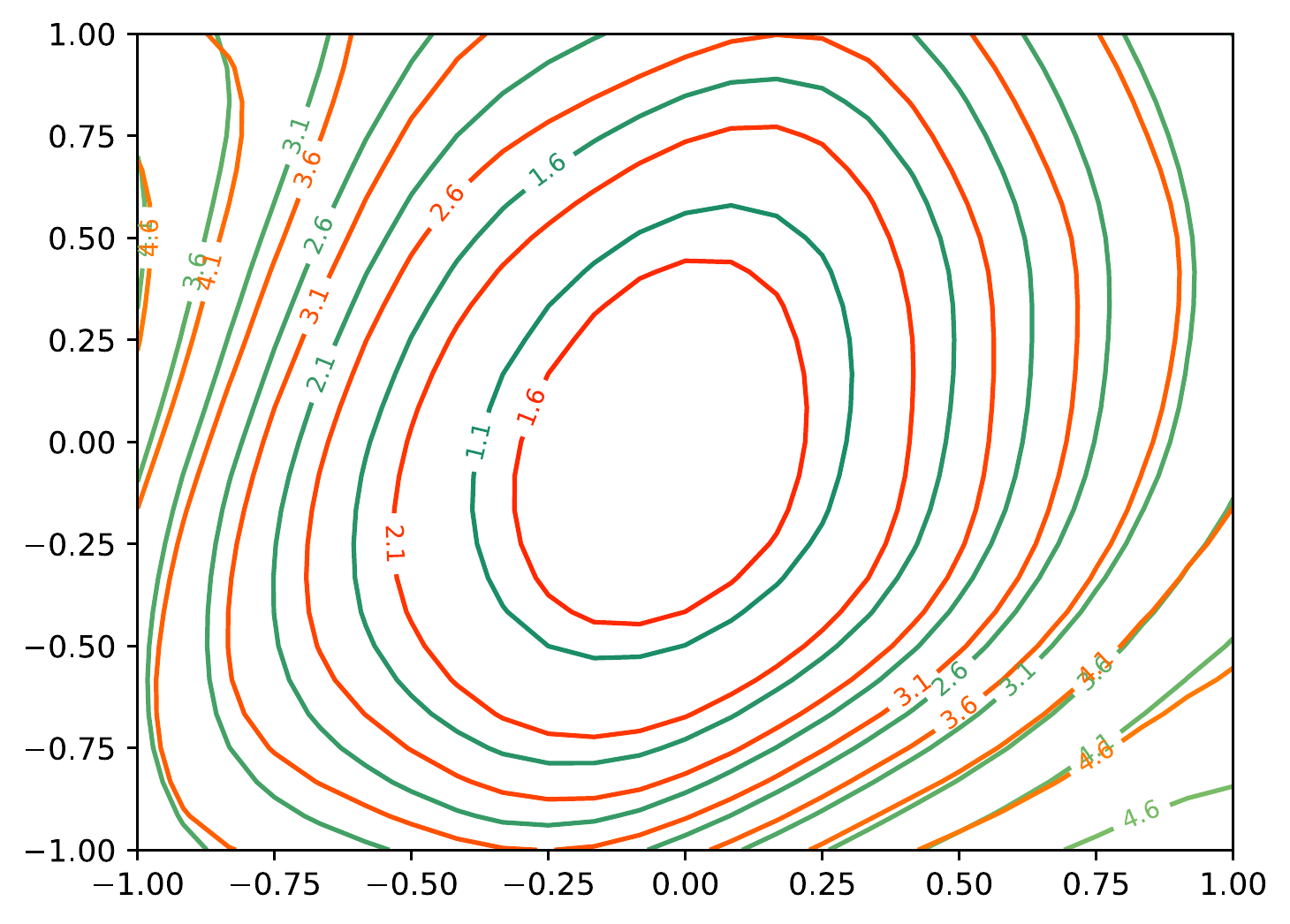}
\includegraphics[width=0.49\linewidth]{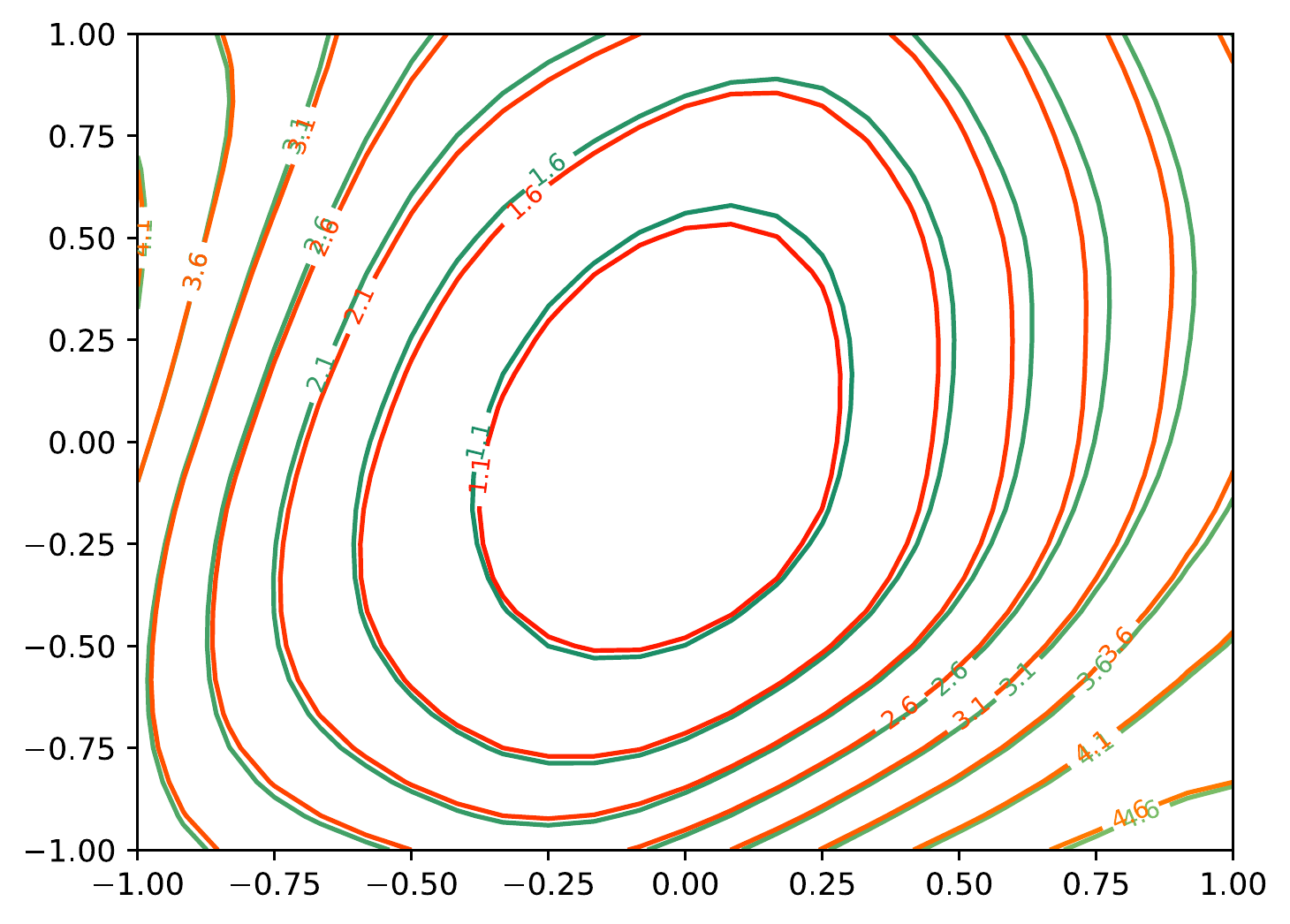}
\caption{Loss contour of $T$ (\textcolor{teal}{green}) and $S$ (\textcolor{red}{red}) for Craig (left) and LCMat-S (right), selected from 1\% fraction of CIFAR-10.}\label{fig:1(b)}
\end{subfigure}%
\caption{(a) The sharpness on loss differences represents the degree of difference on loss surfaces. (b) (left) The data subset, $S$, which is selected by Craig \citep{Craig}, does not match the loss curvature of the training dataset. (right) On the other hand, LCMat-S successfully matches the loss curvatures of $T$ and $S$. We visualize the loss landscape according to the implementation of \citet{lossvisual}.}\label{fig:motivation_fig}
\end{figure*}
\paragraph{Condensation-based Methods}
Instead of selecting $S$ from $T$, a small dataset, $S$ can be directly synthesized to achieve the similar performance from $T$ \cite{datadistill}. Then, $S$ becomes a learnable variable updated via $S \leftarrow S - \gamma \nabla_{S}\mathcal{L}(T,S)$, where $\mathcal{L}(T,S)$ is a general loss function which is dependent on both $T$ and $S$.  
\citet{DC} proposed Dataset Condensation (DC), which matches the gradients between $T$ and $S$ over the optimization path of $S$ as follows:
\begin{align}
\label{condensation}
\min_{S} \mathbb{E}_{\theta^{\small{0}} \sim P_{\theta^{\small{0}}} }\Big{[}\sum^{}_{k}\mathcal{D}(\nabla_{\theta^{k}_{S}} \mathcal{L}(T; \theta^{k}_{S}), \nabla_{\theta^{k}_{S}} \mathcal{L}(S; \theta^{k}_{S})) \Big{]} 
\end{align}
Here, $\theta^{\small{0}}$ is the initialized parameter from $P_{\theta^{\small{0}}}$; and $\theta^{k}_{S}$ is the parameter updated with $k$ iterations on SGD with $S$. The optimization of Eq \eqref{condensation} can be highly-dependent on the learning trajectory of $\theta$ from $S$. 
Other condensation methods\footnote{See Appendix \ref{appendix:condensation-based methods survey} for detailed surveys} utilize either 1) feature vectors \citep{DM, cafe} or 2) kernel products to propagate the task-relevant information of $T$ into $S$ \citep{kip}. However, these methods do not provide theoretical analyses of the relation between $T$ and $S$. 
\subsection{Generalization on Parameter Space}

Apart from dataset reduction, a new research area has emerged by considering generalization over parameter space and its optimization \citep{sun2021exploring, AWP, PNI}. Several studies have focused on the problem of $\theta$ over-fitting to $T$ \citep{SWA, SAM, kim2022fisher}, and they confirmed that optimization on the perturbed parameter region has a strong correlation to the generalization performance of the model. Sharpness-Aware Minimization (SAM) \citep{SAM} is an optimizer for the model parameter, which regularizes the locality region of $\theta$ to be the flat minima on the loss curvature as follows:
\begin{align}
\label{sam}
\min_{\theta}\underset{{||\epsilon||_{2}} \leq \rho }{\text{max}}\mathcal{L}(T;\theta\!+\!\epsilon)
\end{align}
Here, $\epsilon$ is the perturbation vector to the parameter; and $\rho$ denotes the maximum size of the perturbation vector. As the objective is a function defined by both input and model parameter, it is possible to solve the generalization of a model parameter through the optimization of input data. However, there is no such study, which improves the generalization of the perturbed parameter space via optimizing the input data variable, to the best of our knowledge. It should be noted that adversarial training \citep{TRADES} is different from our method because the perturbation for the worst case is conducted on the input space, not on the parameter space.

\section{METHOD}
As described in Section \ref{2.1}, recent methods in dataset reduction propagate the task-relevant information from $T$ to $S$ by aligning the gradients of a specific $\theta$. Given that dataset reduction hinges upon the utilization of $\theta$, the performance depends on the trained $\theta$ at the moment of reduction. Therefore, the optimal dataset reduction $S^{*}$ would be different from $S$, which is biased by $\theta$ at the specific state of $f_\theta$. Therefore, our research question becomes how to design a parameter-robust algorithm for dataset reduction while the algorithm still uses $\theta$ by the necessity of the implementation practice.

\subsection{Parameter Generalization in Dataset Reduction}
A loss function $\mathcal{L}$ quantifies the fitness of $\theta$ under a certain dataset. Accordingly, the optimization of $S$ toward $T$ with respect to $\theta$ would decrease $|\mathcal{L}(T;\theta) - \mathcal{L}(S;\theta)|$, which is the loss difference between $T$ and $S$ on $\theta$. However, if $|\mathcal{L}(T;\theta\!+\! \epsilon) - \mathcal{L}(S;\theta \!+ \!\epsilon)|$ increases with small perturbation $\epsilon$ on $\theta$, then this increment indicates the lack of generalization on $\theta + \epsilon$, or an over-fitted reduction of $S$ by $\theta$. This generalization failure on the locality of $\theta$ subsequently results in the large difference of loss surfaces between $T$ and $S$, as illustrated in Figure \ref{fig:1(a)}. Figure \ref{fig:1(a)} shows that the difference of loss surfaces between $T$ and $S$ could be measured by the sharpness of the loss differences, whose color is pink, on the target parameter region. 
\begin{remark}
\label{remark}
Assuming the strict convexity of $\mathcal{L}$ over $\Theta$, if $|\mathcal{L}(T;\theta)-\mathcal{L}(S;\theta)| = c $ for some fixed constant $c \geq 0$ and any $\theta \in \Theta$,\, $\text{argmin}_{\theta}\mathcal{L}(T;\theta) = \text{argmin}_{\theta}\mathcal{L}(S;\theta)$.
\end{remark}
Remark \ref{remark} explains that the optimal $\theta$ for $T$ and $S$ are the same if the loss difference is constant over the parameter space, which is the state when the loss curvatures of $T$ and $S$ are the same. If this condition is satisfied, we could safely utilize $S$ for learning $\theta$ where the generalization performance of $\theta$ from $S$ is guaranteed to be the same as that of $T$. This motivates us to match the loss curvatures between $T$ and $S$, whose objective is introduced in the next section. 
\subsection{Loss-Curvature Matching (LCMat)} 
This section introduces a parameter-robust objective for dataset reduction, coined \underline{\textbf{L}}oss-\underline{\textbf{C}}urvature \underline{\textbf{Mat}}ching (LCMat), which matches the loss curvature of $T$ and $S$ based on a currently presented $\theta$. The target region of the objective is specified by the $\rho$-ball perturbed region of $\theta$. In Eq \eqref{sam}, SAM optimizes the worst-case sharpness from the target region of $\theta$, where the worst-case optimization becomes efficient when the optimization is requested over the specific region \citep{GroupDRO, SAM}. Following the worst-case optimization scheme, we formulate the primary objective as follows:
\begin{align}
\label{intro_objective}
    &\underset{S}{\min}\underset{{||\epsilon||_{2}} \leq \rho }{\max}\frac{\mathcal{L}_{abs}(T,S;\theta\!+\!\epsilon) \!\!-\!\! {\mathcal{L}_{abs}(T,S;\theta)}}{\rho}
\end{align}
Here, we denote the loss difference between $T$ and $S$ on $\theta$, $\mathcal{L}_{abs}(T,S;\theta)=|\mathcal{L}(T;\theta)-\mathcal{L}(S;\theta)|$. In Eq \eqref{intro_objective}, $S$ is optimized to minimize the sharpness of $\mathcal{L}_{abs}(T,S;\theta)$ over the $\rho$-ball perturbed region from $\theta$. The optimization on Eq \eqref{intro_objective} incurs the maximization of $\mathcal{L}_{abs}(T,S;\theta)$, which could result in the overly under-fitted state of $S$ on $\theta$. 
In our implementation, $\mathcal{L}_{abs}(T,S;\theta)$ is bounded or regularized during the optimization. See Appendix \ref{appendix:LCMat detailed anal} for detailed analyses. Also, Eq \eqref{intro_objective} is defined on the case of single $\theta$ for simplicity, and it could be generalized to any $\theta \in \Theta$. 

The next question is how to optimize $S$ by Eq \eqref{intro_objective}. As our learning target is $S$, not $\theta$; it is intractable to utilize SAM because SAM only provides the gradient of $\theta$ for the corresponding sharpness. We introduce Proposition \ref{proposition}, which provides a tractable and differentiable upper bound of Eq \eqref{intro_objective} as follows:
\begin{proposition}
\label{proposition}
When $\mathbb{H}_D = \nabla^{2}_{\theta}\mathcal{L}(D;\theta)$ is a Hessian matrix of $\mathcal{L}(D;\theta)$, let $\mathbb{H}_{T,S} = \mathbb{H}_{T}\!-\! \mathbb{H}_{S}=\!\nabla^{2}_{\theta}\mathcal{L}(T;\theta)\! - \! \nabla^{2}_{\theta}\mathcal{L}(S;\theta)$ and $\lambda^{T,S}_{1}$ be the maximum eigenvalue of the matrix $\mathbb{H}_{T,S}$, then we have: (Proof in Appendix \ref{appendix:proposition1 proof})
\begin{align}
    \underset{{||\epsilon||_{2}} \leq \rho }{\max}&\frac{\mathcal{L}_{abs}(T,S;\theta\!+\!\epsilon) \!\!-\!\! {\mathcal{L}_{abs}(T,S;\theta)}}{\rho} \\
    & 
    \leq\underbrace{{\Big{\|}\nabla_{\theta}\mathcal{L}(T;\theta) - \nabla_{\theta}\mathcal{L}(S;\theta) \Big{\|}}_2}_{\text{Gradient Matching via $L_{2}$-norm.}} \nonumber \\ 
    & + \underbrace{\frac{1}{2}\rho\lambda^{T,S}_{1}}_{\text{Max eigenvalue}} 
    + \underset{{||\upsilon||_{2}} \leq 1}{\max}O(\rho^{2}\upsilon^3) \nonumber
\end{align}
\end{proposition} 

According to Proposition \ref{proposition}, the upper bound of the Eq \eqref{intro_objective} consists of 1) the $L_2$ norm of gradient differences between $T$ and $S$; 2) the maximum eigenvalue of $\mathbb{H}_{T,S}$; and 3) remaining higher-order terms. Given a certain selection of $\rho$ determining the locality scope of the $\theta$, Proposition \ref{proposition} argues that the gradient matching objective would not be enough for the loss surface matching if $\lambda^{T,S}_{1}$ holds a large proportion in the upper bound. 

\begin{wrapfigure}{h}{0.39\linewidth}
\vspace{-0.1in}
\hspace{-0.2in}
\centering
\includegraphics[width=\linewidth]{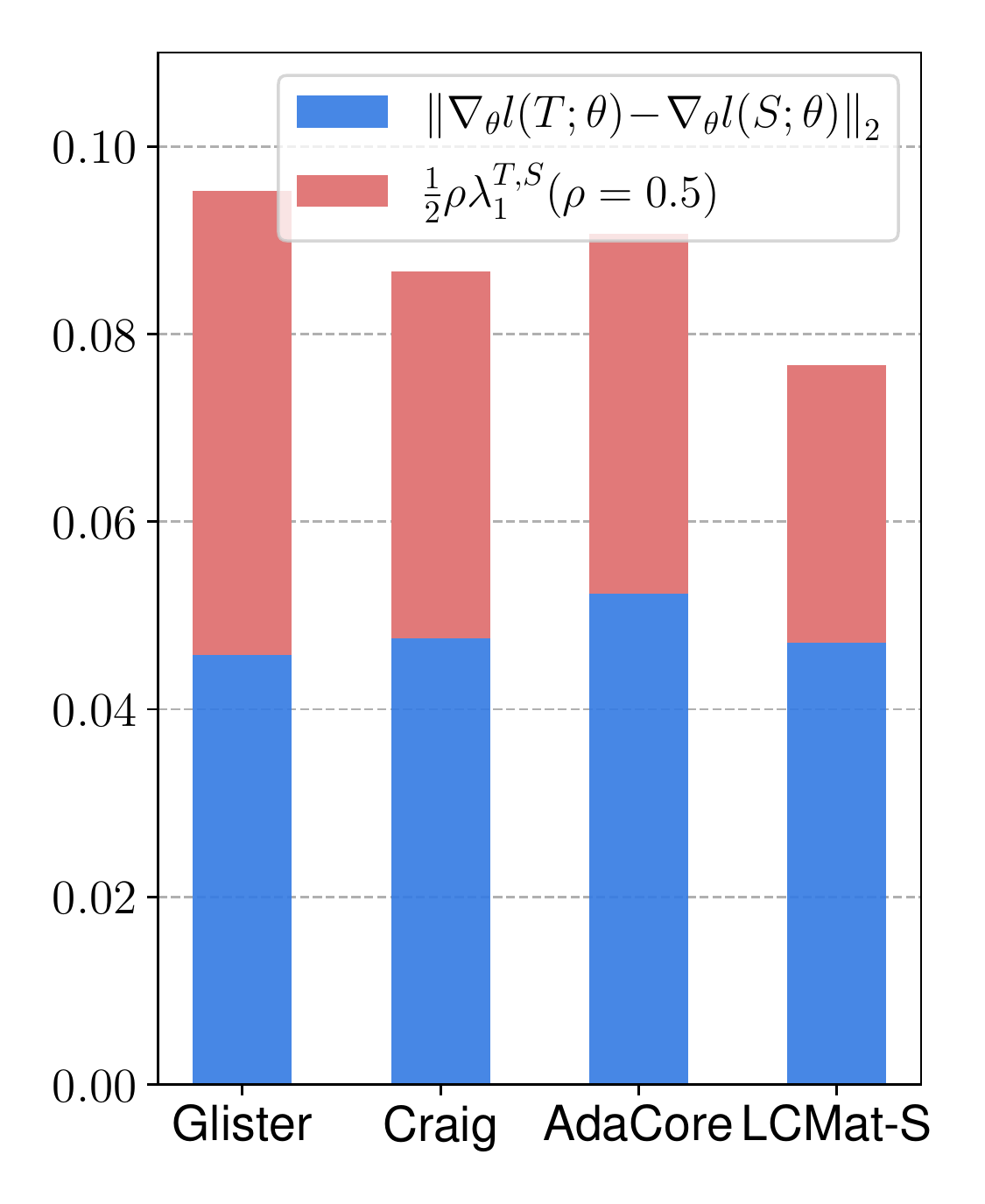}
\hspace{-0.2in}
\caption{Report on each term in Eq \eqref{final_objective} for the selected methods. $\lambda^{T,S}_{1}$ holds a significant proportion.}
\label{proportion}
\vspace{-0.1in}
\end{wrapfigure}

Figure \ref{proportion} shows the value of ${\|\nabla_{\theta}\mathcal{L}(T;\theta) - \nabla_{\theta}\mathcal{L}(S;\theta) \|}_2$ and $\frac{1}{2}\rho\lambda^{T,S}_{1}$ measured from different methods with $\rho=0.5$. For the gradient matching term, all methods show similar values, which means that these methods could not be distinguished by the learning from the gradient matching term. On the contrary, $\lambda^{T,S}_{1}$ holds a large proportion and takes high variance across the tested methods, so the upper bound differences among the methods eventually rely on the value of $\lambda^{T,S}_{1}$. By excluding higher-order terms in Proposition \ref{proposition}, the resulting alternative objective is as follows:
\begin{align}
\label{final_objective}
     \underset{S}{\min}\,\,{\Big{\|}\nabla_{\theta}\mathcal{L}(T;\theta) \!-\! \nabla_{\theta}\mathcal{L}(S;\theta) \Big{\|}}_2 +\frac{1}{2}\rho\lambda^{T,S}_{1}
\end{align}
Directly solving the optimization of Eq \eqref{final_objective} requires an explicit calculation of the Hessian matrices, $\mathbb{H}_{T}$ and $\mathbb{H}_{S}$. This calculation is too costly for over-parameterized models, such as neural networks. To overcome the computational overhead, various methods in machine learning have utilized the diagonal approximation of Hessian \citep{fishr,adahessian} as a common technique. According to \citet{fishr}, $\mathbb{H}$ becomes diagonally dominant at the end of training in most cases. We apply the diagonal approximation on $\mathbb{H}_{T}$ and $\mathbb{H}_{S}$, and we denote the corresponding diagonal Hessian as $\hat{\mathbb{H}}_{T}=\text{diag}(\mathbb{H}_{T})$ and $\hat{\mathbb{H}}_{S}=\text{diag}(\mathbb{H}_{S})$. When we replace $\mathbb{H}_{T}$ and $\mathbb{H}_{S}$ into $\hat{\mathbb{H}}_{T}$ and $\hat{\mathbb{H}}_{S}$, respectively, Eq \eqref{final_objective} is derived\footnote{See Appendix \ref{appendix:equation(7)} for the proof.} as follows:
\begin{align}
\label{main_objective}
     \underset{S}{\min}\,\,{\Big{\|}\nabla_{\theta}\mathcal{L}(T;\theta) \!-\! \nabla_{\theta}\mathcal{L}(S;\theta) \Big{\|}}_2 +\frac{1}{2}\rho\max_{k}\Big{|}\hat{\lambda}^{T}_{k}-\hat{\lambda}^{S}_{k}\Big{|} 
\end{align}
Here, $\hat{\lambda}^{T}_{k}$ and $\hat{\lambda}^{S}_{k}$ are eigenvalues of $\hat{\mathbb{H}}_{T}$ and $\hat{\mathbb{H}}_{S}$ on $k$-th dimension for $\theta$. Having said that, we provide an adaptive application of our objective, Eq \eqref{main_objective}, on two approaches: selection-based methods and condensation-based methods, in Sections \ref{3.2} and \ref{3.3}, respectively.
\subsection{LCMat for Selection-based method}
\label{3.2}
To select $S \subseteq T$, which minimizes Eq \eqref{main_objective}; we transform \eqref{main_objective} into the selection-based objective with the cardinality constraint on $S$, in a sample-wise derivation as follows:
\begin{align}
\label{selection_based_1}
\underset{S \subseteq T}{\min}\Big{(}\,& \Big{\|}\frac1{|T|}\sum\limits_{\substack{(x_{i}, y_{i})\\\in T}}\mathbf{g}^{T}_{i} -  \frac1{|S|}\sum\limits_{\substack{(x_{j}, y_{j})\\\in S}}\gamma_{j}\mathbf{g}^{S}_{j} \Big{\|}_2 \\& +
\frac{1}{2}
\rho \max_{k}\big{|}\frac1{|T|}\sum\limits_{\substack{(x_{i}, y_{i})\\\in T}}\hat{\lambda}^{T}_{i,k} -  \frac1{|S|}\sum\limits_{\substack{(x_{j}, y_{j})\\\in S}}\gamma_{j}\hat{\lambda}^{S}_{j,k}\big{|}\Big{)} \nonumber \\
 &\text{s.t. } |S| \ll |T| \nonumber
\end{align}
Here, we denote the per-sample gradient as $\mathbf{g}^{T}_{i} = \nabla_{\theta}\ell(x_{i},y_{i};
\theta)$ for $(x_{i}, y_{i}) \in T$, and we also denote the $k$-th dimension eigenvalue of the per-sample Hessian as $\hat{\lambda}^{T}_{i,k}$ for $(x_{i}, y_{i}) \in T$. Also, we introduce the learnable weight $\gamma_{j}$ for $(x_{j}, y_{j}) \in S$ to build Eq \eqref{selection_based_1} as a generalized form. 

It is well known the subset selection problem is NP-hard \citep {Kcenter, Craig}. When we maximize $\Big{|}\frac1{|T|}\sum\limits_{\substack{(x_{i}, y_{i})\in T}}\hat{\lambda}^{T}_{i,k} -  \frac1{|S|}\sum\limits_{\substack{(x_{j}, y_{j})\in S}}\gamma_{j}\hat{\lambda}^{S}_{j,k}\Big{|}$ with respect to $k$, $k$ will be different by each subset $S \subseteq T$, where the search for $k$ based on every possible $S \subseteq T$ would be very costly. 
To relax the computational constraints on a search for $k$, we empirically optimize the following equation, which does not need the search of $k$, on behalf of the second term in Eq \eqref{selection_based_1}:
\begin{align}
\label{relaxed_second_term}
\frac{1}{2}\rho\sum_{k \in \mathcal{K}}
 \Big{|}\frac1{|T|}\sum\limits_{\substack{(x_{i}, y_{i})\in T}}\hat{\lambda}^{T}_{i,k} -  \frac1{|S|}\sum\limits_{\substack{(x_{j}, y_{j})\in S}}\gamma_{j}\hat{\lambda}^{S}_{j,k}\Big{|}
\end{align}
Here, $\mathcal{K}$ is a set of indexes for $K$ sub-dimensions on $\theta$. We select $K$ dominant sub-dimensions based on the variance of $\hat{\boldsymbol{\lambda}}^{T}_{k}=[\hat{\lambda}^{T}_{i,k}]^{|T|}_{i=1}$ for each $k$, which is denoted by the set $\mathcal{K} = \underset{\mathcal{K},|\mathcal{K}|=K}{\text{argmax}}\sum_{j\in \mathcal{K}}\text{Var}(\hat{\boldsymbol{\lambda}}^{T}_{k})$. We empirically show that the true $k$ in Eq \eqref{selection_based_1} is always in $\mathcal{K}$, where the hyper-parameter of sub-dimensions $K$ is fixed to $100$ in our experiments. See Appendix \ref{appendix:selection-k-dim} for detailed analyses. 

By the notion of regarding the subset selection as sparse vector approximation \citep{sparse1,Craig}, existing methods utilize submodular optimization with a simple greedy algorithm to get a nearly-optimal solution on their objectives. Similar to \cite{Craig}, we utilize a facility location function \citep{submodular1, submodular2} for the submodular optimization. The facility location function quantifies the cover of $T$ given its subset $S$ by summation of the similarities defined between every $i \in T$ and its closest element $j \in S$. Formally, a facility location is defined as $F(S) = \sum_{i \in T}\max_{j \in S} s_{i,j}$, where $s_{i,j}$ is the similarity between $i,j \in T$. By utilizing the analytical result of Craig, we get an upper bound of the error for Eq \eqref{relaxed_second_term} as follows: (Proof in Appendix \ref{appdendix:equation(10)})
\begin{align}
\label{upper_bound}
  &  \underset{S \subseteq T}{\min}\,\Big{\|}\bar{\mathbf{g}}^{T} -  \mathbf{\gamma}^{S}\bar{\mathbf{g}}^{S} \Big{\|}_2 +
\frac{1}{2}\rho\sum_{k \in \mathcal{K}}  \Big{|}\bar{\lambda}^{T}_{k} -  \mathbf{\gamma}^{S}\bar{\lambda}^{S}_{k}\Big{|} \\ 
  & \leq \underset{i \in T}{\sum}\underset{j \in S}{\min}\Big{(}\Big{\|}{\mathbf{g}}^{T}_{i} -  {\mathbf{g}}^{S}_{j} \Big{\|}_2+\frac{1}{2}\rho\sum_{k \in \mathcal{K}}\Big{|}\hat{\lambda}^{T}_{i,k} -  \hat{\lambda}^{S}_{j,k}\Big{|}\Big{)} \nonumber
\end{align}
Here, $\bar{\mathbf{g}}^{T} = \frac1{|T|}\sum\limits_{\substack{(x_{i}, y_{i})\in T}}\mathbf{g}^{T}_{i}$ \footnote{$\bar{\mathbf{g}} = \frac{1}{N} \sum_{i=1}^N \mathbf{g}_i.$} and $\bar{\lambda}^{T}_{k} = \frac1{|T|}\sum\limits_{\substack{(x_{i}, y_{i})\in T}}\hat{\lambda}^{T}_{i,k}$. 
We aim at minimizing the upper bound from Eq \eqref{upper_bound}, where we denote the upper bound as $L(S)$. Finally, our algorithm will be implemented as follows:
\begin{align}
\label{our_implementation}
\underset{S \subseteq T}{\min}L(S)\,\,\, \text{ s.t. } \,\,\, |S|=m
\end{align}
Similar to \cite{adacore}, we re-formulate Eq \eqref{our_implementation} into the formalized version of facility location algorithm. Let's suppose an auxiliary example $e$, and the minimization of $L(S)$ is turned into the maximization of a facility location objective $F(S)$ as follows:
\begin{align}
\label{submodular_optimization}
\underset{S \subseteq T}{\max}F(S)=L(\{e\})-L(S\cup\{e\})\,\,\, \text{ s.t. } \,\,\, |S|=m
\end{align}
Here, $L(\{e\})$ is a constant, which is an upper bound of $L(S)$. The objective could also be derived as a submodular cover problem, whose objective is to minimize $|S|$ with the constraints on $F(S)$. Finally, we call our method applied to the selection-based method as LCMat-S.\footnote{The code is available at https://github.com/SJShin-AI/LCMat.}

\subsection{LCMat for Condensation-based method}
\label{3.3}
Different from selection-based methods, which need submodular optimization for a subset selection from $T$; condensation-based methods directly optimize $S$ by setting Eq \eqref{main_objective} to $\mathcal{L}(T,S;\theta)$. Eventually, the implemented objective becomes $min_S \mathcal{L}(T,S;\theta)$. Here, $S$ is updated as $S \leftarrow S - \gamma \nabla_{S}\mathcal{L}(T,S;\theta)$. However, the direct optimization of Eq \eqref{main_objective} still remains costly because of derivative computation over the Hessian terms, which are $\hat{\lambda}^{T}_{k}$ and $\hat{\lambda}^{S}_{k}$. This section provides an efficient variation of Eq \eqref{main_objective}, which is adapted to the community of condensation-based methods.

According to \citet{fishr}, the Fisher information $\mathbb{F} = \sum^{|T|}_{i=1}\mathbb{E}_{\hat{y} \sim p_{\theta}(y|x_{i})} \Big{[} \nabla_{\theta}\log p_{\theta}(\hat{y}|x_{i})\nabla_{\theta} \log p_{\theta}(\hat{y}|x_{i})^{\top}\Big{]}$ approximates the Hessian $\mathbb{H}$ with probably bounded errors under mild assumptions \citep{kim2022fisher}. As the Fisher information only requires the first derivative on $\theta$, the computation of Fisher information is more efficient than the computation of the Hessian matrix.
The equation below is the empirical Fisher information $\tilde{\mathbb{F}}$ of a certain dataset $D$:
\begin{align}
\label{empirical_fisher_information}
  \tilde{\mathbb{F}} = \frac{1}{|D|}\sum_{\substack{(x_{i}, y_{i})\in D}}\nabla_{\theta}\ell(x_{i},y_{i};\theta)\nabla_{\theta}\ell(x_{i},y_{i};\theta)^{\top}
\end{align}
$\tilde{\mathbb{F}}$ is equivalent\footnote{We skip the index with $D$ for the simplicity of $\tilde{\mathbb{F}}$, $\mathbf{C}$, and $\mathbf{G}$.} to the gradient covariance matrix $\mathbf{C}=\frac{1}{n-1}\Big{(}\mathbf{G}^{\top}\mathbf{G}-\frac{1}{n}(\mathbf{1}^{\top}\mathbf{G})^{\top}(\mathbf{1}^{\top}\mathbf{G})) \Big{)}$ of size $|\theta| \times |\theta|$ at any first-order stationary point \citep{fishr}, where $\mathbf{G} = [{\mathbf{g}}_{i}]^{|D|}_{i=1}$. As our objective \eqref{main_objective} is constructed based on the Hessian diagonals, such as $\hat{\mathbb{H}}_{T}$ and $\hat{\mathbb{H}}_{S}$; we consider the gradient variance, $\text{Var}(\mathbf{G})$, which is the diagonal components of $\mathbf{C}$ as follows:
\begin{align}
     \text{Var}(\mathbf{G}) = \frac{1}{|D|-1}\sum^{|D|}_{i=1}\big{(}\mathbf{g}_{i}-\bar{\mathbf{g}}\big{)}^{2}
\end{align}
Results from \citet{fishr} support that the similarity between Hessian diagonals and gradient variances is over 99.99$\%$. Similar to Eq \eqref{relaxed_second_term}, we could specify $\mathcal{K}$ to select the sub-dimensions of $\text{Var}(\mathbf{G})$ to match. In practice, we match the whole dimensions of $\text{Var}(\mathbf{G})$, which shows the robustness over the implemented experiments. We provide the adapted application of LCMat to the dataset condensation as follows:
\begin{align}
\label{condense_objective}
     &\min_{S} \mathbb{E}_{\theta^{\small{0}}}\Big{[}
     \sum^{}_{k}\mathcal{D}(\bar{\mathbf{g}}^{T}_{\theta_{k}},\bar{\mathbf{g}}^{S}_{\theta_{k}})
     +\frac{1}{2}\rho|\text{Var}(\mathbf{G}^{T}_{\theta_{k}})-\text{Var}(\mathbf{G}^{S}_{\theta_{k}})|\Big{]} \nonumber \\
     &\,\,\,\,\,\, \,\,\,  \text{ s.t. } \,\,\, \theta_{t+1} = \theta_{t}-\eta\bar{\mathbf{g}}^{T}_{\theta_{t}}\text{ for } t=0,...,k-1. 
\end{align}
We denote $\theta_{k}$ under each term to represent the subject of the derivative. Our objective is composed of 1) $\mathcal{D}(\bar{\mathbf{g}}^{T}_{\theta_{k}},\bar{\mathbf{g}}^{S}_{\theta_{k}})$, which is averaged gradient matching between $T$ and $S$; and 2) $|\text{Var}(\mathbf{G}^{T}_{\theta_{k}})-\text{Var}(\mathbf{G}^{S}_{\theta_{k}})|$, which is gradient variance matching between $T$ and $S$. Note that the averaged gradient matching is the objective of \citet{DC}. 
We also differentiate the learning trajectory of $\theta$ from $S$ to $T$ to satisfy the assumption on the model parameter in Section \ref{3.5}, which is utilized for the theoretical analysis of our method. We call our method applied to the condensation-based method as LCMat-C. 
\begin{table*}[] 
  \centering
  \caption{Test accuracies of Coreset selection tasks on CIFAR-10 and CIFAR-100 with the deployment of ResNet-18 over 5 different random seeds. The best results and second-bests from each setting are shown in \textbf{bold} and \underline{underlines}, respectively.} \label{tab:cifar}
  \resizebox{\textwidth}{!}{%
 \setlength{\tabcolsep}{2pt}
\begin{tabular}{c cccccccc ccccccc}
\toprule
     & \multicolumn{8}{c}{\textbf{CIFAR-10}} & \multicolumn{7}{c}{\textbf{CIFAR-100}}       \\ \cmidrule(lr){2-9} \cmidrule(lr){10-16}

Fraction     & 0.1\%       & 0.5\%       & 1\%       & 5\%       & 10\%      & 20\%      & 30\%     & 100\% &  0.5\%       & 1\%       & 5\%       & 10\%      & 20\%      & 30\%     & 100\%        \\ \cmidrule(lr){1-1} \cmidrule(lr){2-9} \cmidrule(lr){10-16}
Uniform &20.42\scriptsize{$\pm$2.0}&31.98\scriptsize{$\pm$1.9}&36.47\scriptsize{$\pm$1.9}&\underline{64.21}\scriptsize{$\pm$2.1}&\textbf{77.45}\scriptsize{$\pm$1.0}&\underline{87.36}\scriptsize{$\pm$0.4}&90.67\scriptsize{$\pm$0.2}&\multirow{14}{*}{95.48\scriptsize{$\pm$0.1}} &5.04\scriptsize{$\pm$0.5}&8.70\scriptsize{$\pm$0.5}&\underline{25.37}\scriptsize{$\pm$0.3}&34.09\scriptsize{$\pm$2.4}&55.98\scriptsize{$\pm$0.7}&\underline{64.59}\scriptsize{$\pm$0.1}&\multirow{14}{*}{78.91\scriptsize{$\pm$0.2}}\\ 
C-Div      &16.26\scriptsize{$\pm$2.2}&20.97\scriptsize{$\pm$2.3}&23.50\scriptsize{$\pm$2.8}&40.25\scriptsize{$\pm$1.3}&56.85\scriptsize{$\pm$1.7}&83.24\scriptsize{$\pm$1.7}&90.93\scriptsize{$\pm$0.5}&&4.76\scriptsize{$\pm$0.1}&6.01\scriptsize{$\pm$0.5}&13.62\scriptsize{$\pm$0.5}&20.53\scriptsize{$\pm$0.6}&44.91\scriptsize{$\pm$1.9}&58.60\scriptsize{$\pm$2.7}& \\
Herding &18.34\scriptsize{$\pm$2.1}&28.64\scriptsize{$\pm$1.5}&31.91\scriptsize{$\pm$3.8}&48.38\scriptsize{$\pm$2.6}&63.04\scriptsize{$\pm$2.5}&73.24\scriptsize{$\pm$1.8}&79.93\scriptsize{$\pm$1.5}&&4.42\scriptsize{$\pm$0.2}&6.93\scriptsize{$\pm$0.2}&18.24\scriptsize{$\pm$1.6}&26.47\scriptsize{$\pm$0.2}&42.83\scriptsize{$\pm$1.9}&52.14\scriptsize{$\pm$1.4}&\\
k-Center &19.38\scriptsize{$\pm$0.7}&25.80\scriptsize{$\pm$1.1}&31.61\scriptsize{$\pm$1.1}&55.55\scriptsize{$\pm$2.1}&72.12\scriptsize{$\pm$1.7}&86.79\scriptsize{$\pm$0.6}&90.83\scriptsize{$\pm$0.3}&&4.76\scriptsize{$\pm$0.3}&6.74\scriptsize{$\pm$0.8}&18.41\scriptsize{$\pm$0.4}&27.37\scriptsize{$\pm$1.5}&52.1\scriptsize{$\pm$0.8}&63.74\scriptsize{$\pm$0.7}&  \\
L-Conf &13.67\scriptsize{$\pm$2.0}&18.05\scriptsize{$\pm$1.4}&20.31\scriptsize{$\pm$1.8}&36.14\scriptsize{$\pm$2.2}&58.43\scriptsize{$\pm$3.0}&82.64\scriptsize{$\pm$1.2}&\underline{91.21}\scriptsize{$\pm$0.1}&&2.65\scriptsize{$\pm$0.1}&4.38\scriptsize{$\pm$0.1}&11.31\scriptsize{$\pm$0.4}&17.63\scriptsize{$\pm$2.1}&41.29\scriptsize{$\pm$1.1}&58.86\scriptsize{$\pm$1.0}&  \\
Entropy   &15.29\scriptsize{$\pm$1.1}&17.50\scriptsize{$\pm$2.0}&22.42\scriptsize{$\pm$2.0}&37.92\scriptsize{$\pm$2.4}&57.45\scriptsize{$\pm$3.6}&81.72\scriptsize{$\pm$2.2}&91.06\scriptsize{$\pm$0.7}& &2.51\scriptsize{$\pm$0.4}&3.82\scriptsize{$\pm$0.3}&11.32\scriptsize{$\pm$0.5}&16.94\scriptsize{$\pm$0.9}&41.88\scriptsize{$\pm$1.3}&57.45\scriptsize{$\pm$2.0}&  \\
Margin   &17.80\scriptsize{$\pm$2.1}&24.64\scriptsize{$\pm$1.2}&28.26\scriptsize{$\pm$2.9}&44.17\scriptsize{$\pm$2.8}&59.90\scriptsize{$\pm$6.7}&82.34\scriptsize{$\pm$0.9}&90.92\scriptsize{$\pm$0.4}&&3.86\scriptsize{$\pm$0.3}&6.11\scriptsize{$\pm$0.2}&14.57\scriptsize{$\pm$0.2}&20.70\scriptsize{$\pm$1.1}&46.36\scriptsize{$\pm$2.7}&59.45\scriptsize{$\pm$2.2}& \\ \cmidrule(lr){1-1} \cmidrule(lr){2-8} \cmidrule(lr){10-15}
Craig     &18.80\scriptsize{$\pm$2.4}&27.40\scriptsize{$\pm$1.9}&29.76\scriptsize{$\pm$2.0}&39.75\scriptsize{$\pm$3.7}&51.73\scriptsize{$\pm$4.6}&74.09\scriptsize{$\pm$0.9}&87.25\scriptsize{$\pm$0.8}&&\underline{6.38}\scriptsize{$\pm$0.4}&\underline{9.07}\scriptsize{$\pm$0.2}&15.93\scriptsize{$\pm$0.4}&20.32\scriptsize{$\pm$0.6}&32.23\scriptsize{$\pm$0.2}&47.09\scriptsize{$\pm$1.4}&  \\
GradMatch   &15.31\scriptsize{$\pm$0.6}&23.88\scriptsize{$\pm$1.2}&27.78\scriptsize{$\pm$2.0}&40.75\scriptsize{$\pm$3.1}&51.11\scriptsize{$\pm$2.3}&71.84\scriptsize{$\pm$3.5}&84.88\scriptsize{$\pm$1.4}&&4.28\scriptsize{$\pm$0.4}&6.26\scriptsize{$\pm$0.5}&14.19\scriptsize{$\pm$1.1}&20.23\scriptsize{$\pm$0.5}&40.28\scriptsize{$\pm$1.1}&51.03\scriptsize{$\pm$1.5}&  \\ 
GradMatch-Val &15.39\scriptsize{$\pm$1.2}&22.18\scriptsize{$\pm$1.1}&25.1\scriptsize{$\pm$1.7}&37.76\scriptsize{$\pm$1.2}&49.21\scriptsize{$\pm$2.4}&71.14\scriptsize{$\pm$1.7}&83.34\scriptsize{$\pm$1.4}& &4.43\scriptsize{$\pm$0.5}&5.57\scriptsize{$\pm$0.2}&13.45\scriptsize{$\pm$0.6}&22.99\scriptsize{$\pm$0.6}&39.84\scriptsize{$\pm$2.0}&51.72\scriptsize{$\pm$1.8}&   \\ 
Glister   &19.08\scriptsize{$\pm$2.1}&26.35\scriptsize{$\pm$1.7}&29.46\scriptsize{$\pm$3.4}&40.74\scriptsize{$\pm$3.1}&56.89\scriptsize{$\pm$2.7}&78.27\scriptsize{$\pm$0.5}&89.73\scriptsize{$\pm$0.4}&&4.22\scriptsize{$\pm$0.4}&6.46\scriptsize{$\pm$0.7}&16.49\scriptsize{$\pm$0.5}&24.07\scriptsize{$\pm$0.4}&44.42\scriptsize{$\pm$1.4}&56.81\scriptsize{$\pm$1.2}& \\
Glister-Val &17.53\scriptsize{$\pm$1.2}&23.97\scriptsize{$\pm$0.8}&28.64\scriptsize{$\pm$1.7}&39.74\scriptsize{$\pm$1.1}&52.98\scriptsize{$\pm$2.1}&77.54\scriptsize{$\pm$2.3}&87.46\scriptsize{$\pm$1.1}&  &4.54\scriptsize{$\pm$0.2}&5.5\scriptsize{$\pm$0.5}&14.78\scriptsize{$\pm$1.1}&25.72\scriptsize{$\pm$1.0}&43.22\scriptsize{$\pm$1.0}&55.98\scriptsize{$\pm$1.2}& \\ 
AdaCore   & \underline{22.54}\scriptsize{$\pm$0.9}&\underline{32.02}\scriptsize{$\pm$1.1}&\underline{39.09}\scriptsize{$\pm$1.0}&63.97\scriptsize{$\pm$1.1}&76.44\scriptsize{$\pm$1.5}&87.21\scriptsize{$\pm$0.2}&90.54\scriptsize{$\pm$0.4}&
&5.43\scriptsize{$\pm$0.2}&7.96\scriptsize{$\pm$0.2}&23.96\scriptsize{$\pm$1.0}&\underline{35.26}\scriptsize{$\pm$1.8}&\underline{56.54}\scriptsize{$\pm$0.6}&64.06\scriptsize{$\pm$0.9}& \\ \cmidrule(lr){1-1} \cmidrule(lr){2-8} \cmidrule(lr){10-15}
\rowcolor{Gray} LCMat-S &\textbf{23.87}\scriptsize{$\pm$1.1} &\textbf{33.17}\scriptsize{$\pm$0.6} &\textbf{39.54}\scriptsize{$\pm$0.7} &\textbf{64.72}\scriptsize{$\pm$1.3} &\underline{77.41}\scriptsize{$\pm$2.0} &\textbf{88.12}\scriptsize{$\pm$0.2} &\textbf{91.32}\scriptsize{$\pm$0.2}& &\textbf{7.65}\scriptsize{$\pm$0.8} &\textbf{11.82}\scriptsize{$\pm$0.8} &\textbf{27.3}\scriptsize{$\pm$1.2} &\textbf{36.66}\scriptsize{$\pm$1.0} &\textbf{56.66}\scriptsize{$\pm$0.6} &\textbf{64.81}\scriptsize{$\pm$0.9} & \\ \bottomrule
\end{tabular}
}
\end{table*}
\subsection{Theoretical Understanding of LCMat}
\label{3.5}
This section analyzes the generalization bound of Eq \eqref{intro_objective}, which is our primary objective. First, we define $\hat{\Theta}$, which is the application range of generalization bound as follows:
\begin{definition}
    $\hat{\Theta} = \{\theta : \mathcal{L}(T;\theta) \leq \mathcal{L}(\mathbb{D};\theta) \,\text{ for }\, \theta \in \Theta\}$
\end{definition}
In practice, $\mathcal{L}(T;\theta)$ and $\mathcal{L}(\mathbb{D};\theta)$ are approximated by the training loss and test loss, respectively. $\hat{\Theta}$ specifies $\theta$ whose generalization gap is more than equal to zero, which is intuitive when we optimize $\theta$ based on $T$. We first derive the generalization bound of $\underset{{||\epsilon||_{2}} \leq \rho }{\max}\mathcal{L}_{abs}(T,S;\theta\!+\!\epsilon)$, which is subpart of Eq \eqref{intro_objective}, as follows: 
\begin{theorem}
\label{theorem_v1} (Generalization Bound of $\underset{{||\epsilon||_{2}} \leq \rho }{\max}\mathcal{L}_{abs}(T,S;\theta\!+\!\epsilon)$) For $\theta \in \hat{\Theta}$, with probability at least $1-\delta$ over the choice of the training set $T$ with $|T|=n$, the following holds. (Proof in Appendix \ref{appendix:theorem1})
\begin{align}
\label{theorem_eq}
&\mathbb{E}_{\epsilon \sim \mathcal{N}(0,\rho)}[\mathcal{L}_{abs}(\mathbb{D},S;\theta\!+\!\epsilon)]  \\ &\leq \max_{\|\epsilon\|_{2}\leq \rho}\mathcal{L}_{abs}(T,S;\theta\!+\!\epsilon) + \sqrt{\frac{O(k+\log\frac{n}{\delta})}{n-1}}
\nonumber
\end{align}
\end{theorem}
Please note that proof of Theorem \ref{theorem_v1} largely referred to the proof concept of SAM \citep{SAM}. Having said that, Theorem \ref{theorem_v1} states that $\underset{{||\epsilon||_{2}} \leq \rho }{\max}\mathcal{L}_{abs}(T,S;\theta\!+\!\epsilon)$ can become the upper bound of $\mathbb{E}_{\epsilon \sim \mathcal{N}(0,\rho)}[\Big{|}\mathcal{L}(\mathbb{D};\theta\!+\!\epsilon)-\mathcal{L}(S;\theta\!+\!\epsilon)\Big{|}]$, which is the expected loss difference between $\mathbb{D}$ and $S$ over the $\epsilon$-perturbed space of the current parameter $\theta$. 

From the theoretical view, Theorem \ref{theorem_v1} provides the generalization property of the loss difference between two arbitrary datasets. As an extension of Theorem \ref{theorem_v1}, Corollary 1 directly investigates the generalization property of our main objective in Eq (4), which is the first term in R.H.S of Corollary 1, with an additional assumption, $\mathcal{L}_{abs}(T,S;\theta)\leq \mathcal{L}_{abs}(\mathbb{D},S;\theta)$. The assumption is acceptable if the loss difference from $\mathbb{D}$ is larger than $T$'s. 
\begin{corollary}
\label{corollary_v1} (Generalization Bound of Eq \eqref{intro_objective}) If $\mathcal{L}_{abs}(T,S;\theta) \!\leq\! \mathcal{L}_{abs}(\mathbb{D},S;\theta)$ for $\theta \in \hat{\Theta}$, with probability at least $1-\delta$ over the choice of the training set $T$ with $|T|=n$, the following holds: (Proof in Appendix \ref{appendix:corollary})
\begin{align}
&\Big{(}\mathbb{E}_{\epsilon \sim \mathcal{N}(0,\rho)}[\mathcal{L}_{abs}(\mathbb{D},S;\theta+\epsilon)]-\mathcal{L}_{abs}(\mathbb{D},S;\theta)\Big{)}\Big{/}\rho 
\\ &\leq \Big{(}\underset{{||\epsilon||_{2}} \leq \rho }{\max}\mathcal{L}_{abs}(T,S;\theta+\epsilon)-\mathcal{L}_{abs}(T,S;\theta)\Big{)}\Big{/}\rho \nonumber \\ 
&+ \sqrt{\frac{O(k+\log\frac{n}{\delta})}{n-1}}
\nonumber
\end{align}
\end{corollary}
According to Corollary \ref{corollary_v1}, Eq \eqref{intro_objective} can be an upper bound of $\Big{(}\mathbb{E}_{\epsilon \sim \mathcal{N}(0,\rho)}[\mathcal{L}_{abs}(\mathbb{D},S;\theta+\epsilon)]-\mathcal{L}_{abs}(\mathbb{D},S;\theta)\Big{)}\Big{/}\rho$, which is the expected sharpness of loss differences between $\mathbb{D}$ and $S$ over the $\epsilon$-perturbed space of the parameter $\theta$. This implies that the minimization of Eq \eqref{intro_objective} would lead to the local curvature matching between S and $\mathbb{D}$, when $\mathbb{D}$ is our target population distribution.

\section{EXPERIMENTS}
This section investigates the validity of our method, LCMat, through experiments on various datasets and tasks. First, we check the efficacy of LCMat through the application of LCMat on coreset selection and dataset condensation tasks. In addition, we investigate the performance of LCMat on a continual learning framework as a practical application.

\subsection{Coreset Selection Evaluation}
\label{coreset_evaluation}
\textbf{Experiment Details}
To investigate the efficacy of each selection-based algorithm, we follow the selection evaluation scenario of \cite{deepcore}, which is provided as follows. Each selection-based method learns $S$ by utilizing the neural network, $f_{\theta_{T}}$, which is pre-trained on $T$. Next, we introduce another randomly initialized neural network $f_{\theta_{S}}$; and we optimize $\theta_{S}$ with $S$. Finally, we measure the test accuracy on $f_{\theta_{S}}$ to evaluate the quality of $S$. During the selection, we assume that $\theta$ is fixed without alternative optimization between $S$ and $\theta$. It should be noted that our method could also be evaluated on the dynamic coreset selection scenario \citep{Craig,adacore}.

\textbf{Baselines}
We choose the baselines in the past works of selection-based methods. The selected baselines can be divided into two modelling categories. Baselines in the first category utilize the output from the forward-pass of the model, e.g. layer-wise feature vector, softmax output (Contextual Diversity (C-Div) \citep{contextualdiversity}, Herding \citep{Herding}, k-CenterGreedy (k-Center) \citep{Kcenter}, Least Confidence (L-Conf), Entropy, and Margin \citep{Uncertainty}). Baselines in another category are a set of variants for gradient matching (Craig \citep{Craig}, GradMatch \citep{gradmatchcoreset}, Glister \citep{glister} and AdaCore \citep{adacore}). We also report results from a randomly chosen subset (Uniform). For all methods, We select $S$ in a class-balanced manner. We provide the detailed implementation of each method and the corresponding wall-clock time in Appendix \ref{experimental_details}.

\textbf{Implementation of LCMat-S and Gradient-based Methods}
We compute the gradient and the Hessian matrix of the last layer of $f_{\theta}$, which is common practice in the theoretical analyses \citep{Craig,adacore}. For AdaCore \citep{adacore} and our method, LCMat-S; we skip the training of $\mathbf{w}$, which is learnable weights for the instances in subset $S$ because it significantly decreases the test performances. We conjecture that the problem is caused by the over-fitting of $\mathbf{w}$. We tune $\rho$, which is the only hyper-parameter of LCMat-S, from the value list of [0.01, 0.05, 0.1, 0.5]. We also implement the variants of GradMatch and Glister, which we call as GradMatch-Val and Glister-Val, by matching the gradient of $T$ with the gradient over the validation dataset as specified in the original paper.

\textbf{Benchmark Evaluation Result} 
Table \ref{tab:cifar} reports the test accuracy of the ResNet-18 trained using $S$ from each method. We evaluate $S$ with different fractions in dataset reduction, which is the cardinality budget of $S$ from $T$. Uniform, which is a random selection baseline, shows competitive performances over other baselines. This shows the weak robustness of the existing selection methods. LCMat-S shows the improved or competitive performances over the implemented baselines by relieving the over-fitting issue of $S$ to the provided $\theta$. Particularly, the gain from LCMat-S becomes significant when the tested dataset becomes difficult and the reduction rate becomes small, i.e. the dataset reduction to 0.5\%, 1\%, and 5\% in CIFAR-100. In Appendix \ref{selected_images}, we report image samples selected by each method of all classes for CIFAR-10 dataset. LCMat-S selects a set of examples with diverse characteristics, e.g. the diverse shape of each object and different backgrounds without redundancy.

\begin{table}[h]
  \centering
  \caption{Cross-architecture generalization performance (\%) on CIFAR-100 with ResNet-18. \textbf{Bold} represents best result. Experiments are repeated over 3 times.}
 \resizebox{0.49\textwidth}{!}{
  \begin{tabular}{c|c|cccc}
    \toprule
    Fraction &Test Model & ResNet-18 & VGG-16 & Inception-v3 & WRN-16-8 \\\midrule
    \multirow{6}{*}{1$\%$}&Uniform&8.35\scriptsize{$\pm$0.4}&3.5\scriptsize{$\pm$1.1}&6.22\scriptsize{$\pm$0.3}&8.57\scriptsize{$\pm$0.2} \\ 
    &Craig& 9.65\scriptsize{$\pm$0.3}&2.53\scriptsize{$\pm$0.6}&6.07\scriptsize{$\pm$0.5}&10.37\scriptsize{$\pm$0.2}\\ 
    & GradMatch &6.72\scriptsize{$\pm$0.2}&2.11\scriptsize{$\pm$0.6}&4.70\scriptsize{$\pm$0.5}&7.14\scriptsize{$\pm$0.2}\\
    & Glister &6.66\scriptsize{$\pm$0.1}&3.98\scriptsize{$\pm$0.6}&5.24\scriptsize{$\pm$0.2}&6.96\scriptsize{$\pm$0.4}\\
    & AdaCore &7.85\scriptsize{$\pm$0.1}&2.53\scriptsize{$\pm$0.6}&5.88\scriptsize{$\pm$0.3}&8.61\scriptsize{$\pm$0.1}\\
    & LCMat-S &\textbf{12.17}\scriptsize{$\pm$0.1}&\textbf{5.09}\scriptsize{$\pm$1.0}&\textbf{9.04}\scriptsize{$\pm$0.2}&\textbf{12.53}\scriptsize{$\pm$0.2}\\\midrule
    \multirow{6}{*}{5\%}&Uniform&25.85\scriptsize{$\pm$0.0}&18.22\scriptsize{$\pm$0.8}&21.00\scriptsize{$\pm$0.4}&30.13\scriptsize{$\pm$0.5} \\ 
    &Craig &17.08\scriptsize{$\pm$0.6}&10.00\scriptsize{$\pm$0.7}&12.11\scriptsize{$\pm$1.2}&18.85\scriptsize{$\pm$0.4} \\
    & GradMatch &15.63\scriptsize{$\pm$0.0}&12.59\scriptsize{$\pm$0.2}&13.43\scriptsize{$\pm$0.2}&19.16\scriptsize{$\pm$0.7} \\
    & Glister &17.01\scriptsize{$\pm$0.3}&13.82\scriptsize{$\pm$0.7}&14.14\scriptsize{$\pm$0.3}&20.53\scriptsize{$\pm$0.7}\\
    & AdaCore &24.71\scriptsize{$\pm$0.4}&19.38\scriptsize{$\pm$1.2}&21.66\scriptsize{$\pm$0.9}&29.77\scriptsize{$\pm$1.2}\\
    & LCMat-S &\textbf{27.29}\scriptsize{$\pm$0.7}&\textbf{20.42}\scriptsize{$\pm$1.2}&\textbf{24.87}\scriptsize{$\pm$0.7}&\textbf{33.20}\scriptsize{$\pm$0.7}\\ 
    \bottomrule
    \end{tabular}
}
\label{tab:cross}
\end{table}

\begin{table*}[h]
\centering
\caption{Test accuracies of coreset selection task on VGG-16 network (first, second row) and TinyImageNet dataset (third row), respectively. We denote the best performance as \textbf{Bold}; and the second best performance as \underline{Underline}, respectively.}
\resizebox{0.9\linewidth}{!}{
\begin{tabular}{c|c|ccccccc}
\toprule
Experiment &Frac&Uniform&k-Center&Craig&GradMatch&Glister&AdaCore&LCMat-S\\ \midrule
\multirow{2}{*}{CIFAR-10 w/ \textbf{VGG-16}} &0.5$\%$ &13.61$\pm$\scriptsize{1.8}&12.81$\pm$\scriptsize{1.1}&\textbf{15.83}$\pm$\scriptsize{1.9}&11.33$\pm$\scriptsize{0.6}&12.4$\pm$\scriptsize{0.7}&13.84$\pm$\scriptsize{1.6}&\underline{15.37}$\pm$\scriptsize{0.0}\\
&1$\%$ &19.81$\pm$\scriptsize{2.4}&15.78$\pm$\scriptsize{4.1}&15.19$\pm$\scriptsize{1.4}&13.7$\pm$\scriptsize{1.8}&\underline{22.84}$\pm$\scriptsize{3.1}&19.08$\pm$\scriptsize{8.4}&\textbf{25.41}$\pm$\scriptsize{6.4}\\ \midrule
\multirow{2}{*}{CIFAR-100 w/ \textbf{VGG-16}} &0.5$\%$ &1.85$\pm$\scriptsize{0.4}&1.51$\pm$\scriptsize{0.2}&2.13$\pm$\scriptsize{0.6}&\textbf{2.41}$\pm$\scriptsize{0.8}&2.03$\pm$\scriptsize{0.6}&1.79$\pm$\scriptsize{0.3}&\underline{2.34}$\pm$\scriptsize{0.2}\\
&1$\%$ &3.6$\pm$\scriptsize{1.5}&2.07$\pm$\scriptsize{0.6}&\underline{4.73}$\pm$\scriptsize{1.0}&2.63$\pm$\scriptsize{0.5}&4.36$\pm$\scriptsize{1.1}&2.9$\pm$\scriptsize{0.8}&\textbf{5.91}$\pm$\scriptsize{0.3}\\\midrule
\multirow{2}{*}{\textbf{TinyImageNet} w/ ResNet-18} &0.5$\%$&2.07$\pm$\scriptsize{0.2}&1.72$\pm$\scriptsize{0.2}&\underline{2.99}$\pm$\scriptsize{0.2}&2.44$\pm$\scriptsize{0.2}&2.75$\pm$\scriptsize{0.0}&1.81$\pm$\scriptsize{0.1}&\textbf{3.18}$\pm$\scriptsize{0.4}\\
&1$\%$&3.57$\pm$\scriptsize{0.1}&2.45$\pm$\scriptsize{0.2}&5.16$\pm$\scriptsize{0.3}&4.81$\pm$\scriptsize{0.1}&\underline{5.20}$\pm$\scriptsize{0.3}&3.43$\pm$\scriptsize{0.1}&\textbf{5.43}$\pm$\scriptsize{0.4}\\\bottomrule
\end{tabular}
}

\label{tab:additional_table}
\end{table*} 
\textbf{Robustness on Cross-Architecture}
From our scenario, the network structure of $f_{\theta_{S}}$ could be different from $f_{\theta_{T}}$. We test the robustness of LCMat-S on the specific scenario, which we call as Cross-Architecture Generalization \citep{DC}.  We utilize VGG-16 \citep{vgg}, Inception-v3 \citep{inception}, and WRN-16-8 \citep{wrn} as $f_{\theta_{S}}$. Table \ref{tab:cross} reports the test accuracy of LCMat-S and other gradient-based methods. LCMat-S consistently shows better generalization performances than the implemented baselines. We conjecture that the robustness over the different network architectures could be improved from our loss-curvature matching objective.

\begin{figure}{h}
\vspace{-0.2in}
\hspace{-0.3in}
\centering
\includegraphics[width=0.85\linewidth]{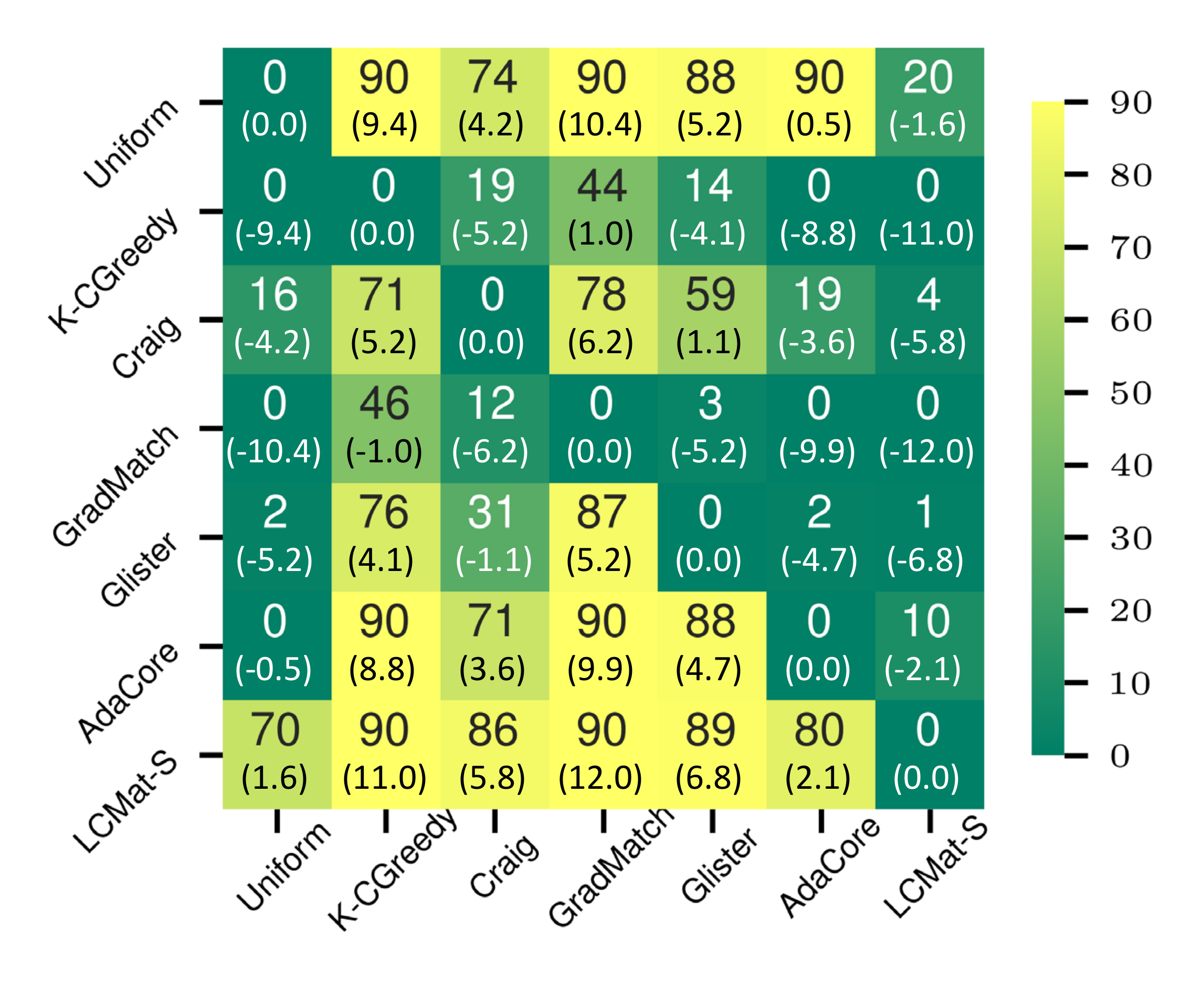}
\hspace{-0.3in}
\caption{Heatmap plot which shows the number of times that each method beats the others from each case; and the averaged improvements over the other methods in parenthesis (\%). }
\label{fig:heatmap}
\vspace{-0.1in}
\end{figure}

\textbf{Robustness on the pre-training of $f_{\theta_{T}}$} From our evaluation scenario, $f_{\theta_{T}}$ could be pre-trained with different hyper-parameters for each experiment, where $f_{\theta_{T}}$ significantly influences the selection of $S$. To test the robustness over the $\theta$ pre-training, We conduct the coreset selection experiments over the differently pre-trained ResNet-18 with combinations of epochs [2,5,10,20,100]; weight decay [1e-4, 5e-4,1e-3]; optimizers [SGD, Adam]; and 3 seeds, which result in 90 cases. Fig \ref{fig:heatmap} shows the number of times that each method beats the others from each case; and the averaged improvements over the other methods in parenthesis (\%). LCMat-S beats other baselines with large numbers.

\begin{figure}[h]
\centering
\begin{subfigure}[b]{0.475\columnwidth}
    \centering
    \includegraphics[width=\columnwidth]{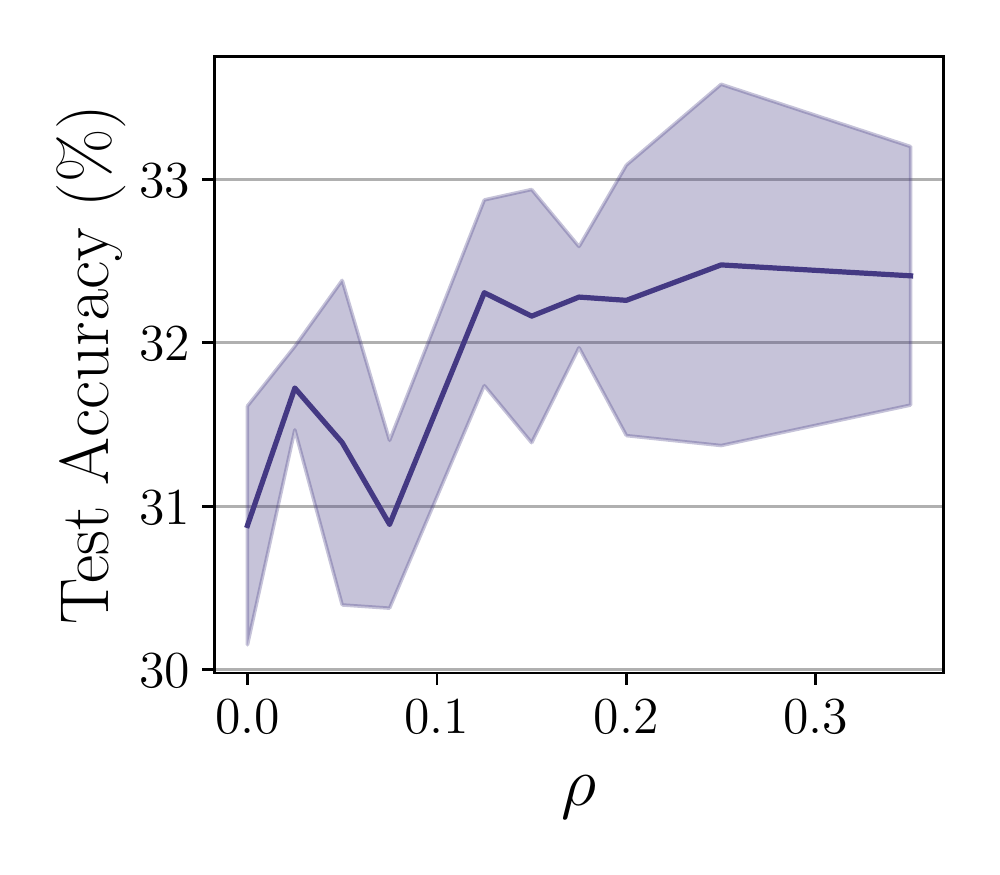}
    \vskip-0.1in
    \caption{Fraction = 0.5$\%$}
\end{subfigure}\hfill
\begin{subfigure}[b]{0.475\columnwidth}
    \centering
    \includegraphics[width=\columnwidth]{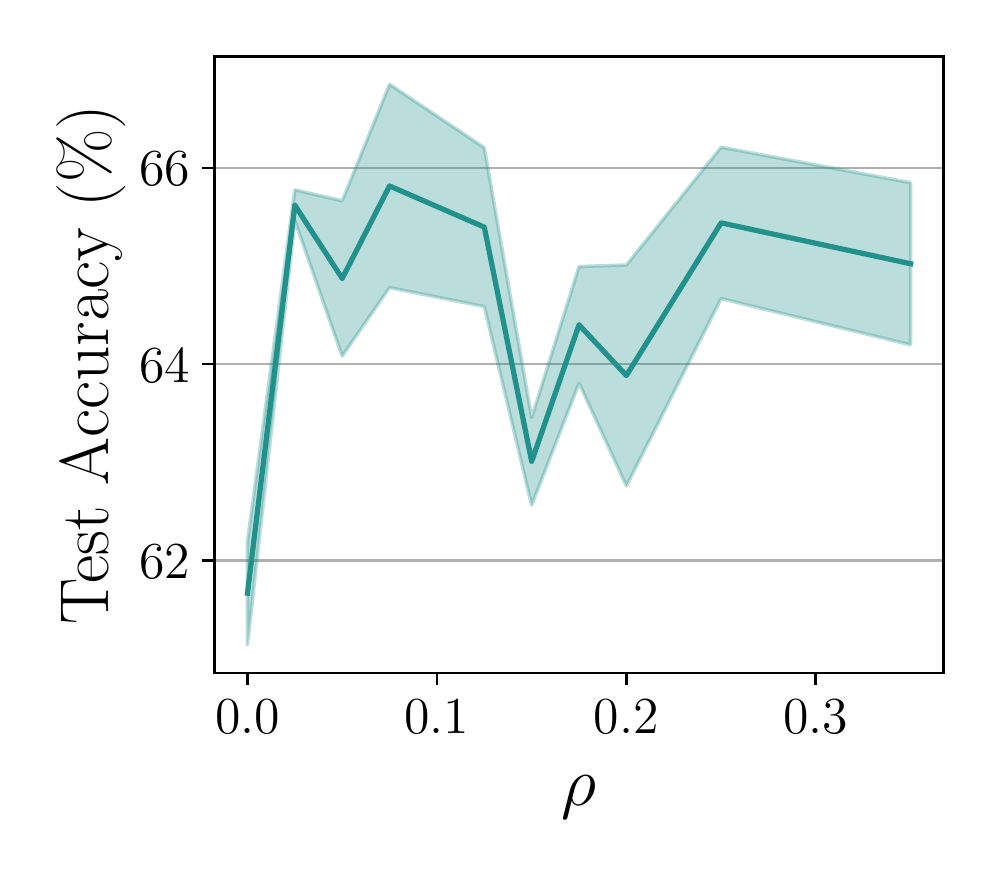}
    \vskip-0.1in
    \caption{Fraction = 5$\%$}
\end{subfigure}
\caption{The sensitivity analyses based on $\rho$ for fraction = 0.5$\%$ and 5$\%$ in CIFAR-10.}
\label{fig:sensitivity}
\end{figure}

\textbf{Additional Results} We demonstrate the efficacy of LCMat-S over the baselines from the experiments of 1) Selection with different network architecture (VGG-16); and 2) Selection on a different dataset (TinyImageNet). Table \ref{tab:additional_table} shows that LCMat is consistently competitive over the selected baselines on the evaluated settings.

\textbf{Ablation Study}
When we set $\rho=0$ in Eq \eqref{our_implementation}, our method is reduced to the gradient matching with the facility location algorithm. To validate the efficacy of loss-curvature matching over the gradient matching, we provide the ablation study of LCMat-S by conducting sensitivity analyses over $\rho$. Figure \ref{fig:sensitivity} shows that the test performances when $\rho > 0$ are consistently higher than when $\rho=0$, which shows the efficacy of loss-curvature matching over the gradient matching.

\subsection{Dataset Condensation Evaluation}
\textbf{Experiment Details} The condensation evaluation scenario is very similar to our selection scenario explained in Section \ref{coreset_evaluation}. The only difference is the existence of alternative training between $S$ and $\theta_{T}$ during the condensation, which arises from the nature of condensation.

We condense $S$ based on CIFAR-10 and CIFAR-100 with the utilization of ConvNet-3 as $f_{\theta_{T}}$. As specified in Eq \eqref{condense_objective}, we optimize $\theta_{T}$ from $T$ than the current $S$ during the alternative training of $S$ and $\theta_{T}$, which is shown to be effective for the condensation \citep{EDC}. It also fits with our parameter coverage on the Theorem \ref{theorem_v1}. All methods utilize the Differential Siamese Augmentation \citep{DSA} and the additional augmentation strategy specified in \cite{EDC} to further improve the performance. During the alternative update of $S$ and $\theta_{T}$, we re-initialize $\theta_{T}$ periodically as a common practice \citep{DC, EDC}. All experiments in this section are repeated over 3 times.

\begin{table}[h!]
\centering
\caption{Condensation performances on CIFAR-10 and CIFAR-100 with ConvNet-3. \textbf{Bold} represents best result. $\dagger$ means reported results from the original papers.}
\resizebox{0.48\textwidth}{!}
{\begin{tabular}{c cc cc}
\toprule & \multicolumn{2}{c}{CIFAR-10} & \multicolumn{2}{c}{CIFAR-100} \\ \cmidrule(lr){2-3} \cmidrule(lr){4-5} 
Fraction & 0.2$\%$  & 1$\%$ & 2$\%$ & 10$\%$\\ \cmidrule(lr){1-1} \cmidrule(lr){2-3} \cmidrule(lr){4-5} 
Random   & 37.13\scriptsize{$\pm$0.3}& 56.67\scriptsize{$\pm$0.5}& 20.60\scriptsize{$\pm$0.3} & 40.90\scriptsize{$\pm$0.0} \\
KIP & {${\text{47.30}}^{\dagger}$\scriptsize{$\pm$0.3}} & {${\text{50.10}}^{\dagger}$\scriptsize{$\pm$0.2}} & {${\text{13.40}}^{\dagger}$\scriptsize{$\pm$0.2}} & - \\
DM& 54.47\scriptsize{$\pm$0.5}& 65.23\scriptsize{$\pm$0.2}& 33.99\scriptsize{$\pm$0.2} & 43.35\scriptsize{$\pm$0.2} \\
DSA& 54.90\scriptsize{$\pm$0.3}& 61.90\scriptsize{$\pm$0.4}& 33.75\scriptsize{$\pm$0.1} & 38.71\scriptsize{$\pm$0.3} \\
\rowcolor{Gray} LCMat-C& \textbf{56.83}\scriptsize{$\pm$0.2} & \textbf{65.90}\scriptsize{$\pm$0.4} & \textbf{36.47}\scriptsize{$\pm$0.0}& \textbf{43.53}\scriptsize{$\pm$0.1}\\ \cmidrule(lr){1-1} \cmidrule(lr){2-3} \cmidrule(lr){4-5}
Full& \multicolumn{2}{c}{89.77\scriptsize{$\pm$0.2}} & \multicolumn{2}{c}{{65.13\scriptsize{$\pm$0.5}}} \\\bottomrule
\end{tabular}
}
\label{tab:cifar10-initial}
\end{table}

\textbf{Baselines} To validate the efficacy of LCMat-C, we compare the test performances over the baselines with different objectives. Baselines include the gradient matching (DSA) \citep{DSA}, feature output matching (DM) \citep{DM}, and kernel-based (KIP) methods \citep{kip}.

\textbf{Implementation of LCMat-C}
The gradient variance, $\text{Var}(\mathbf{G}^{T}_{\theta_{k}})$, in Eq \eqref{condense_objective}, requires the costly computation of per-sample gradients over $\theta$. We utilize BackPACK \citep{backpack}, which provides the computation of per-sample gradients at almost no time overhead. Also, we compute the gradient variance term only for the last layer, which is an efficient practice to improve the test performance with low computational costs. 

\textbf{Results}
Table \ref{tab:cifar10-initial} shows that condensed $S$ from LCMat-C consistently improves the test performances of all baselines over different fractions of CIFAR-10 and CIFAR-100. We especially observe significant improvements from the experiments on the low fraction budgets. We also test the robustness of LCMat-C on the cross-architecture scenario, which utilizes ResNet-10 \citep{resnet} and DenseNet-121 \citep{densenet} as testing backbones. Table \ref{tab:dc_cross} shows the consistent improvements of LCMat-C over baselines.

\begin{table}[h!]
  \centering
  \caption{Cross-architecture generalization performance ($\%$) on CIFAR-10 with ConvNet-3. \textbf{Bold} means best.}
\resizebox{0.48\textwidth}{!}
{\begin{tabular}{c|c|ccc}
\toprule
Fraction &Test Model & ConvNet-3 & ResNet-10 & DenseNet-121 \\ 
\midrule
\multirow{4}{*}{0.2$\%$}&Random&37.13\scriptsize{$\pm$0.3}&35.27\scriptsize{$\pm$0.4}&36.93\scriptsize{$\pm$0.6} \\ 
& DM &54.47\scriptsize{$\pm$0.5}&44.73\scriptsize{$\pm$1.1}&44.97\scriptsize{$\pm$0.3}\\
& DSA &54.90\scriptsize{$\pm$0.3}&46.03\scriptsize{$\pm$0.3}&45.63\scriptsize{$\pm$1.8}\\
& LCMat-C &\textbf{56.83}\scriptsize{$\pm$0.2}&\textbf{48.00}\scriptsize{$\pm$1.5}&\textbf{47.27}\scriptsize{$\pm$1.1} \\
\midrule
\multirow{4}{*}{1$\%$}&Random&56.67\scriptsize{$\pm$0.5}&53.57\scriptsize{$\pm$0.4}&56.77\scriptsize{$\pm$0.4} \\ 
& DM &65.23\scriptsize{$\pm$0.2}&56.77\scriptsize{$\pm$0.1}&55.80\scriptsize{$\pm$0.4}\\
& DSA &61.90\scriptsize{$\pm$0.4}&57.97\scriptsize{$\pm$0.2}&55.00\scriptsize{$\pm$0.8}\\
&LCMat-C &\textbf{65.90}\scriptsize{$\pm$0.4}&\textbf{60.93}\scriptsize{$\pm$0.4}&\textbf{57.93}\scriptsize{$\pm$0.1} \\ \midrule
100$\%$&Full&89.72\scriptsize{$\pm$0.2}&93.80\scriptsize{$\pm$0.3}&96.17\scriptsize{$\pm$0.2}\\
\bottomrule
\end{tabular}}
\label{tab:dc_cross}
\end{table}

\subsection{Application : Continual Learning with Memory Replay}
\begin{figure}[h!]
\centering
\begin{subfigure}[b]{0.49\columnwidth}
    \centering
    \includegraphics[width=0.99\columnwidth]{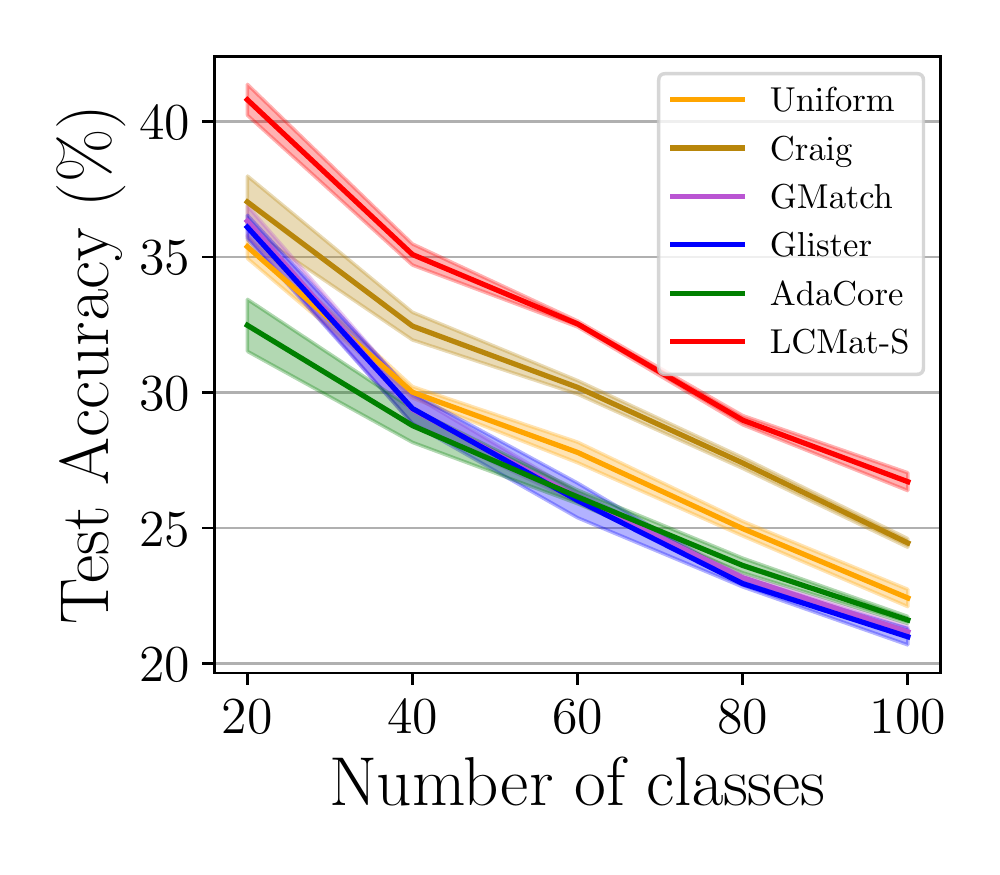}
    \caption{Selection Methods}
\end{subfigure}\hfill
\begin{subfigure}[b]{0.49\columnwidth}
    \centering
    \includegraphics[width=0.99\columnwidth]{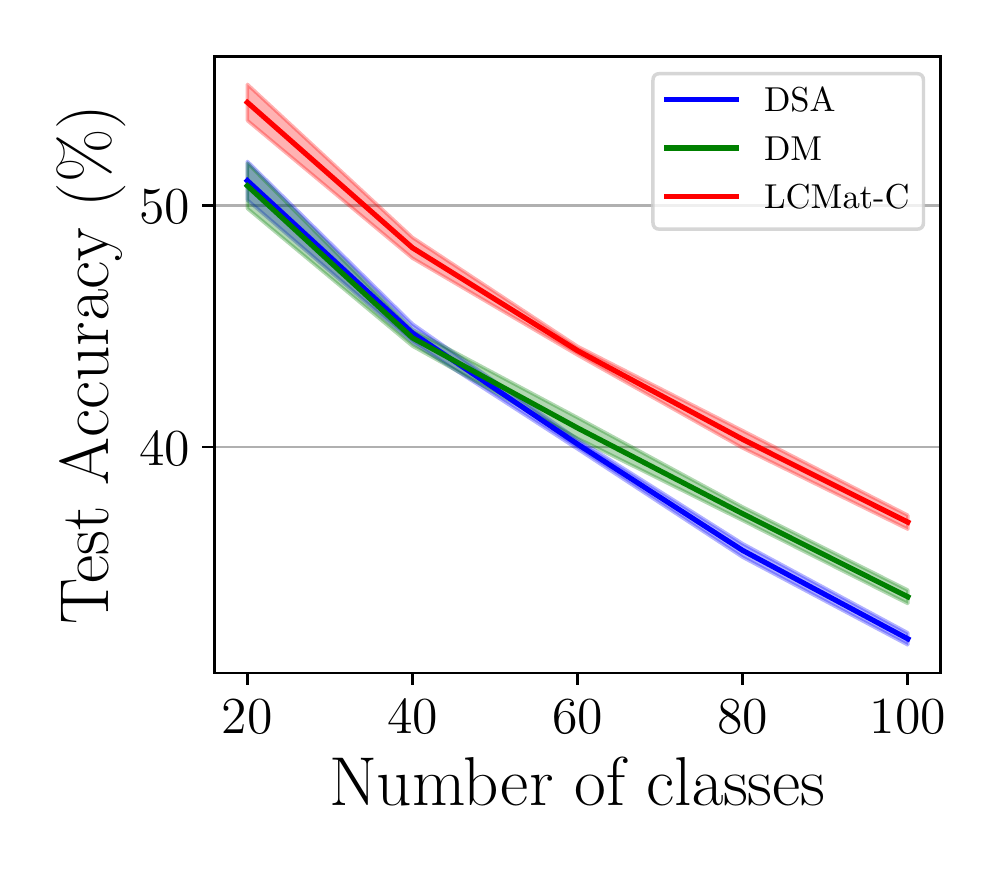}
    \caption{Condensation Methods}
\end{subfigure}
\caption{Test accuracy from the continual learning scenario with the selected or condensed data from CIFAR-100. We compare the methods from each category separately.} \label{fig:continual}
\end{figure}

Methods for memory-based continual learning store small representative instances; and these methods optimize its classifier with the samples stored in the memory to alleviate the catastrophic forgetting of previously observed tasks \citep{memory}. 
As an application practice, we utilize $S$ from each method as a memory exemplar for previously seen classes under the class incremental setting of \cite{DC,DM}. From the setting, CIFAR-100 is divided into 5 sets of sub-classes with a memory budget of 10 images per class, where each set of classes means a separate task stage. This setting purely trains a model based on the latest memory at each task stage. Figure \ref{fig:continual} shows that the variants of LCMat, LCMat-S and LCMat-C, significantly improve the test performances under the defined setting, which represents the minimization of catastrophic forgetting. 
\section{CONCLUSION}
We propose a new objective for dataset reduction named Loss-Curvature Matching, or LCMat. LCMat identifies the worst loss-curvature gap between the original dataset and the reduced dataset around the local parameter region, which is closely related to the parameter-based generalization on dataset reduction procedure. From the adaptive application of LCMat, such as selection-based methods and condensation-based methods; LCMat consistently shows improved performances over baselines from both lines of research in dataset reduction. Especially, LCMat shows clear performance merits on the extreme reduction ratio, which is a specialized property for on-device learning where the memory capacity is limited.


\subsubsection*{Acknowledgements}
This research was supported by AI Technology Development for Commonsense Extraction, Reasoning, and Inference from Heterogeneous Data (IITP)
funded by the Ministry of Science and ICT(2022-0-00077). Also, authors would like to acknowledge Dongjun Kim and Byeonghu Na for their invaluable discussions and supports.

\bibliography{bib}





\appendix
\onecolumn

\section{Proofs}
\subsection{Proof of Proposition 1}
\label{appendix:proposition1 proof}
\begin{proposition}
\label{proposition_appen}
When $\mathbb{H}_D = \nabla^{2}_{\theta}\mathcal{L}(D;\theta)$ is a Hessian matrix of $\mathcal{L}(D;\theta)$, let $\mathbb{H}_{T,S} = \mathbb{H}_{T}\!-\! \mathbb{H}_{S}=\!\nabla^{2}_{\theta}\mathcal{L}(T;\theta)\! - \! \nabla^{2}_{\theta}\mathcal{L}(S;\theta)$ and $\lambda^{T,S}_{1}$ be the maximum eigenvalue of the matrix $\mathbb{H}_{T,S}$, then we have: 
\begin{align}
    \underset{{||\epsilon||_{2}} \leq \rho }{\max}\frac{\mathcal{L}_{abs}(T,S;\theta\!+\!\epsilon) \!\!-\!\! {\mathcal{L}_{abs}(T,S;\theta)}}{\rho} \leq\underbrace{{\Big{\|}\nabla_{\theta}\mathcal{L}(T;\theta) - \nabla_{\theta}\mathcal{L}(S;\theta) \Big{\|}}_2}_{\text{Gradient Matching via $L_{2}$-norm.}}+ \underbrace{\frac{1}{2}\rho\lambda^{T,S}_{1}}_{\text{Max eigen}}+ \underset{{||\upsilon||_{2}} \leq 1 }{\max}O(\rho^{2}\upsilon^3) \nonumber
\end{align}
\end{proposition} 
\textit{Proof.}
By leveraging the Taylor-series with finite-order approximation, we can derive $\underset{{||\epsilon||_{2}} \leq \rho }{\max}\mathcal{L}_{abs}(T,S;\theta\!+\!\epsilon)$, which is an abbreviated term introduced in Section 3 of main paper, as follows: 
\begin{align}
&\underset{{||\epsilon||_{2}} \leq \rho }{\max}\mathcal{L}_{abs}(T,S;\theta\!+\!\epsilon) =\max_{\|\epsilon\|_{2}\leq \rho}\Big{|} \mathcal{L}(T;\theta+\epsilon)-\mathcal{L}(S;\theta+\epsilon)\Big{|} \nonumber\\
&=\max_{\|\epsilon\|_{2}\leq \rho}\Big{|} 
\mathcal{L}(T;\theta)+\epsilon^{\top}\nabla_{\theta}\mathcal{L}(T;\theta)+\frac{1}{2}\epsilon^{\top}\nabla^{2}_{\theta}\mathcal{L}(T;\theta)\epsilon-\mathcal{L}(S;\theta)-\epsilon^{\top}\nabla_{\theta}\mathcal{L}(S;\theta)-\frac{1}{2}\epsilon^{\top}\nabla^{2}_{\theta}\mathcal{L}(S;\theta)\epsilon)+O(\epsilon^3)\Big{|}
\\&\leq|\mathcal{L}(T;\theta)-\mathcal{L}(S;\theta)|+\max_{\|\epsilon\|_{2}\leq\rho}|\epsilon^{\top}(\nabla_{\theta}\mathcal{L}(T;\theta)-\nabla_{\theta}\mathcal{L}(S;\theta))| +\max_{\|\epsilon\|_{2}\leq\rho}|\frac{1}{2}\epsilon^{\top}\nabla^{2}_{\theta}\mathcal{L}(T;\theta)\epsilon - \frac{1}{2}\epsilon^{\top}\nabla^{2}_{\theta}\mathcal{L}(S;\theta)\epsilon|
\\ \nonumber
&+ \max_{\|\upsilon\|_{2}\leq \rho}O(\upsilon^3)
\\&=|\mathcal{L}(T;\theta)-\mathcal{L}(S;\theta)|+\rho\|\nabla_{\theta}\mathcal{L}(T;\theta)-\nabla_{\theta}\mathcal{L}(S;\theta)\|^{2} +\max_{\|\epsilon\|_{2}\leq \rho}|\frac{1}{2}\epsilon^{\top}\nabla^{2}_{\theta}\mathcal{L}(T;\theta)\epsilon - \frac{1}{2}\epsilon^{\top}\nabla^{2}_{\theta}\mathcal{L}(S;\theta)\epsilon| + \max_{\|\upsilon\|_{2}\leq 1}O(\rho^{3}\upsilon^3) \label{equation_4}
\end{align}
From here, we denote the difference of hessian, $\mathbb{H}_{T,S} = \nabla^{2}_{\theta}l(T;\theta) - \nabla^{2}_{\theta}l(S;\theta)$, and we derive $\mathbb{H}_{T,S}$ as follows:
\begin{align}
    &\max_{\|\epsilon\|_{2}\leq \rho}|\frac{1}{2}\epsilon^{\top}\nabla^{2}_{\theta}l(T;\theta)\epsilon - \frac{1}{2}\epsilon^{\top}\nabla^{2}_{\theta}l(S;\theta)\epsilon|=\max_{\|\epsilon\|_{2}\leq \rho}|\frac{1}{2}\epsilon^{\top}\mathbb{H}_{T,S}\epsilon| =\max_{\|\epsilon\|_{2}\leq \rho}\frac{1}{2}\|\epsilon\|\|\mathbb{H}_{T,S}\epsilon\| \nonumber \\
    &=\frac{1}{2}\rho^{2}\max_{\|\nu\|_{2}\leq 1}\|\mathbb{H}_{T,S}\nu\|=\frac{1}{2}\rho^{2}\lambda^{T,S}_{1}
\label{equation_7}
\end{align}
Here, $\lambda^{T,S}_{i}$ is maximum eigenvalue of the matrix $\mathbb{H}_{T,S}$. By replacing $\lambda^{T,S}_{i}$ into the Eq \eqref{equation_4}, $\underset{{||\epsilon||_{2}} \leq \rho }{\max}\mathcal{L}_{abs}(T,S;\theta\!+\!\epsilon)$ is derived as follows:
\begin{align}
&\underset{{||\epsilon||_{2}} \leq \rho }{\max}\mathcal{L}_{abs}(T,S;\theta\!+\!\epsilon) =\max_{\|\epsilon\|_{2}\leq \rho}\Big{|} \mathcal{L}(T;\theta+\epsilon)-\mathcal{L}(S;\theta+\epsilon)\Big{|} \nonumber \\
&\leq\mathcal{L}_{abs}(T,S;\theta)+\rho\|\nabla_{\theta}\mathcal{L}(T;\theta)-\nabla_{\theta}\mathcal{L}(S;\theta)\|^{2} +\frac{1}{2}\rho^{2}\lambda^{T,S}_{1} + \max_{\|\upsilon\|_{2}\leq 1}O(\rho^{3}\upsilon^3) \label{equation_5}
\end{align}
After moving $\mathcal{L}_{abs}(T,S;\theta)$ to L.H.S, dividing both sides by $\rho > 0$ finishes the proof as follows:
\begin{align}
&\frac{\underset{{||\epsilon||_{2}} \leq \rho }{\max}\mathcal{L}_{abs}(T,S;\theta\!+\!\epsilon)-\mathcal{L}_{abs}(T,S;\theta)}{\rho} 
\leq\|\nabla_{\theta}\mathcal{L}(T;\theta)-\nabla_{\theta}\mathcal{L}(S;\theta)\|^{2} +\frac{1}{2}\rho\lambda^{T,S}_{1} + \max_{\|\upsilon\|_{2}\leq 1}O(\rho^{2}\upsilon^3) \label{equation_6}
\end{align}

\subsection{Proof of Eq \eqref{main_objective}}
\label{appendix:equation(7)}
From the replacement of $\mathbb{H}_T$ and $\mathbb{H}_S$; into $\hat{\mathbb{H}}_T$ and $\hat{\mathbb{H}}_S$, which are diagonal version of Hessian matrices of $\mathbb{H}_T$ and $\mathbb{H}_S$, respectively, we could further simplify the derivation in Eq \eqref{equation_7} of the supplementary material into following equations:
\begin{align}
    &\max_{\|\epsilon\|_{2}\leq \rho}\Big{|}\frac{1}{2}\epsilon^{T}\nabla^{2}_{\theta}\mathcal{L}(T;\theta)\epsilon - \frac{1}{2}\epsilon^{T}\nabla^{2}_{\theta}\mathcal{L}(S;\theta)\epsilon\Big{|} \nonumber \\
    &=\frac{1}{2}\max_{\|\epsilon\|_{2}\leq \rho}\Big{|}\epsilon^{T}(\hat{\mathbb{H}}_T-\hat{\mathbb{H}}_S)\epsilon\Big{|}=\frac{1}{2}\rho\max_{\|\epsilon\|_{2}\leq \rho}\Big{|}
    \sum_{k}\epsilon_{k}(\hat{\lambda}^{T}_{k}-\hat{\lambda}^{S}_{k})\Big{|} =\frac{1}{2}\rho^{2}\max_{k}\Big{|}\hat{\lambda}^{T}_{k}-\hat{\lambda}^{S}_{k}\Big{|} \nonumber
\end{align}
Here, $\hat{\lambda}^{T}_{k}$ and $\hat{\lambda}^{S}_{k}$ are eigenvalues of $\hat{\mathbb{H}}_{T}$ and $\hat{\mathbb{H}}_{S}$ on $k$-th parameter dimension for $\theta$.
\subsection{Proof of Eq \eqref{upper_bound}}
\label{appdendix:equation(10)}
In this section, we prove that Eq \eqref{upper_bound} with per-sample weight $\gamma_{j}$ could be changed as follows:
 \begin{align}
 \label{intro_upperbound_appen}
     \underset{S \subseteq T }{\min}\Big{\|}\sum_{i \in T}\hat{\lambda}^{T}_{i,\mathcal{K}}
     -\sum_{j \in S}\gamma_{j}\hat{\lambda}^{S}_{j,\mathcal{K}}\Big{\|} +
     \Big{\|}\sum_{i \in T}\mathbf{g}^{T}_{i}
     -\sum_{j \in S}\gamma_{j}\mathbf{g}^{S}_{j}\Big{\|} \leq 
     \sum_{i \in T}
     \underset{j \in S}{\min}
     \Big{\|}\mathbf{g}^{T}_{i} - \mathbf{g}^{S}_{j}
     \Big{\|}+\Big{\|}\hat{\lambda}^{T}_{i,\mathcal{K}} -
     \hat{\lambda}^{S}_{j,\mathcal{K}}\Big{\|}
\end{align}

To derive the Eq \eqref{intro_upperbound_appen}, we first re-phrase the upper-bound derivation of Craig \citep{Craig} with the notation based on our paper. \cite{Craig} showed that the norm-based error between sum of whole elements in $T$ and a weighted sum of a subset of elements is upper-bounded by facility location objective. For the complete proof of Eq (10) in main paper, we also follow the proof of \cite{Craig} here. We assume that there is a mapping function $\zeta_{\theta}(i) :  T \rightarrow S$ which assigns every data point $i \in T$ to one of the elements $j \in S$, i.e. $\zeta_{\theta}(i)=j\in S$. Corresponding set $C_{j}=\{i \in [n]| \zeta(i)=j\} \subseteq T$ is defined as a set of data points that are assigned to $j \in S$, and $\gamma_{j} = |C_{j}|$ be the number of samples assigned to $j$. From this derivation, we can write as follows:
 \begin{align}
     \sum_{i \in T}\mathbf{g}^{T}_{i}
     =\sum_{i \in T}\Big{(}\mathbf{g}^{T}_{i}-\mathbf{g}^{T}_{\zeta_{\theta}(i)}+\mathbf{g}^{T}_{\zeta_{\theta}(i)}\Big{)} = \sum_{i \in T}\Big{(}\mathbf{g}^{T}_{i}-\mathbf{g}^{T}_{\zeta_{\theta}(i)}\Big{)}+\sum_{j \in S}\gamma_{j}\mathbf{g}^{S}_{j}
\end{align}
From above equation, subtracting $\sum_{j \in S}\gamma_{j}\mathbf{g}^{S}_{j}$ and taking norm with triangle inequality, we get the upper bound as follows:
 \begin{align}
 \label{grad_inequal_appen}
     \Big{\|}\sum_{i \in T}\mathbf{g}^{T}_{i}
     -\sum_{j \in S}\gamma_{j}\mathbf{g}^{S}_{j}\Big{\|} \leq 
     \sum_{i \in T}\Big{\|}\mathbf{g}^{T}_{i} - \mathbf{g}^{T}_{\zeta_{\theta}(i)}
     \Big{\|}
\end{align}
To construct the upper bound based on $\hat{\lambda}^{T}_{i,k}$ and $\hat{\lambda}^{S}_{j,k}$, we first denote a vector $\hat{\mathbf{\lambda}}^{T}_{i,\mathcal{K}}=(\hat{\lambda}^{T}_{i,a},\hat{\lambda}^{T}_{i,b},\hat{\lambda}^{T}_{i,c},...)$, where $a,b,c...$ are naively introduced indices for a specific index $k \in \mathcal{K}$. Then for $i,j \in T$, following holds by definition.
\begin{align}
\sum_{k \in \mathcal{K}}\Big{|}\hat{\lambda}^{T}_{i,k} -  \hat{\lambda}^{S}_{j,k}\Big{|} =\Big{\|}\hat{\mathbf{\lambda}}^{T}_{i,\mathcal{K}} - \hat{\mathbf{\lambda}}^{S}_{i,\mathcal{K}}\Big{\|}_{1}
\end{align}
As Eq \eqref{grad_inequal_appen} holds for any bounded vector, we could also extend the result of Eq \eqref{grad_inequal_appen} as follows:
 \begin{align}
 \label{lambda_inequal_appen}
     \Big{\|}\sum_{i \in T}\hat{\lambda}^{T}_{i,\mathcal{K}}
     -\sum_{j \in S}\gamma_{j}\hat{\lambda}^{S}_{j,\mathcal{K}}\Big{\|} \leq
     \sum_{i \in T}\Big{\|}\hat{\lambda}^{T}_{i,\mathcal{K}} -
     \hat{\lambda}^{T}_{\zeta_{\theta}(i),\mathcal{K}}\Big{\|}
\end{align}
Integrating Eq \eqref{grad_inequal_appen} and \eqref{lambda_inequal_appen},
 \begin{align}
 \label{whole_inequal_appen}
     \Big{\|}\sum_{i \in T}\hat{\lambda}^{T}_{i,\mathcal{K}}
     -\sum_{j \in S}\gamma_{j}\hat{\lambda}^{S}_{j,\mathcal{K}}\Big{\|} +
     \Big{\|}\sum_{i \in T}\mathbf{g}^{T}_{i}
     -\sum_{j \in S}\gamma_{j}\mathbf{g}^{S}_{j}\Big{\|} \leq 
     \sum_{i \in T}\Big{\|}\mathbf{g}^{T}_{i} - \mathbf{g}^{T}_{\zeta_{\theta}(i)}
     \Big{\|}+\Big{\|}\hat{\lambda}^{T}_{i,\mathcal{K}} -
     \hat{\lambda}^{T}_{\zeta_{\theta}(i),\mathcal{K}}\Big{\|}
\end{align}
Then, we set the mapping function $\zeta_{\theta}(i) = \text{argmin}_{j \in S}\Big{\|}\mathbf{g}^{T}_{i} - \mathbf{g}^{S}_{j}
     \Big{\|}+\Big{\|}\hat{\lambda}^{T}_{i,\mathcal{K}} -
     \hat{\lambda}^{S}_{j,\mathcal{K}}\Big{\|}$. Hence,

 \begin{align}
 \label{min_inequal_appen}
     \underset{S \subseteq V }{\min}\Big{\|}\sum_{i \in T}\hat{\lambda}^{T}_{i,\mathcal{K}}
     -\sum_{j \in S}\gamma_{j}\hat{\lambda}^{S}_{j,\mathcal{K}}\Big{\|} +
     \Big{\|}\sum_{i \in T}\mathbf{g}^{T}_{i}
     -\sum_{j \in S}\gamma_{j}\mathbf{g}^{S}_{j}\Big{\|} \leq 
     \sum_{i \in T}
     \underset{j \in S}{\min}
     \Big{\|}\mathbf{g}^{T}_{i} - \mathbf{g}^{S}_{j}
     \Big{\|}+\Big{\|}\hat{\lambda}^{T}_{i,\mathcal{K}} -
     \hat{\lambda}^{S}_{j,\mathcal{K}}\Big{\|}
\end{align}
Based on the upper bound, we provide the pseudo-code, which is the greedy algorithm of LCMat-S, in Algorithm \ref{alg:greedy}. The provided pseudo-code is motivated from \cite{adacore}. The notations are all defined from the main paper. As mentioned in the main paper, we report the performance of LCMat-S without the application of weighting procedure because the performance with weighting shows degraded performance than the performance without weights. It should be noted that the greedy algorithm only with the incremental selection procedure provides us a logarithmic approximation \citep{facilitylocation_greedy_1,facilitylocation_greedy_2}.

\begin{algorithm}
\caption{The Greedy Algorithm of LCMat-S}\label{alg:greedy}
\begin{algorithmic}[1]
\Ensure Subset $S \subseteq V$ with corresponding per-element stepsizes $\{\gamma\}_{j\in S}$.
\Procedure{LCMat-S}{}
\State $S_0 \leftarrow \emptyset, i=0$
\While{$F(S) < C_1  - \epsilon$}\Comment{Selection Procedure of LCMat-S}
\State $j\in {\arg\max}_{e \in V\setminus S_{i-1}} F (e|S_{i-1})$
\State $S_i = S_{i-1}\cup \{j\}$
\State $i = i + 1$
\EndWhile
\For{$j=1$ to $|S|$}\Comment{Weighting Procedure of LCMat-S (Optional)}
\State{$\gamma_j = \sum_{i\in V} \mathbb{I} \left[ j = {\arg\min}_{j \in S}  \Big{\|}\mathbf{g}^{T}_{i} - \mathbf{g}^{S}_{j}
\Big{\|}+\Big{\|}\hat{\lambda}^{T}_{i,\mathcal{K}} -
\hat{\lambda}^{S}_{j,\mathcal{K}}\Big{\|}  \right]$}    
\EndFor
\EndProcedure
\end{algorithmic}
\end{algorithm}

\newpage
\subsection{Proof of Theorem \ref{theorem_v1}}
\label{appendix:theorem1}
First, we define a set of $\theta$, $\hat{\Theta}$, which is the application range of generalization bound as follows:
\begin{definition}
    $\hat{\Theta} = \{\theta : \mathcal{L}(T;\theta) \leq \mathcal{L}(\mathbb{D};\theta) \,\text{ for }\, \theta \in \Theta\}$
\end{definition}
As noted in the main paper, $\mathcal{L}(T;\theta)$ and $\mathcal{L}(\mathbb{D};\theta)$ are approximated by the training loss and test loss from the experimental practices, respectively. $\hat{\Theta}$ specifies $\theta$ whose generalization gap is more than equal to zero, which is intuitive when we optimize $\theta$ based on $T$ based on the valid setting.
\begin{theorem}
\label{theorem_v1_appen}
(Generalization Bound of $\underset{{||\epsilon||_{2}} \leq \rho }{\max}\mathcal{L}_{abs}(T,S;\theta\!+\!\epsilon)$)
For $\theta \in \hat{\Theta}$, with probability at least $1-\delta$ over the choice of the training set $T$ with $|T|=n$, the following holds. 
\begin{align}
\label{theorem_eq_appen}
&\mathbb{E}_{\epsilon \sim \mathcal{N}(0,\rho)}[\Big{|}\mathcal{L}(\mathbb{D};\theta\!+\!\epsilon)-\mathcal{L}(S;\theta\!+\!\epsilon)\Big{|}] \leq \max_{\|\epsilon\|_{2}\leq \rho}|\mathcal{L}(T;\theta\!+\!\epsilon)-\mathcal{L}(S;\theta\!+\!\epsilon)| + \sqrt{\frac{O(k+\log\frac{n}{\delta})}{n-1}}
\nonumber
\end{align}
\end{theorem}
\textit{Proof.} 
We start the proof by utilizing the triangle inequality when each metric is provided by absolute difference as follows:
\begin{align}
    |x-z| \leq |x-y| + |y-z|\text{ for all }\text{ }x,y,z
\end{align}
From the triangle inequality, we can derive the inequality between the losses from the different population as follows:
\begin{align}
    |\mathcal{L}(\mathbb{D};\theta)-\mathcal{L}(S;\theta)| \leq |\mathcal{L}(\mathbb{D};\theta)-\mathcal{L}(T;\theta)| + |\mathcal{L}(T;\theta)-\mathcal{L}(S;\theta)| \text{ for all } \theta \in \Theta
\end{align}
It can also be extended into the following inequality, which is inequality between the expected loss on the $\epsilon$-perturbed region of $\theta$:
\begin{align}
\label{ineq1}
    &\mathbb{E}_{\epsilon \sim \mathcal{N}(0,\rho)}\big{[}|\mathcal{L}(\mathbb{D};\theta+\epsilon)-\mathcal{L}(S;\theta+\epsilon)|\big{]} \leq \mathbb{E}_{\epsilon \sim \mathcal{N}(0,\rho)}\left[|\mathcal{L}(\mathbb{D};\theta+\epsilon)-\mathcal{L}(T;\theta+\epsilon)|\right] \nonumber \\ & + \mathbb{E}_{\epsilon \sim \mathcal{N}(0,\rho)}\left[|\mathcal{L}(T;\theta+\epsilon)-\mathcal{L}(S;\theta+\epsilon)|\right] \text{ for all } \theta \in \Theta
\end{align}

Here, we refer the PAC-Bayes theorem \cite{pac_v1} to derive the bound between them. It should be noted that the provided proof referred the proof concept of SAM \cite{SAM} and fisher-SAM \citep{kim2022fisher}. The PAC-Bayes generalization bound of \cite{pac_v1,pac_v2} provides that, for any prior distribution $P(\theta)$ with probability at least 1-$\delta$ over the choice of the training set $T$ with $|T|=n$, it holds that
\begin{align}
\label{original_kl}
    \mathbb{E}_{Q(\theta)}\left[\mathcal{L}({\mathbb{D}};\theta)\right]\leq \mathbb{E}_{Q(\theta)}\big{[}\mathcal{L}({T};\theta)\big{]}+\sqrt{\frac{\text{KL}(Q(\theta)||P(\theta))+\log{\frac{n}{\delta}}}{2(n-1)}}
\end{align}
Posterior distribution, $Q(\theta)$, is assumed to be dependent on the training dataset $T$ and synthetic data variable $S$, which are both accessible during the training procedure of $\theta$. Let $k$ be the dimensionality of the model parameter $\theta$. Following \cite{kim2022fisher}, if we assume that $Q(\theta)=\mathcal{N}(\mu_{Q},\sigma^{2}_{Q}I)$ and $P(\theta)=\mathcal{N}(\mu_{P},\sigma^{2}_{P}I)$, the KL divergence can be written as follows:
\begin{align}
    \text{KL}(Q||P)=\frac{1}{2}\left[\frac{k\sigma^{2}_{Q}+\| \mu_{Q} - \mu_{P}\| ^{2}_{2} }{\sigma^{2}_{P}}-k+k\log(\frac{\sigma^{2}_{P}}{\sigma^{2}_{Q}})\right]
\end{align}
It should be noted that the prior distribution $P(\theta)$ do not have access into the training dataset $T$, which makes it hard to adapt the $P(\theta)$ to minimize KL-divergence with the corresponding posterior $Q(\theta)$. It inspires the utilization of the covering approach from \cite{SAM,intriguing}, which introduces a pre-defined set of parameter distributions with the constraint that each prior distribution holds the PAC-Bayes bound. Afterwards, we can select the one from the set which has minimal KL-divergence in junction with the posterior $Q(\theta)$.

From a pre-defined set of prior distributions $\{P_{j}(\theta)\}^{J}_{j=1}$ where $P_{j}(\theta) = \mathcal{N}(\bar{\theta_{j}},\bar{\sigma_{j}})$ and posterior distribution $Q(\theta)$, we set $\mu_{Q}=\theta$, $\sigma_{Q}=\rho$, and $\bar{\theta_{j}}=0$. Here, the point is how to set $\bar{\sigma_{j}}$. Motivated from \cite{intriguing}, we introduce $\{c \text{exp}((1-j)/k)|j\in\mathbb{N}\}$, which is a set of pre-defined parameter values for $\sigma^{2}_{P}$. For the detailed analyses about the inclusion of $c$, see \citet{intriguing} for the detailed explanation of the technique. From this setting, PAC-Bayes bound holds with probability $1-\delta_{j}$ when $\delta_{j}=\frac{6\delta}{\pi^{2}j^{2}}$, which is generalized by the union bound theorem that all bounds hold simultaneously with probability at least $1-\sum^{\infty}_{j=1}\frac{6\delta}{\pi^{2}j^{2}}=1-\delta$. 

With the specified $Q(\theta)$ and $P(\theta)$, we have:
\begin{align}
    k\sigma^{2}_{Q}+\| \mu_{Q} - \mu_{P}\| ^{2}_{2} = k\rho^{2}+\|\theta\|^{2}_{2} 
\end{align}
With the replacement of $Q(\theta)$ and $P(\theta)$, we rephrase Eq \eqref{original_kl} as follows:
\begin{align}
\label{second_kl}
    \mathbb{E}_{\epsilon \sim \mathcal{N}(0,\rho)}\big{[}\mathcal{L}_{\mathbb{D}}(\theta + \epsilon)\big{]}\leq \mathbb{E}_{\epsilon \sim \mathcal{N}(0,\rho)}\big{[}\mathcal{L}_{T}(\theta + \epsilon)\big{]}+\sqrt{\frac{\frac{1}{4}k(\frac{\rho^{2}+{\|\theta\|}^{2}_{2}/{k}}{\sigma^{2}_{P}}-1+\log{\frac{\sigma^{2}_{P}}{\rho^{2}}})+\log{\frac{n}{\delta}}}{n-1}}
\end{align}
Here, we first restrict the value range of $\sigma^{2}_{P}$ to further derive the bound of KL divergence. Afterwards, we provide that the specified value range is strictly feasible with some $j \in \mathbb{N}$. Having said that, we provide the range of $\sigma^{2}_{P}$ as follows:
\begin{align}
\label{sigma_bound}
    \rho^{2}+\|\theta\|^{2}_{2}/k \leq \sigma^{2}_{P} \leq \text{exp}(1/k)(\rho^{2}+\|\theta\|^{2}_{2}/k)
\end{align}
From the specified region of $\sigma_{P}$, KL divergence is bounded as follows:
\begin{align}
    KL(Q(\theta)||P(\theta))&=\frac{k}{2}(\frac{\rho^{2}+{\|\theta\|}^{2}_{2}/{k}}{\sigma^{2}_{P}}-1+\log{\frac{\sigma^{2}_{P}}{\rho^{2}}}) \\ &\leq \frac{k}{2}(\frac{\rho^{2}+{\|\theta\|}^{2}_{2}/{k}}{\rho^{2}+{\|\theta\|}^{2}_{2}/{k}}-1 + \log\big{(}\frac{\text{exp}(1/k)(\rho^{2}+\|\theta\|^{2}_{2}/k)}{\rho^{2}}\big{)}) \\ &=
    \frac{k}{2}(\frac{1}{k}+\log(1+\frac{\|\theta\|^{2}_{2}}{k\rho^{2}}))
\end{align}
It should be noted that above bound holds only for specific $j \in \mathbb{N}$, which is not specified yet. Utilizing the provided bound of KL-divergence for specific $j$ and substracting the first term in R.H.S of Eq \eqref{second_kl}, it is further derived as follows:
\begin{align}
\label{pac_v2}
    \mathbb{E}_{\epsilon \sim \mathcal{N}(0,\rho)}\big{[}\mathcal{L}_{\mathbb{D}}(\theta + \epsilon)-\mathcal{L}_{T}(\theta + \epsilon)\big{]}\leq \sqrt{\frac{\frac{1}{4}k\log{(1+\frac{\|\theta\|^{2}_{2}}{k\rho^{2}})}+\frac{1}{4}+\log{\frac{n}{\delta_{j}}}}{n-1}}
\end{align}
From the above bound, the value range of $\|\theta \|^{2}_{2}$ divided into a range in which the bound holds trivially and a range in which it does not. The right hand side of \eqref{pac_v2} is lower-bounded by $\sqrt{\frac{k}{4n}\log(1+\|\theta\|^{2}_{2}/\rho^{2})}$, which is greater than 1 when $\|\theta\|^{2}_{2} \geq \rho^{2}(\text{exp}(4n/k)-1)$. It gaurantees that the right hand side of \eqref{pac_v2} is greather than 1, which results in the trivial proof of inequality. Having said that, we focus on the case when $\|\theta\|^{2}_{2} \leq \rho^{2}(\text{exp}(4n/k)-1)$.

When $\|\theta\|^{2}_{2} \leq \rho^{2}(\text{exp}(4n/k)-1)$, we have:
\begin{align}
    \rho^{2}+\|\theta\|^{2}_{2}/k \leq \rho^{2}(1+\text{exp}(4n/k))
\end{align}
By considering the bound where $j = \left\lfloor 1-k\log((\rho^{2}+\|\theta\|^{2}_{2}/k)/c) \right\rfloor \in \mathcal{N}$ and setting $c = \rho^{2}(1+\text{exp}(4n/k))$, we can derive the feasible bound of $\sigma^{2}_{P}$ as follows:
\begin{align}
\label{sigma_bound_v2}
    \rho^{2}+\|\theta\|^{2}_{2}/k \leq \sigma^{2}_{P} \leq \text{exp}(1/k)(\rho^{2}+\|\theta\|^{2}_{2}/k)
\end{align}
It is exactly same with the provided range of $\sigma^{2}_{P}$ in Eq \eqref{sigma_bound}. The bound which corresponds to $j$ holds with probability $1-\delta_{j}$ for $\delta_{j}=\frac{6\delta}{\pi^{2}j^{2}}$. By leveraging it, we transform the log term $\log\frac{n}{\delta_{j}}$ as follows:
\begin{align}
    &\log\frac{n}{\delta_{j}} 
    = \log\frac{n}{\delta}+\log\frac{\pi^{2}j^{2}}{6} 
    \leq \log\frac{n}{\delta} + \log \frac{\pi^{2}k^{2}\log^{2}(\frac{c}{\rho^{2}+\|\theta\|^{2}_{2}/k})}{6} \\
    & \leq \log\frac{n}{\delta} + \log \frac{\pi^{2}k^{2}\log^{2}(\frac{c}{\rho^{2}})}{6} 
    \leq \log\frac{n}{\delta} + \log \frac{\pi^{2}k^{2}\log^{2}(1+\exp(4n/k))}{6} \\
    & \leq \log\frac{n}{\delta} + \log \frac{\pi^{2}k^{2}(2+4n/k)^{2}}{6} \leq \log\frac{n}{\delta} +2\log(6n+3k)
\end{align}
By replacing the log term and utilizing $\theta \in \hat{\Theta}$, the absolute difference is bounded as follows:
\begin{align}
\label{absolute_difference}
    &\mathbb{E}_{\epsilon \sim \mathcal{N}(0,\rho)}\big{[}\mathcal{L}(\mathbb{D};\theta + \epsilon)-\mathcal{L}(T;\theta + \epsilon)\big{]} \\
    &= \mathbb{E}_{\epsilon \sim \mathcal{N}(0,\rho)}\big{[}|\mathcal{L}(\mathbb{D};\theta + \epsilon)-\mathcal{L}(T;\theta + \epsilon)|\big{]}
    \leq \sqrt{\frac{\frac{1}{4}k\log{(1+\frac{\|\theta\|^{2}_{2}}{k\sigma^{2}})}+\frac{1}{4}+\log{\frac{n}{\delta}}+2\log(6n+3k)}{n-1}}
\end{align}
Utilizing the inequality in Eq \eqref{absolute_difference}, we replace the Eq \eqref{ineq1} as follows:
\begin{align} 
\label{before_final_objective}
    &\mathbb{E}_{\epsilon \sim \mathcal{N}(0,\rho)}\big{[}|\mathcal{L}(\mathbb{D};\theta+\epsilon)-\mathcal{L}(S;\theta+\epsilon)|\big{]} \nonumber \\
    &\leq\mathbb{E}_{\epsilon \sim \mathcal{N}(0,\rho)}\big{[}|\mathcal{L}(T;\theta+\epsilon)-\mathcal{L}(S;\theta+\epsilon)|\big{]}+ \sqrt{\frac{\frac{1}{4}k\log{(1+\frac{\|\theta\|^{2}_{2}}{k\sigma^{2}})}+\frac{1}{4}+\log{\frac{n}{\delta}}+2\log(6n+3k)}{n-1}}
\end{align}
Finally, we are to bound the expectation term in R.H.S with the $\max_{\|\epsilon\|_{2}\leq \rho}|\mathcal{L}(T;\theta\!+\!\epsilon)-\mathcal{L}(S;\theta\!+\!\epsilon)|$ by utilizing the results from \cite{laurent2000adaptive} as follows:
\begin{align}
    z \sim \mathcal{N}(0,\gamma I) \rightarrow \|z\|^{2}\leq k\gamma{\Big{(} 1+\sqrt{\frac{\log{n}}{k}}\Big{)}}^{2} \text{ with probability at least } 1-\frac{1}{\sqrt{n}}
\end{align}
Here we denote $\rho = k\gamma{\Big{(} 1+\sqrt{\frac{\log{n}}{k}}\Big{)}}^{2}$. To provide the upper-bound of $\mathbb{E}_{\epsilon \sim \mathcal{N}(0,\rho)}\big{[}|\mathcal{L}(T;\theta+\epsilon)-\mathcal{L}(S;\theta+\epsilon)|\big{]}$, we partition the $\epsilon$ space into those with $\|\epsilon\|_{2} \leq \rho$ and $\|\epsilon\|_{2} > \rho$. As $\|\epsilon\|_{2} \leq \rho$ with probability at least $1-\frac{1}{\sqrt{n}}$, we have:
\begin{align}
    &\mathbb{E}_{\epsilon \sim \mathcal{N}(0,\rho)}\big{[}|\mathcal{L}(T;\theta+\epsilon)-\mathcal{L}(S;\theta+\epsilon)|\big{]} \nonumber \\
    &\leq (1-\frac{1}{\sqrt{n}})\max_{\|\epsilon\|_{2}\leq \rho}\Big{|} \mathcal{L}(T;\theta+\epsilon)-\mathcal{L}(S;\theta+\epsilon)\Big{|} + \frac{1}{\sqrt{n}}l_{max}\\
    &\leq  \max_{\|\epsilon\|_{2}\leq \rho}\Big{|} \mathcal{L}(T;\theta+\epsilon)-\mathcal{L}(S;\theta+\epsilon)\Big{|} + \frac{1}{\sqrt{n}}l_{max}
\end{align}
Here, $l_{max} =\max_{\|\epsilon\|_{2}\geq \rho}\Big{|} \mathcal{L}(T;\theta+\epsilon)-\mathcal{L}(S;\theta+\epsilon)\Big{|}$. By replacing original expectation term to $\max_{\|\epsilon\|_{2}\leq \rho}\Big{|} \mathcal{L}(T;\theta+\epsilon)-\mathcal{L}(S;\theta+\epsilon)\Big{|}$, Eq \eqref{before_final_objective} is derived as follows:
\begin{align}
    &\mathbb{E}_{\epsilon \sim \mathcal{N}(0,\rho)}\big{[}|\mathcal{L}(\mathbb{D};\theta+\epsilon)-\mathcal{L}(S;\theta+\epsilon)|\big{]} \nonumber \\
    & \leq \max_{\|\epsilon\|_{2}\leq \rho}\Big{|} \mathcal{L}(T;\theta+\epsilon)-\mathcal{L}(S;\theta+\epsilon)\Big{|}+ \frac{1}{\sqrt{n}}l_{max} + \sqrt{\frac{\frac{1}{4}k\log{(1+\frac{\|\theta\|^{2}_{2}}{k\sigma^{2}})}+\frac{1}{4}+\log{\frac{n}{\delta}}+2\log(6n+3k)}{n-1}} 
\end{align}
With the bounded $\theta$ with $k$ dimensions, The summation of second term and last term could be asymptotically described as $\sqrt{\frac{O(k+\log\frac{n}{\delta})}{n-1}}$. With the replacement of last two terms to the corresponding asymptotical term, we re-arrange above equation as follows:
\begin{align}
\label{theorem_proof_final_v1}
    &\mathbb{E}_{\epsilon \sim \mathcal{N}(0,\rho)}\big{[}|\mathcal{L}(\mathbb{D};\theta+\epsilon)-\mathcal{L}(S;\theta+\epsilon)|\big{]} \leq \max_{\|\epsilon\|_{2}\leq \rho}\Big{|} \mathcal{L}(T;\theta+\epsilon)-\mathcal{L}(S;\theta+\epsilon)\Big{|} + \sqrt{\frac{O(k+\log\frac{n}{\delta})}{n-1}}
\end{align}
By re-phrasing each term in Eq \eqref{theorem_proof_final_v1} into the shorter description, we conclude the proof as follows:
\begin{align}
&\mathbb{E}_{\epsilon \sim \mathcal{N}(0,\rho)}[\mathcal{L}_{abs}(\mathbb{D},S;\theta+\epsilon)]
\leq \underset{{||\epsilon||_{2}} \leq \rho }{\max}\mathcal{L}_{abs}(T,S;\theta+\epsilon)+ \sqrt{\frac{O(k+\log\frac{n}{\delta})}{n-1}}
\end{align}

\subsection{Proof of Corollary \ref{corollary_v1}}
\label{appendix:corollary}
We first refer the Corollary \ref{corollary_v1} here as follows:
\begin{corollary}
\label{corollary_v1_appendix} 
If $\mathcal{L}_{abs}(T,S;\theta) \!\leq\! \mathcal{L}_{abs}(\mathbb{D},S;\theta)$ for $\theta \in \hat{\Theta}$, with probability at least $1-\delta$ over the choice of the training set $T$ with $|T|=n$, the following holds: 
\begin{align}
&\Big{(}\mathbb{E}_{\epsilon \sim \mathcal{N}(0,\rho)}[\mathcal{L}_{abs}(\mathbb{D},S;\theta+\epsilon)]-\mathcal{L}_{abs}(\mathbb{D},S;\theta)\Big{)}\Big{/}\rho \leq \Big{(}\underset{{||\epsilon||_{2}} \leq \rho }{\max}\mathcal{L}_{abs}(T,S;\theta+\epsilon)-\mathcal{L}_{abs}(T,S;\theta)\Big{)}\Big{/}\rho + \sqrt{\frac{O(k+\log\frac{n}{\delta})}{n-1}}
\end{align}
\end{corollary}
\textit{Proof.} 
We first revisit the resulting equation by Theorem \ref{theorem_v1} as follows:
\begin{align}
\label{from_theorem_1}
&\mathbb{E}_{\epsilon \sim \mathcal{N}(0,\rho)}[\mathcal{L}_{abs}(\mathbb{D},S;\theta+\epsilon)]
\leq \underset{{||\epsilon||_{2}} \leq \rho }{\max}\mathcal{L}_{abs}(T,S;\theta+\epsilon)+ \sqrt{\frac{O(k+\log\frac{n}{\delta})}{n-1}}
\end{align}
As we assume that $\mathcal{L}_{abs}(T,S;\theta) \!\leq\! \mathcal{L}_{abs}(\mathbb{D},S;\theta)$ for $\theta \in \hat{\Theta}$, we can extend Eq \eqref{from_theorem_1} as follows:
\begin{align}
\label{from_theorem_2}
&\mathbb{E}_{\epsilon \sim \mathcal{N}(0,\rho)}[\mathcal{L}_{abs}(\mathbb{D},S;\theta+\epsilon)] - \mathcal{L}_{abs}(\mathbb{D},S;\theta)
\leq \underset{{||\epsilon||_{2}} \leq \rho }{\max}\mathcal{L}_{abs}(T,S;\theta+\epsilon) - \mathcal{L}_{abs}(T,S;\theta) + \sqrt{\frac{O(k+\log\frac{n}{\delta})}{n-1}}
\end{align}
By dividing both terms by $\rho$ we finish the proof. As $\rho$ is controllable hyper-parameter, which is usually set to value between 0.01 and 0.5, we do not reflect $\rho$ on the asymptotical term.
\begin{align}
&\Big{(}\mathbb{E}_{\epsilon \sim \mathcal{N}(0,\rho)}[\mathcal{L}_{abs}(\mathbb{D},S;\theta+\epsilon)]-\mathcal{L}_{abs}(\mathbb{D},S;\theta)\Big{)}\Big{/}\rho \leq \Big{(}\underset{{||\epsilon||_{2}} \leq \rho }{\max}\mathcal{L}_{abs}(T,S;\theta+\epsilon)-\mathcal{L}_{abs}(T,S;\theta)\Big{)}\Big{/}\rho + \sqrt{\frac{O(k+\log\frac{n}{\delta})}{n-1}}
\end{align}

\section{Further Analyses of LCMat}
\subsection{Analyses on $L_{abs}$}
\label{appendix:LCMat detailed anal}
First, we recap our objective as follows:
\begin{align}
\label{intro_objective_appen}
    &\underset{S}{\min}\underset{{||\epsilon||_{2}} \leq \rho }{\max}\frac{\mathcal{L}_{abs}(T,S;\theta\!+\!\epsilon) \!\!-\!\! {\mathcal{L}_{abs}(T,S;\theta)}}{\rho}
\end{align}
As stated in the main paper, optimization of Eq \eqref{intro_objective_appen} will lead to 1) the minimization of $\mathcal{L}_{abs}(T,S;\theta\!+\!\epsilon)$; and 2) the maximization of ${\mathcal{L}_{abs}(T,S;\theta)}$, respectively. The minimization of $\mathcal{L}_{abs}(T,S;\theta\!+\!\epsilon)$ is profitable, which is also shown in Theorem 1. The maximization of $\mathcal{L}_{abs}(T,S;\theta\!+\!\epsilon)$ is beneficial to some extent, in that it slightly regularizes the over-fitting of $S$ to $T$ based on the current parameter $\theta$. However, it could also lead to the under-fitting of $S$ based on $\theta$ if ${\mathcal{L}_{abs}(T,S;\theta)}$ increases too much. In our practical implementation, the value of $\mathcal{L}_{abs}(T,S;\theta)$ is bounded or regularized during the optimization. 

\textbf{Selection-based methods} For selection-based methods, A subset $S$ is constructed from $T$ as $S \subseteq T$, where our current parameter $\theta$ is assumed to be pre-trained on $T$. As the optimization of $\theta$ based on $T$ incurs $\mathcal{L}(T;\theta)$ to be small, we assume that the increase of $\mathcal{L}_{abs}(T,S;\theta)$ is induced by the large value of $\mathcal{L}(S;\theta)$ than $\mathcal{L}(T;\theta)$. Having said that, $\mathcal{L}_{abs}(T,S;\theta)$ gets the bound from the fixed state of $\theta$ and $T$ as follows:
\begin{align}
\label{upper_bound_selection_s}
    \mathcal{L}_{abs}(T,S;\theta) \leq \underset{S \subseteq T}{\max }{\mathcal{L}_{abs}(T,S;\theta)}=\underset{S \subseteq T}{\max }\Big{(}\mathcal{L}(S;\theta) - \mathcal{L}(T;\theta)\Big{)} \leq \underset{S \subseteq T}{\max}\mathcal{L}(S;\theta) - \mathcal{L}(T;\theta)
\end{align}
\textbf{Condensation-based methods} We recap our objective for application of condensation-based methods, LCMat-C, as follows:
\begin{align}
\label{condense_objective_appen}
     &\min_{S} \mathbb{E}_{\theta^{\small{0}} \sim P_{\theta^{\small{0}}}}\Big{[}
     \sum^{}_{k}\mathcal{D}(\bar{\mathbf{g}}^{T}_{\theta_{k}},\bar{\mathbf{g}}^{S}_{\theta_{k}})
     +\frac{1}{2}\rho|\text{Var}(\mathbf{G}^{T}_{\theta_{k}})-\text{Var}(\mathbf{G}^{S}_{\theta_{k}})|\Big{]} \nonumber \\
     &\,\,\,\,\,\, \,\,\,  \text{ s.t. } \,\,\, \theta_{t+1} = \theta_{t}-\eta\bar{\mathbf{g}}^{T}_{\theta_{t}}\text{ for } t=0,...,k-1. 
\end{align}
As noted in the main paper, the objective is composed of 1) $\mathcal{D}(\bar{\mathbf{g}}^{T}_{\theta_{k}},\bar{\mathbf{g}}^{S}_{\theta_{k}})$, which is averaged gradient matching between $T$ and $S$; and 2) $|\text{Var}(\mathbf{G}^{T}_{\theta_{k}})-\text{Var}(\mathbf{G}^{S}_{\theta_{k}})|$, which is gradient variance matching between $T$ and $S$. In practice, gradient variance matching is conducted based on the classifier parameters, where the classifier parameter weight and bias term is denoted as $w$ and $b$, respectively. We utilize the findings from \cite{fishr} as follows:

When we utilize cross-entropy as a loss function, the derivative of sample $(x,y)$ with respect to $b$ is $\nabla_{b}\ell(x,y;\theta)=(\hat{y}-y)$, where $\hat{y}$ is softmax output; and $y$ is true label. Hence, when we compute the gradient variance based on a certain dataset $D$, the gradient variance is computed as $\mathbf{v}^{D}_{b} = \frac{1}{|D|}\sum^{|D|}_{i=1}(\hat{y_{i}}-y_{i})^{2}$, which is equivalent to the mean squared error between the $\hat{y}$ and $y$. Accordingly, the gradient variance matching of $T$ and $S$ based on the classifier bias term is equivalent to matching the mean squared error of $T$ and $S$. Although the exact loss function is defined as cross-entropy, matching the mean squared error implicitly regularizes the difference between $\mathcal{L}(T;\theta)$ and $\mathcal{L}(S;\theta)$ to be small during the condensation procedure.
\begin{figure*}[h!]
\begin{subfigure}{0.495\textwidth}
\includegraphics[width=\linewidth]{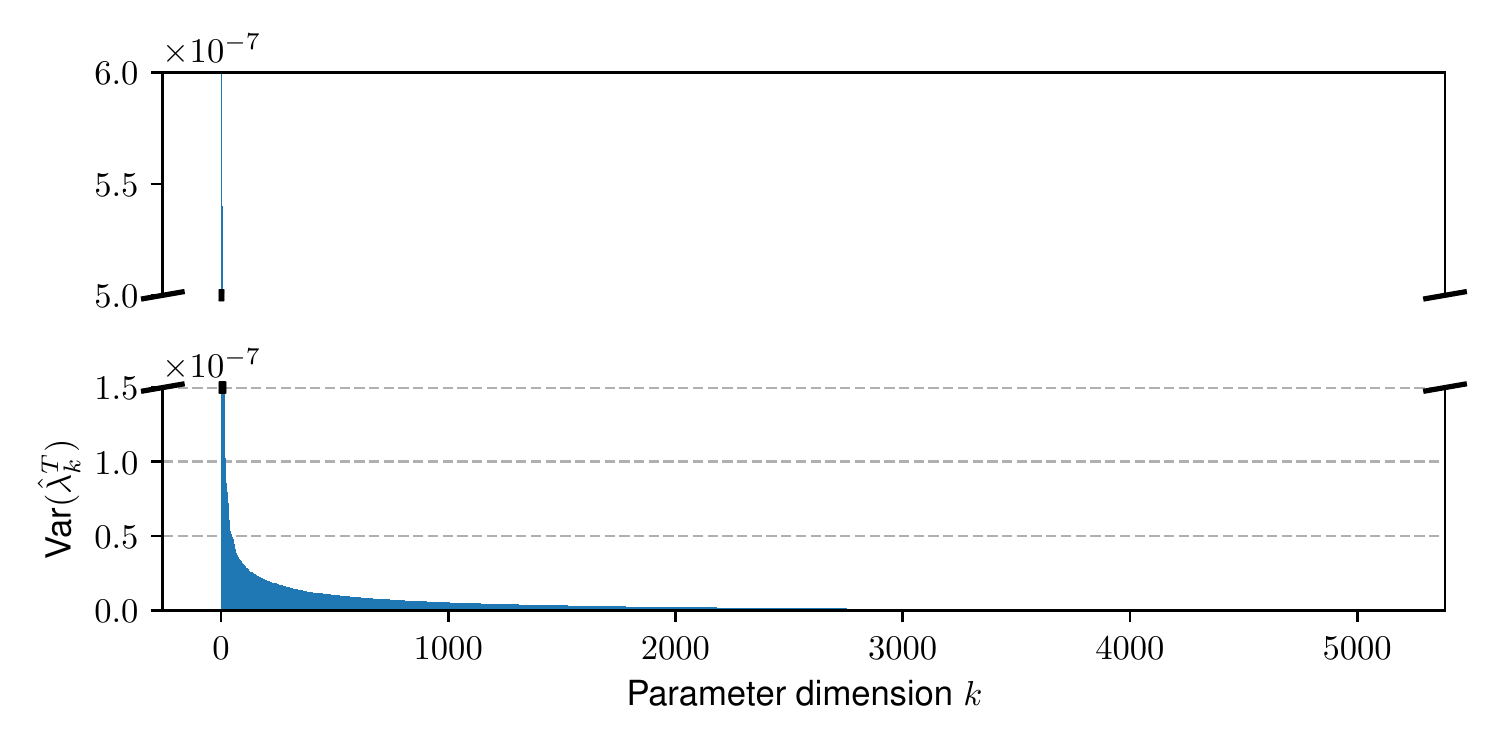}
\caption{Sorted $\text{Var}(\hat{\boldsymbol{\lambda}}^{T}_{k})$ for whole dimensions of classifier parameter.}
\label{fig:whole_dimension}
\end{subfigure}
\hfill
\begin{subfigure}{0.495\textwidth}
\includegraphics[width=\linewidth]{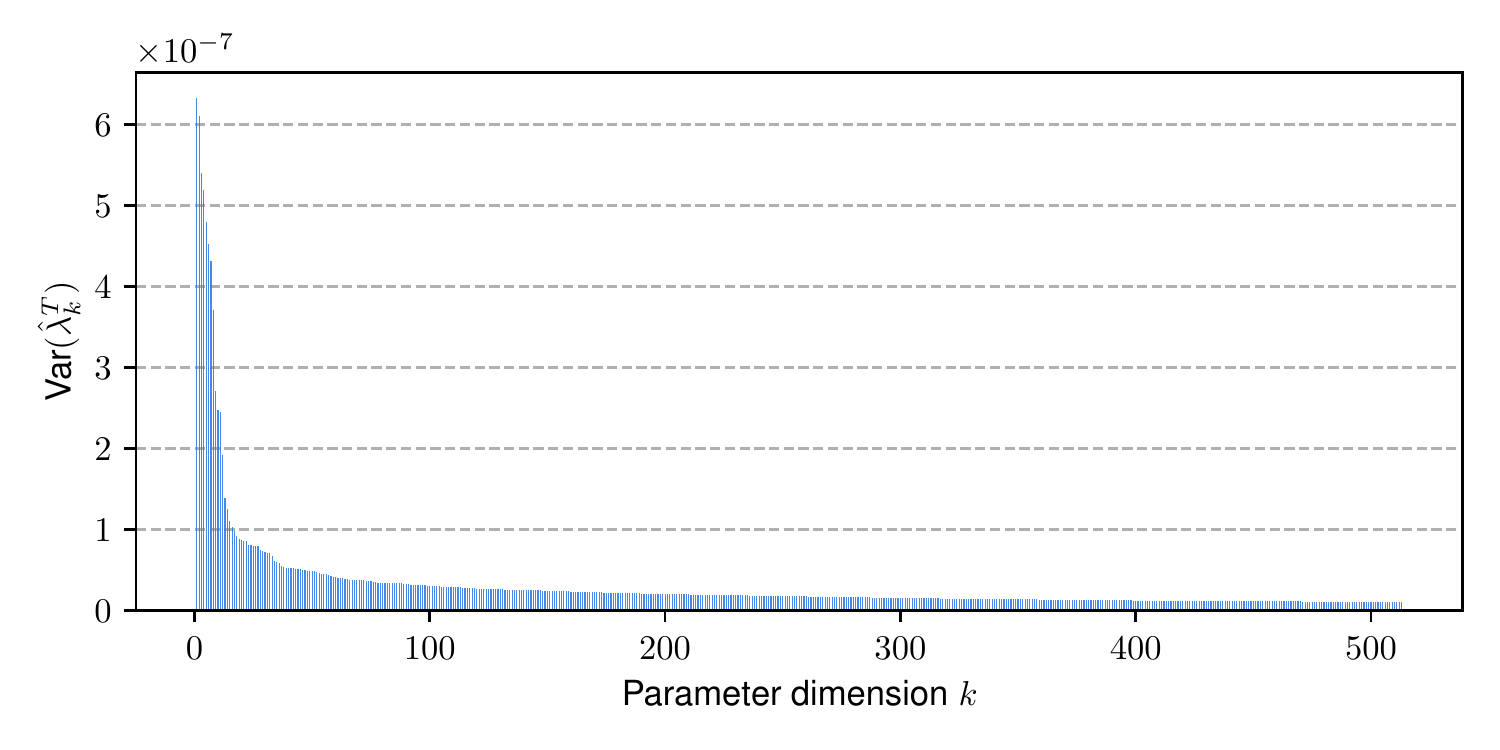}
\caption{Sorted $\text{Var}(\hat{\boldsymbol{\lambda}}^{T}_{k})$ for Top-500 dimensions of classifier parameter.}
\label{fig:sub_dimension}
\end{subfigure}
\caption{Empirically measured and sorted $\text{Var}(\hat{\boldsymbol{\lambda}}^{T}_{k})$ for whole dimensions of classifier parameters in ResNet-18 on learning with CIFAR-10. As we construct the subset based on class-wise comparison, we measure $\text{Var}(\hat{\boldsymbol{\lambda}}^{T}_{k})$ from the samples with same class. (Airplane for these figures)}
\label{fig:sub_dimension_selection}
\end{figure*}%
\subsection{Analyses on Sub-Dimension Selection}a
\label{appendix:selection-k-dim}
As noted in the main paper, we recap the sub-dimension selection criteria from whole parameter dimension. Let $\mathcal{K}$ be a set of indexes for $K$ sub-dimensions on $\theta$. We select $K$ dominant sub-dimensions based on the variance of $\hat{\boldsymbol{\lambda}}^{T}_{k}=[\hat{\lambda}^{T}_{i,k}]^{|T|}_{i=1}$ for each $k$, which is denoted by the set $\mathcal{K} = \underset{\mathcal{K},|\mathcal{K}|=K}{\text{argmax}}\sum_{j\in \mathcal{K}}\text{Var}(\hat{\boldsymbol{\lambda}}^{T}_{k})$. We assume that the large variance from specific parameter dimension means that there is a big difference in the corresponding eigenvalue of per-sample Hessian matrix for each sample. The difference between the averaged eigenvalue gap between the randomly selected subset and the entire training dataset would also likely to be large. In Figure \ref{fig:sub_dimension_selection}, $\text{Var}(\hat{\boldsymbol{\lambda}}^{T}_{k})$ shows the long-tailed distribution with concentration on specific dimensions from whole dimensions of parameter. In practice over the experiments of ResNet-18, we choose Top-$100$ dimensions from $5130$ dimensions of classifier parameters. Although setting $K = 100$ shows robust results over the experiments with ResNet-18, the optimal $K$ could be slightly different if we change the network structure for measuring $\text{Var}(\hat{\boldsymbol{\lambda}}^{T}_{k})$.

\subsection{Discussion on the Limitations and Social Impacts of LCMat}
\textbf{Limitations} The calculation of Hessian matrix over the model parameter induces the computational overhead during the optimization. As our method introduces the computation of Hessian over the classifier parameter, the computation of Hessian matrix could be costly when the number of feature dimensions and class dimensions further increases from the current experimental setting.

\textbf{Social Impacts} The data selection inevitably accompanies the discrimination of some samples than other samples, which are discarded from the dataset reduction procedure. Recently, \cite{privacy_for_free} found out that the condensation-based methods can be utilized to relieve the privacy issues by erasing the privacy-related information of each sample during the condensation. As we provide an application of our method for condensation-based method, we conjecture that our method can also be utilized as a privacy-robust method for dataset reduction task.

\section{Technical Survey of Methods for Dataset Reduction}
\subsection{Selection-based Methods}
\label{appendix:selection-based methods survey}
Selection-based methods find a data subset $S \subset T$ that satisfies the cardinality constraint while maximizing the objective defined by the informativeness of $S$. We report details of the previous researches of selection-based methods in this section.

\textbf{Herding} \citep{Herding} selects data points considering the distance between the feature center of the full dataset; and the feature center of the selected subset, and it selects samples to regularize the centers from each dataset to be similar. 

\textbf{k-CenterGreedy} \citep{Kcenter} solves the coreset selection problem as k-Center problem (minimax facility location \citep{minimaxfacility}). Since k-Center problem is NP-Hard, it provides an approximate greedy solution for the problem by firstly selecting any sample as initialization and adding samples with maximum distances that have not been included to the coreset gradually. 

\textbf{ContextualDiversity} \citep{contextualdiversity} is similar to \cite{Kcenter}, but it calculates the distance between two feature inputs using the summation of KL-divergence and reverse KL-divergence.

\textbf{Forgetting} \citep{Forgetting} assumes samples, which are not forgettable during the training procedure, are reducible. It defines forgetting as the event of wrong classification a sample when the model prediction of the sample was correct in the previous epoch. After a few epochs of training, samples are selected based on the number of forgotten times, which is counted for each sample. Therefore, it requires 1) saving all model prediction results from whole iterations and 2) an adequate number of training to get credible forgetting score. 

\textbf{GraND} \citep{Data-Diet} calculates the expectation of the loss gradient with regard to model parameter. It is analytically regarded as the contribution of each sample to the averaged training loss. GraND also utilizes the outputs from multiple models, where each model is randomly initialized. Since these two methods both necessarily requires multiple times of model training with full dataset, we consider the framework of these methods is quiet different from our method. Hence, we do not report them as our baselines. 

\textbf{Uncertainty} based methods \citep{Uncertainty}, which include LeastConfidence, Entropy and Margin in our baselines, assume that data samples with lower level of model prediction confidence would have larger impact on the construction of decision boundary. The scores of LeastConfidence, Entropy and Margin are defined as  $1-\text{max}_{i=1,...,C}P(\hat{y}=i|x)$, $-\sum_{i=1}^{C}P(\hat{y}=i|x)\text{log}P(\hat{y}=i|x)$, and $1-\text{min}_{y\neq \hat{y}}\left(P\left(\hat{y}|x \right)-P\left(y|x \right)\right)$, respectively. They select samples based on the computed scores in descending order.

Gradient-based methods minimize the distance between the gradients from the training dataset $T$; and the (weighted) gradients from $S$ as follows:
\begin{align}
\label{gradmatch_v1}
\min_{\mathbf{w}, S}  \mathcal{D}\Big{(}\frac1{|T|}\sum\limits_{(x, y)\in T}&\nabla_{\theta} \ell(x, y; \theta), \frac1{\|\mathbf{w}\|_{1}}\sum\limits_{(x, y)\in S} w_{x}\nabla_{\theta} \ell(x, y; \theta)\Big{)} \\
&\text{s.t.} \quad  S \subset T, \;  w_{x}\geq0 \nonumber
\end{align}
Here, $\mathbf{w}$ is the vector of learnable weights for the data instances in $S$; $\|\mathbf{w}\|_{1}$ is l1-norm of $\mathbf{w}$; and $\mathcal{D}$ measures the distance between gradients. 

To solve the problem, \textbf{Craig} \citep{Craig} converts Eq \eqref{gradmatch_v1} into the submodular maximization problem, and this research utilizes the greedy approach to optimize Eq \eqref{gradmatch_v1}. 

Compared to Craig \citep{Craig}, \textbf{GradMatch} \citep{gradmatchcoreset} utilizes orthogonal matching pursuit algorithm \citep{omp} and squared $L_{2}$ regularization term over $\mathbf{w}$ to stabilize the optimization. 

\textbf{Glister} \citep{glister} introduces the generalization-based method, which results in the extraction of subsets which approximate the gradient of a training dataset or additional validation dataset well. 

\textbf{AdaCore} \citep{adacore} replaces $\nabla_\theta l(x,y;\theta)$ in Eq \eqref{gradmatch_v1} with a preconditioned gradient with the Hessian matrix, which leverages the second-order information for optimization. It firstly suggests a way of utilizing hessian information for coreset selection, however, the optimization is conducted to match the pre-conditioned gradients, which could also be generalized into loss-curvature matching method based on the pre-conditioned gradients.

\subsection{Condensation-based Methods}
\label{appendix:condensation-based methods survey}
\textbf{DC} \citep{DC} formulates the condensation method as a gradient matching task between the gradients of deep neural network weights, that are trained on the original and our
synthetic data. Recently, \cite{EDC} have introduced bag-of-tricks to improve the condensation quality with the gradient matching objective. It should be noted that these tricks could be orthogonally applied upon the choice of objective functions.

\textbf{DSA} \citep{DSA} proposes the Differentiable Siamese Augmentation (DSA), which utilizes the same data transformation to original data instances and synthetic data instances at each training iteration. Additionaly, it enables the update of data transformation policy by back-propagating the gradient of the loss with respect to synthetic data into the augmentation parameters. Similar to \cite{EDC}, DSA are orthogonally applied upon the condensation objectives.

\textbf{DM} \citep{DM} proposes matching feature distributions of the original dataset and synthetic dataset in sampled embedding spaces. As feature matching do not necessarily need the bi-level optimization between the model parameter, $\theta$, and the condensed dataset, $S$, it significantly reduces the computational costs of the gradient matching \cite{DC}. However, the distribution matching do not provide the theoretical meaning of the introduced objective.

\textbf{KIP} \citep{kip} proposes a kernel-based objective which utilizes infinitely-wide neural networks. As condensed dataset is equivalent to the kernel inducing points from the kernel ridge-regression, it could be recognized as dataset summarization with kernel.

\section{Experimental Details and Further Results}
\label{experimental_details}
\subsection{Experimental Details}
\paragraph{Coreset Selection Evaluation}
For coreset selection task, we use batch size of 128 for both CIFAR-10 dataset and CIFAR-100 dataset, for both model training for coreset selection; and model training with the selected instances. For the model optimization, We use SGD optimizer which utilizes learning rate of 0.1, momentum of 0.9, and weight decay (L2 regularization) parameter of 5$\times 10^{-4}$. After the extraction of $S$, we train the model with the selected instances, $S$, for 200 epochs. To validate the robustness of LCMat-S for each fraction budget, we report results with fraction over $[0.1\%, 0.5\%, 1\%, 5\%, 10\%, 20\%, 30\%]$. We omit fraction of $0.1\%$ condition for CIFAR-100, which chooses only one sample per class. We also report the performance of the model trained with the full dataset (100$\%$). We consider it as the upper bound. For augmentation module, we utilize RandomCrop with reflection padding 4, RandomHorizontalFlip with probability 0.5, and Normalization for both CIFAR-10 and CIFAR-100 dataset. There are methods which require either outputs of a model or gradient signals from a model for each sample. To get such information, we trained a model with random initialized parameter for 10 epochs (Please refer to Appendix 4.5 for the sensitivity analysis on the number of this training epochs.). For gradient matching method such as Craig, GradMatch, Glister and AdaCore, we use the gradient signal of the last layer of the model because of the computation issue, as they did in the original paper \citep{Craig, gradmatchcoreset, glister, adacore}. We select samples with class-balanced manner, meaning that the number of samples in the selected subset for each class should be balanced.
\paragraph{Condensed Dataset Evaluation}
As we learn $\theta$ with $T$ from the inner loop of bi-level optimization, we learn $\theta$ with 1 epoch per one inner loop. The learning rates for model network and data variable are set to 0.01 and 0.005, respectively. Similar to \cite{EDC}, we utilize augmentation sequecne of color transform, crop, and cutout for data objective learning. Additionally, the initialization of the synthetic data is set to noise initialization. The evaluation scenario comes with the fraction budget, where we set 10, 50 samples per class as a practice setting.

\subsection{Wall-Clock Time Analyses of Selection-based Methods} 
In this section, we compare the computation time taken for each method over the different fraction budget: $0.1\%$, $1\%$, and $10\%$. Wall-clock time calculation includes 1) pre-training model with full dataset (with 10 epochs); and 2) subset selection process. As Uniform do not need the process of pre-training model with full-dataset, we skip the process for  Uniform. AdaCore and LCMat-S, which are methods which utilizes Hessian matrix during the selection, show significant increase of Wall-Clock time on the large fraction budget. It should noted that the computation time of LCMat-S could be reduced if we utilize the faster approximation of $L_{1}$ norm, which is introduced on the computation of Hessian matrix difference. In addition, it should be noted that AdaCore and LCMat-S only shows the consistently competitive performances over the Uniform baseline, which emphasizes the importance of modelling Hessian matrix during the selection procedure.

\begin{table}[h]
\centering
\resizebox{\textwidth}{!}{%
\begin{tabular}{c cccccccccccc}
\toprule
Fraction & Uniform & C-Div & Herding & k-Center & L-Conf & Entropy & Margin & Craig & GradMatch & Glister & AdaCore & LCMat-S \\ \cmidrule(lr){1-1} \cmidrule(lr){2-8} \cmidrule(lr){9-13}
0.001 & \multirow{3}{*}{0.04} & 201.15 & 205.09 & 458.78 & 202.35 & 203.93 & 204.14 & 224.82 & 199.39 & 204.68 & 204.38 & 458.92  \\
0.01 &  & 205.88 & 202.69 & 455.34 & 204.32 & 201.68 & 204.27 & 234.17 & 209.72 & 203.56 & 304.50 & 627.50 \\
0.1 &  & 202.89 & 205.86 & 458.80 & 203.68 & 196.51 & 202.23 & 246.97 & 259.59 & 204.74 & 1255.49 & 1400.56\\ \bottomrule
\end{tabular}%
}
\label{tab:wallclock-time}
\end{table}

\subsection{Results with Inception-v3}
Here, we report the test accuracy of the Inception-v3 trained by the $S$ from each method. As reported in the main paper, we evaluate $S$ with different fractions in dataset reduction, which is the cardinality budget of $S$ from $T$. Similar to the results computed from the ResNet-18 network, Uniform shows competitive performances over other baselines, meaning the weak robustness of the existing selection methods. LCMat-S shows competitive performances over the implemented baselines by relieving the over-fitting issue of $S$ to the provided $\theta$. Comparing the results from ResNet-18 and Inception-V3, the results of LCMat-S from Inception-V3 shows degraded performance than the ones from ResNet-18. As the number of classifier dimensions from Inception-V3 is 4 times bigger than the one from ResNet-18, our method could not cover the whole dimensions to compute the corresponding Hessian matrix. We assume that sub-dimension computation of Hessian matrix could be naive when original parameter dimension is too large to cover the whole dimensions by sub-dimension computation.

\begin{table}[h]
  \centering
  \caption{Coreset selection performances on CIFAR10. We train randomly initialized Inception-v3 on the coresets selected by different methods and then test on the real testing set.}  \label{tab:cifar-inception}
  \resizebox{0.8\textwidth}{!}{
\begin{tabular}{c cccccccc}
\toprule
     & \multicolumn{8}{c}{\textbf{CIFAR10}}    \\ \cmidrule(lr){2-9} 
Fraction     & 0.1\%       & 0.5\%       & 1\%       & 5\%       & 10\%      & 20\%      & 30\%     & 100\%      \\ \cmidrule(lr){1-1} \cmidrule(lr){2-9}
Uniform &17.59$\pm$\scriptsize{2.9}&27.24$\pm$\scriptsize{2.3}&35.29$\pm$\scriptsize{0.3}&60.09$\pm$\scriptsize{1.1}&76.73$\pm$\scriptsize{1.3}&85.52$\pm$\scriptsize{0.6}&89.25$\pm$\scriptsize{0.5}&\multirow{12}{*}{95.62$\pm$\scriptsize{0.1}}\\
C-Div &11.94$\pm$\scriptsize{0.4}&19.26$\pm$\scriptsize{0.7}&21.9$\pm$\scriptsize{2.4}&35.89$\pm$\scriptsize{3.7}&55.18$\pm$\scriptsize{2.4}&82.99$\pm$\scriptsize{0.7}&90.3$\pm$\scriptsize{0.5}&\\
Herding &14.52$\pm$\scriptsize{0.5}&26.03$\pm$\scriptsize{2.4}&32.06$\pm$\scriptsize{2.4}&49.86$\pm$\scriptsize{4.4}&64.98$\pm$\scriptsize{1.5}&75.56$\pm$\scriptsize{1.0}&80.99$\pm$\scriptsize{0.2}&\\
k-Center &15.81$\pm$\scriptsize{1.0}&20.4$\pm$\scriptsize{0.4}&25.48$\pm$\scriptsize{0.4}&48.8$\pm$\scriptsize{3.3}&75.47$\pm$\scriptsize{1.8}&85.72$\pm$\scriptsize{0.4}&90.08$\pm$\scriptsize{0.1}&\\
L-Conf &13.36$\pm$\scriptsize{1.3}&14.88$\pm$\scriptsize{0.7}&19.19$\pm$\scriptsize{2.2}&34.85$\pm$\scriptsize{3.1}&60.75$\pm$\scriptsize{2.8}&82.66$\pm$\scriptsize{1.3}&89.92$\pm$\scriptsize{0.1}&\\
Entropy &12.73$\pm$\scriptsize{0.6}&16.26$\pm$\scriptsize{2.2}&17.91$\pm$\scriptsize{1.5}&37.53$\pm$\scriptsize{3.1}&54.94$\pm$\scriptsize{1.8}&82.54$\pm$\scriptsize{1.1}&89.97$\pm$\scriptsize{0.9}&\\
Margin &15.29$\pm$\scriptsize{1.3}&23.81$\pm$\scriptsize{1.5}&26.71$\pm$\scriptsize{1.2}&43.14$\pm$\scriptsize{1.5}&63.29$\pm$\scriptsize{3.3}&83.36$\pm$\scriptsize{1.2}&90.14$\pm$\scriptsize{0.5}&\\
Craig &13.54$\pm$\scriptsize{0.8}&22.50$\pm$\scriptsize{1.8}&24.55$\pm$\scriptsize{5.9}&38.05$\pm$\scriptsize{1.6}&52.13$\pm$\scriptsize{6.6}&71.00$\pm$\scriptsize{3.1}&82.68$\pm$\scriptsize{1.7}&\\
GradMatch &12.73$\pm$\scriptsize{1.1}&18.24$\pm$\scriptsize{1.4}&18.69$\pm$\scriptsize{0.6}&35.56$\pm$\scriptsize{2.9}&50.91$\pm$\scriptsize{4.5}&68.95$\pm$\scriptsize{1.3}&83.34$\pm$\scriptsize{0.6}&\\
Glister &15.52$\pm$\scriptsize{1.5}&21.82$\pm$\scriptsize{1.6}&22.11$\pm$\scriptsize{1.9}&34.71$\pm$\scriptsize{0.7}&48.98$\pm$\scriptsize{4.6}&70.13$\pm$\scriptsize{2.9}&84.33$\pm$\scriptsize{0.6}&\\
AdaCore &15.79$\pm$\scriptsize{1.8}&27.48$\pm$\scriptsize{1.2}&33.93$\pm$\scriptsize{0.2}&58.57$\pm$\scriptsize{1.5}&71.97$\pm$\scriptsize{3.9}&86.2$\pm$\scriptsize{0.6}&90.54$\pm$\scriptsize{0.4}&\\ \cmidrule(lr){1-1} \cmidrule(lr){2-8}
LCMat-S  &18.55$\pm$\scriptsize{1.8}&29.33$\pm$\scriptsize{0.5}&36.09$\pm$\scriptsize{0.6}&53.23$\pm$\scriptsize{2.1}&69.14$\pm$\scriptsize{1.3}&85.21$\pm$\scriptsize{1.2}&89.89$\pm$\scriptsize{0.2}& \\ \bottomrule
\end{tabular}}
\end{table}
\newpage
\subsection{Selected Images for All Class in Cifar-10}
\label{selected_images}
\begin{figure}[h!]
\centering
\begin{subfigure}[b]{0.195\columnwidth}
    \centering
    \includegraphics[width=\columnwidth]{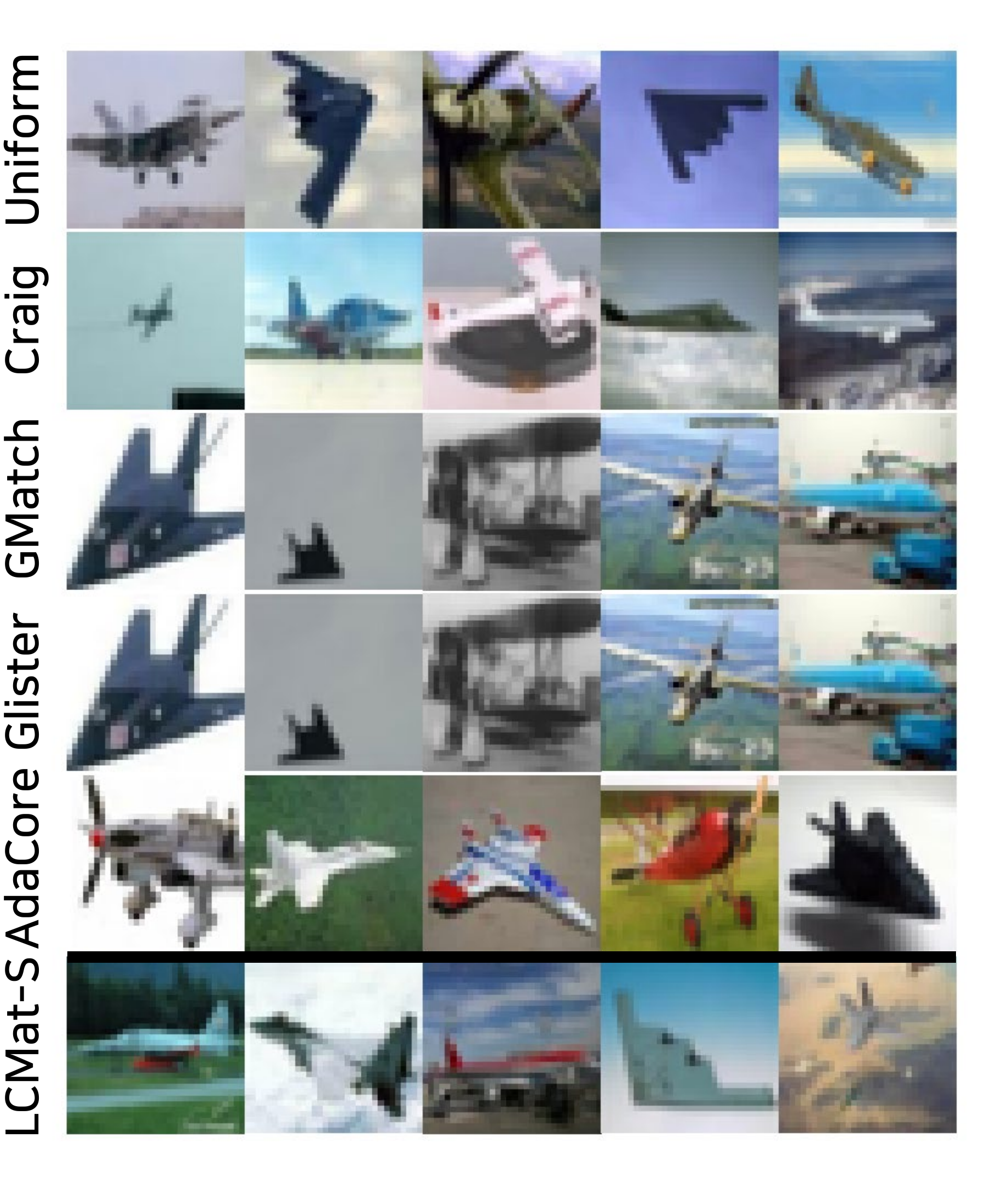}
    \caption{Airplane}
\end{subfigure}\hfill
\begin{subfigure}[b]{0.195\columnwidth}
    \centering
    \includegraphics[width=\columnwidth]{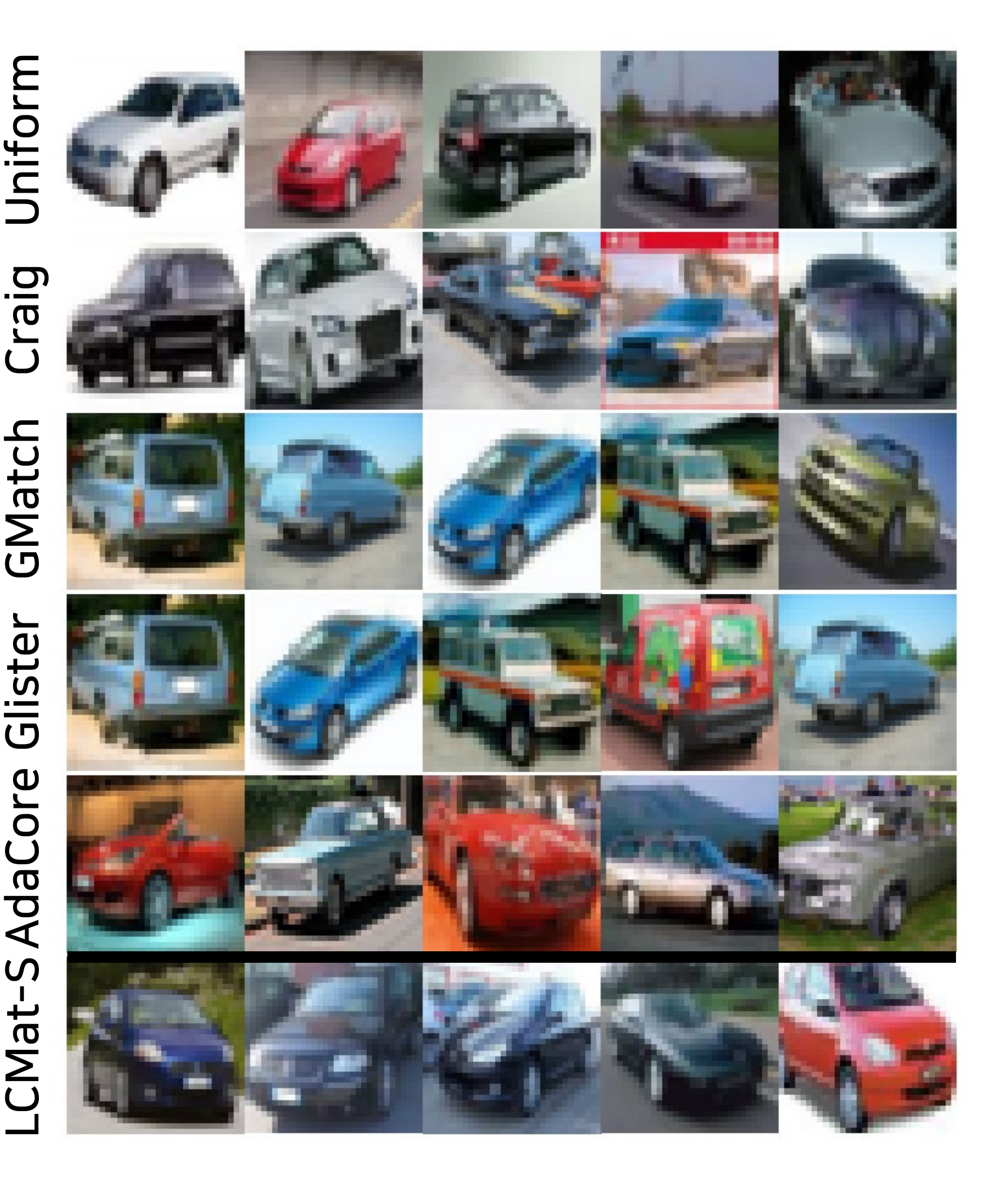}
    \caption{Automobile}
\end{subfigure}\hfill
\begin{subfigure}[b]{0.195\columnwidth}
    \centering
    \includegraphics[width=\columnwidth]{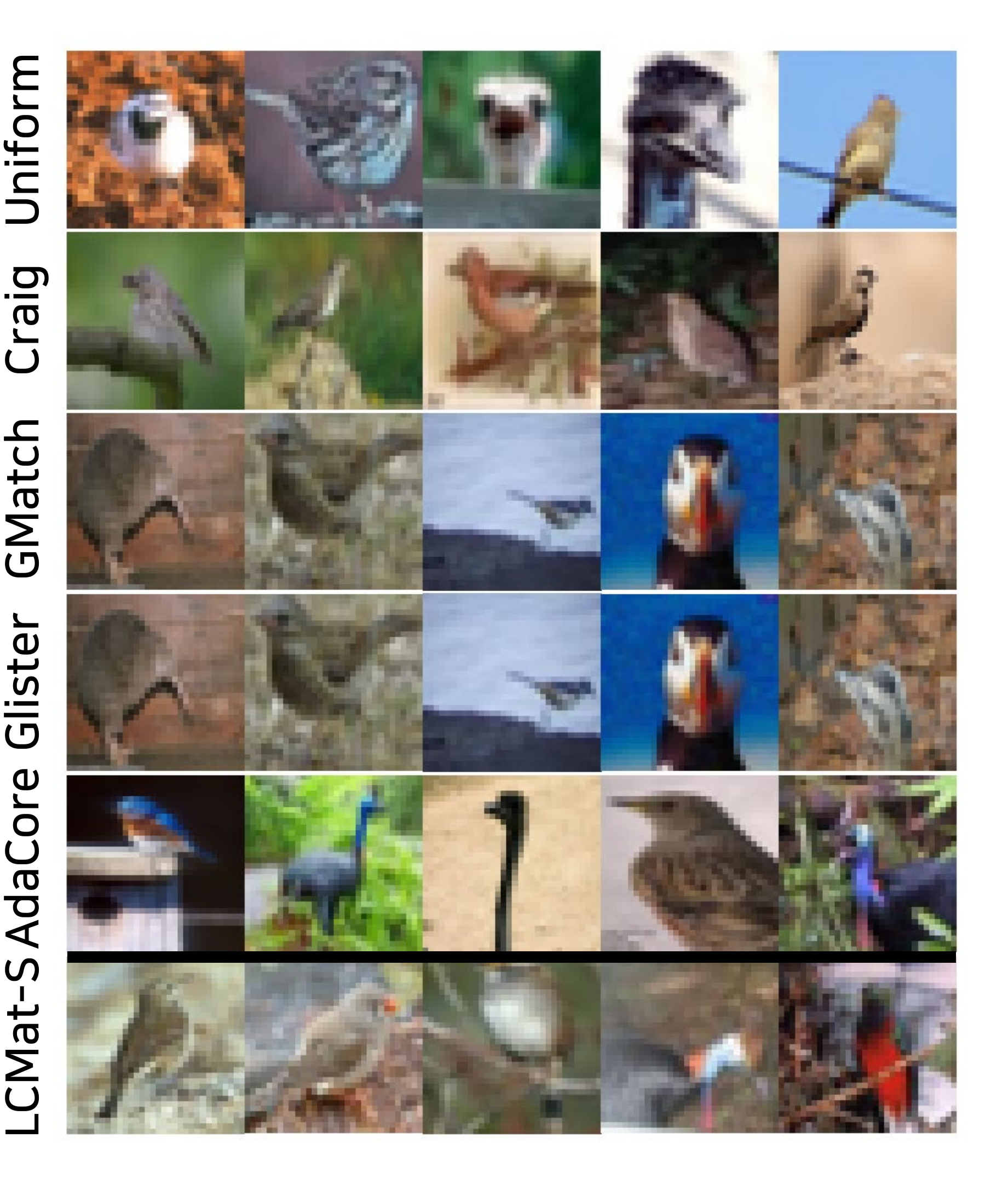}
    \caption{Bird}
\end{subfigure}\hfill
\begin{subfigure}[b]{0.195\columnwidth}
    \centering
    \includegraphics[width=\columnwidth]{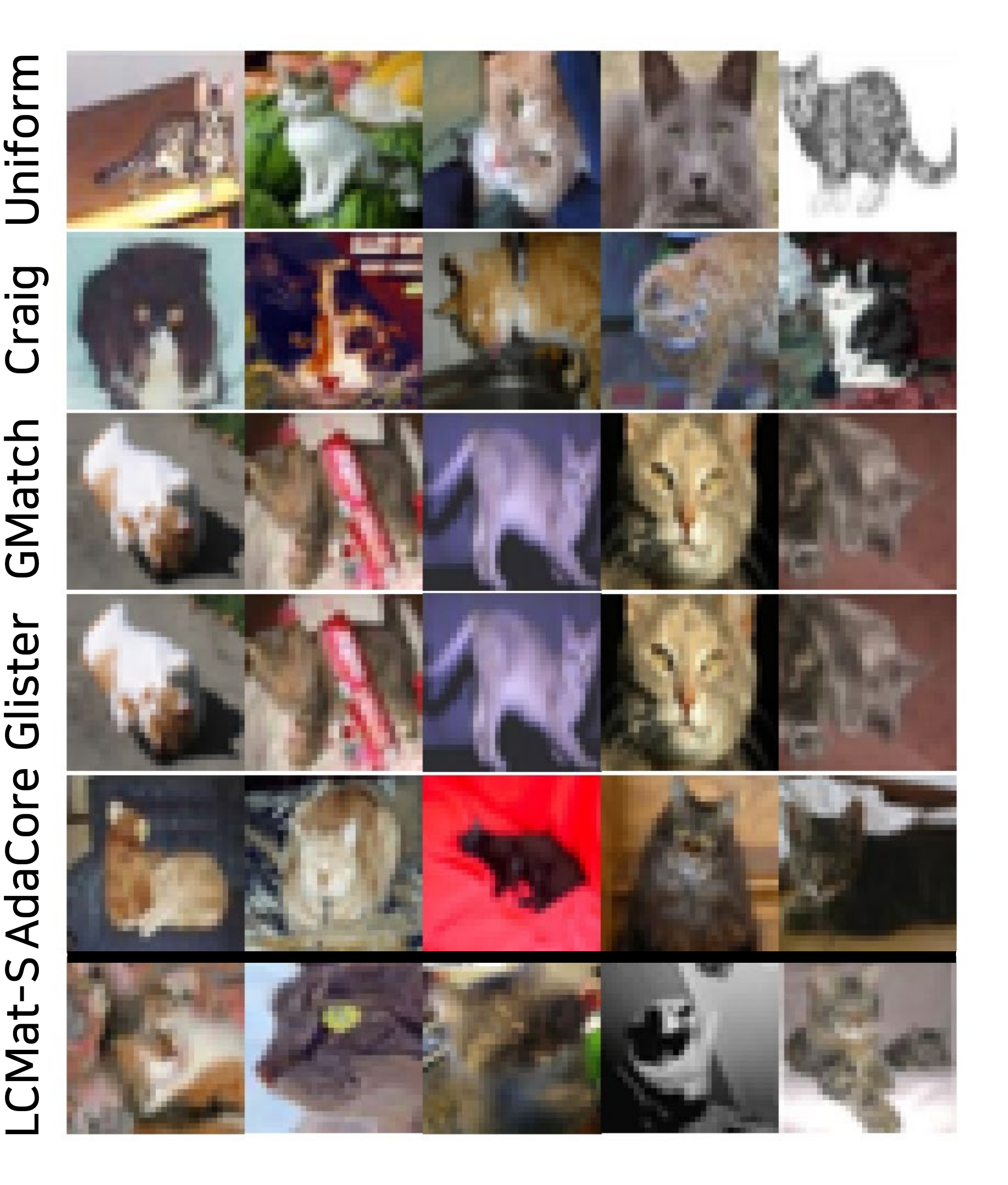}
    \caption{Cat}
\end{subfigure}\hfill
\begin{subfigure}[b]{0.195\columnwidth}
    \centering
    \includegraphics[width=\columnwidth]{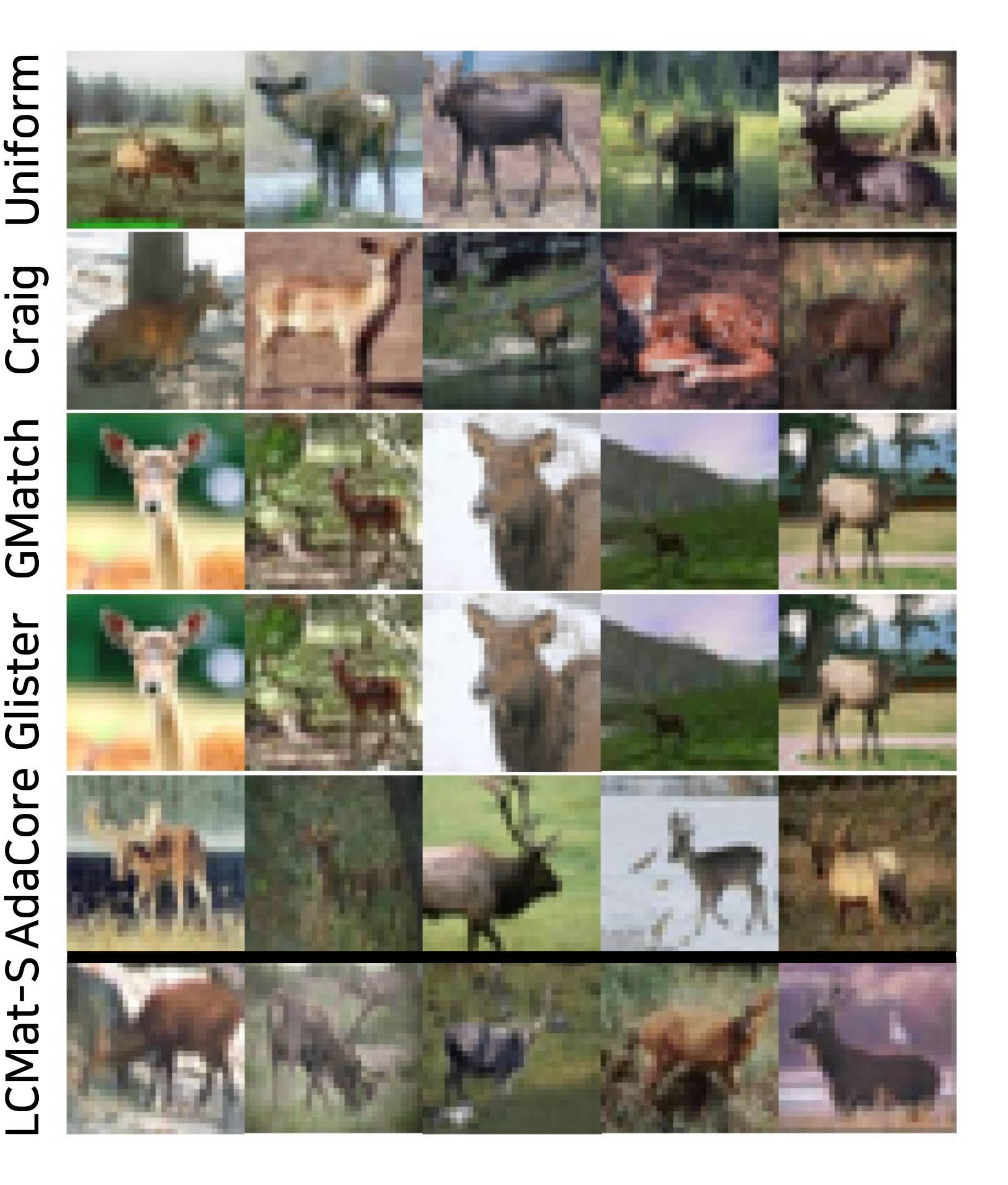}
    \caption{Deer}
\end{subfigure}\hfill
\begin{subfigure}[b]{0.195\columnwidth}
    \centering
    \includegraphics[width=\columnwidth]{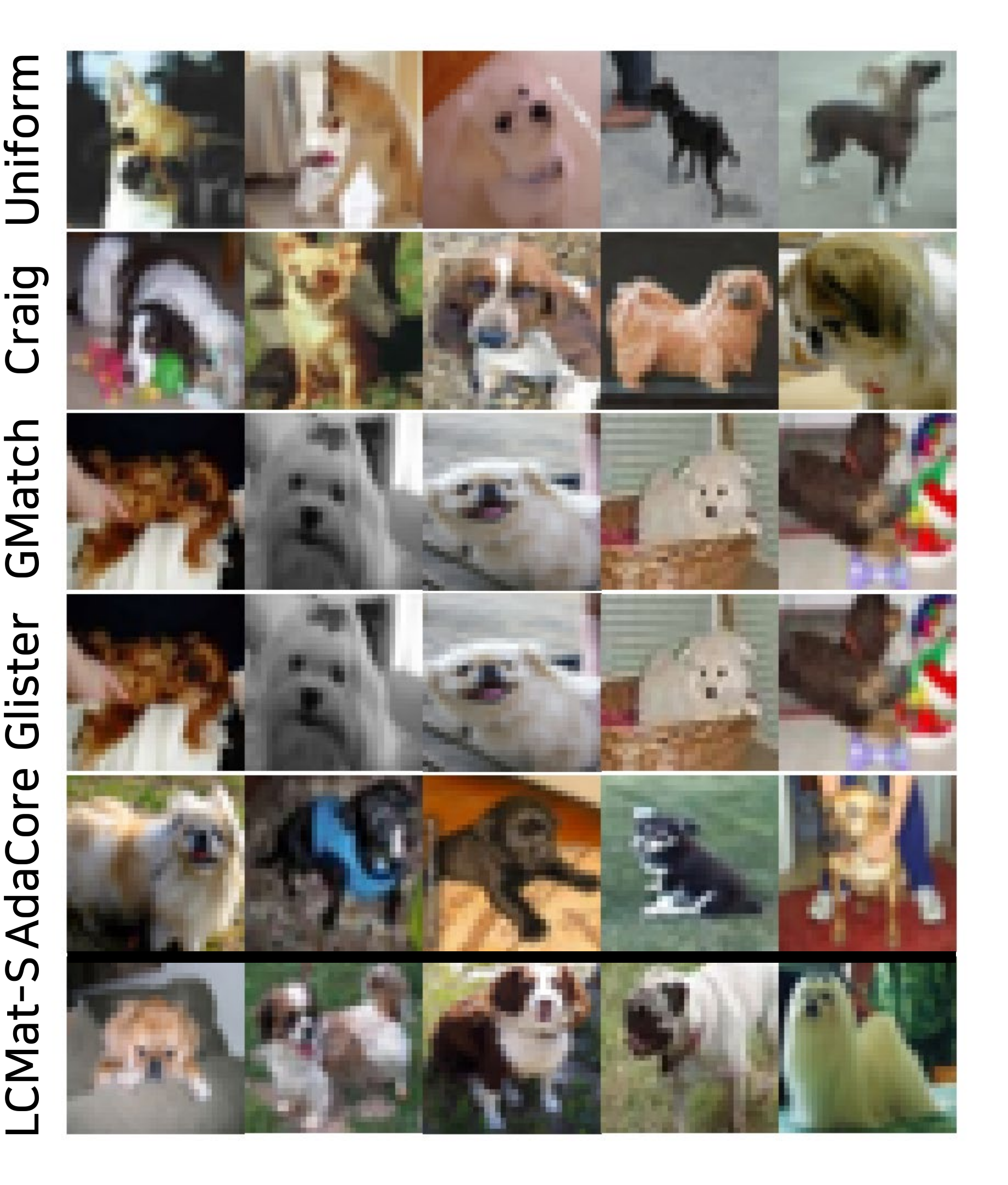}
    \caption{Dog}
\end{subfigure}\hfill
\begin{subfigure}[b]{0.195\columnwidth}
    \centering
    \includegraphics[width=\columnwidth]{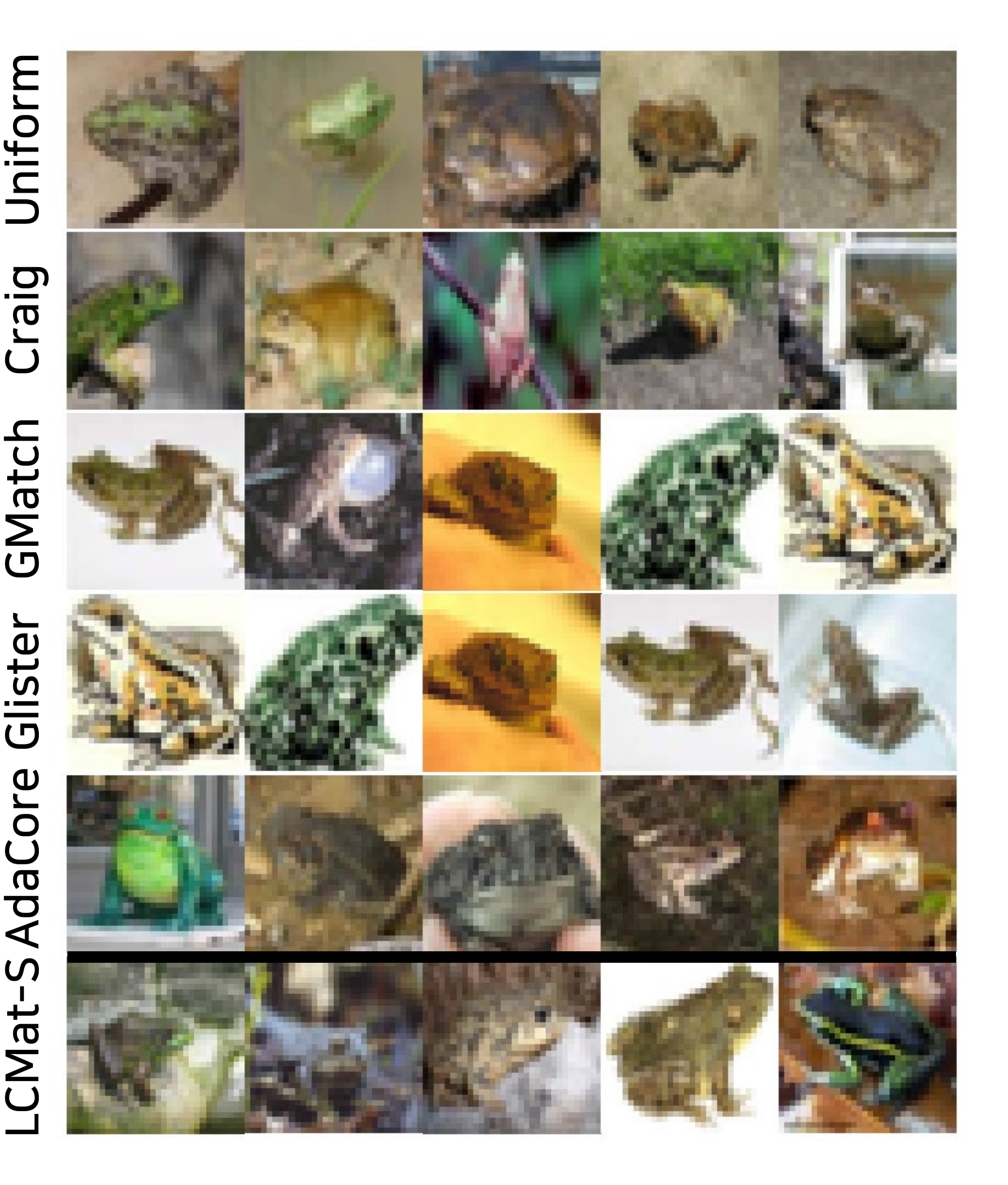}
    \caption{Frog}
\end{subfigure}\hfill
\begin{subfigure}[b]{0.195\columnwidth}
    \centering
    \includegraphics[width=\columnwidth]{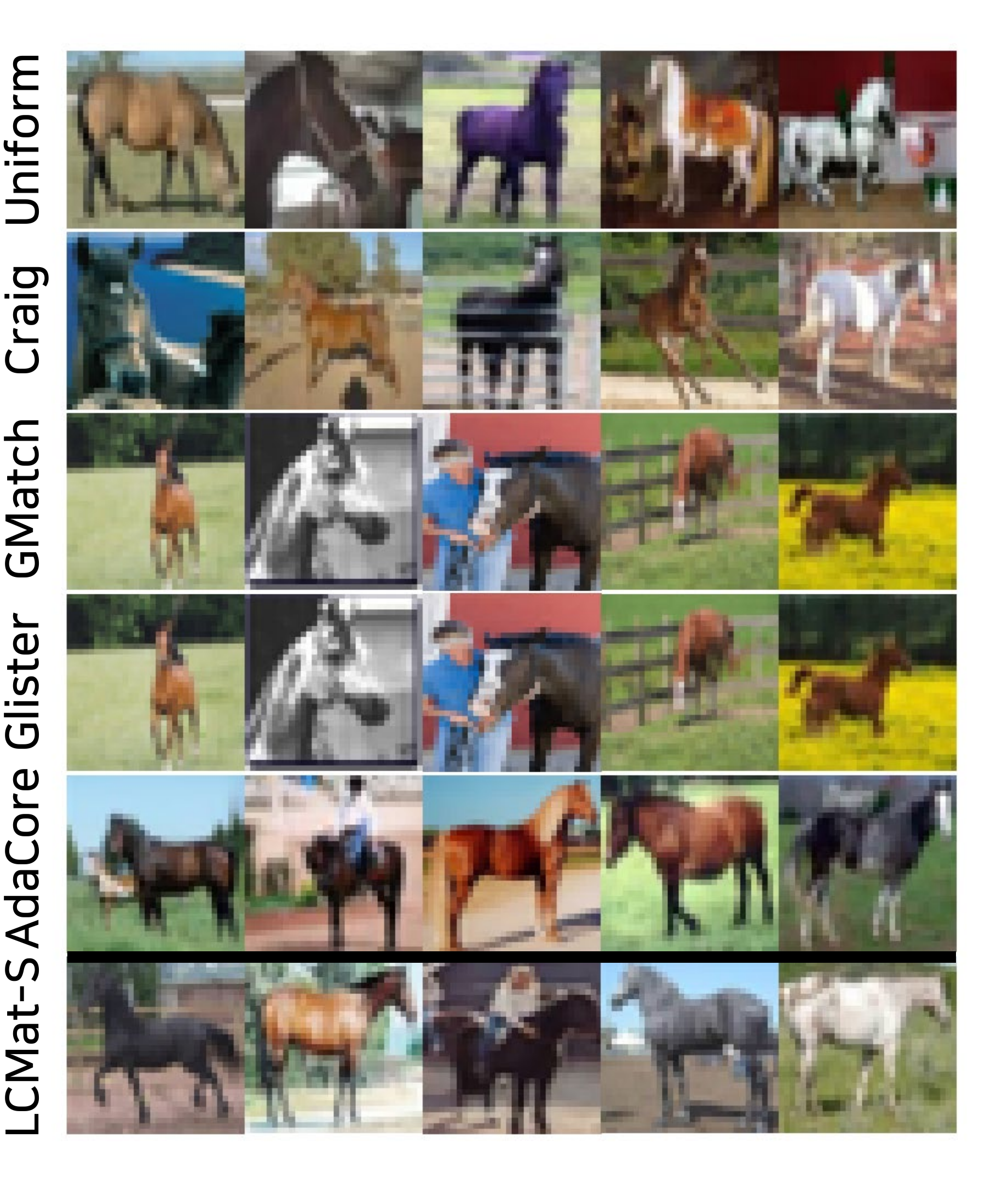}
    \caption{Horse}
\end{subfigure}\hfill
\begin{subfigure}[b]{0.195\columnwidth}
    \centering
    \includegraphics[width=\columnwidth]{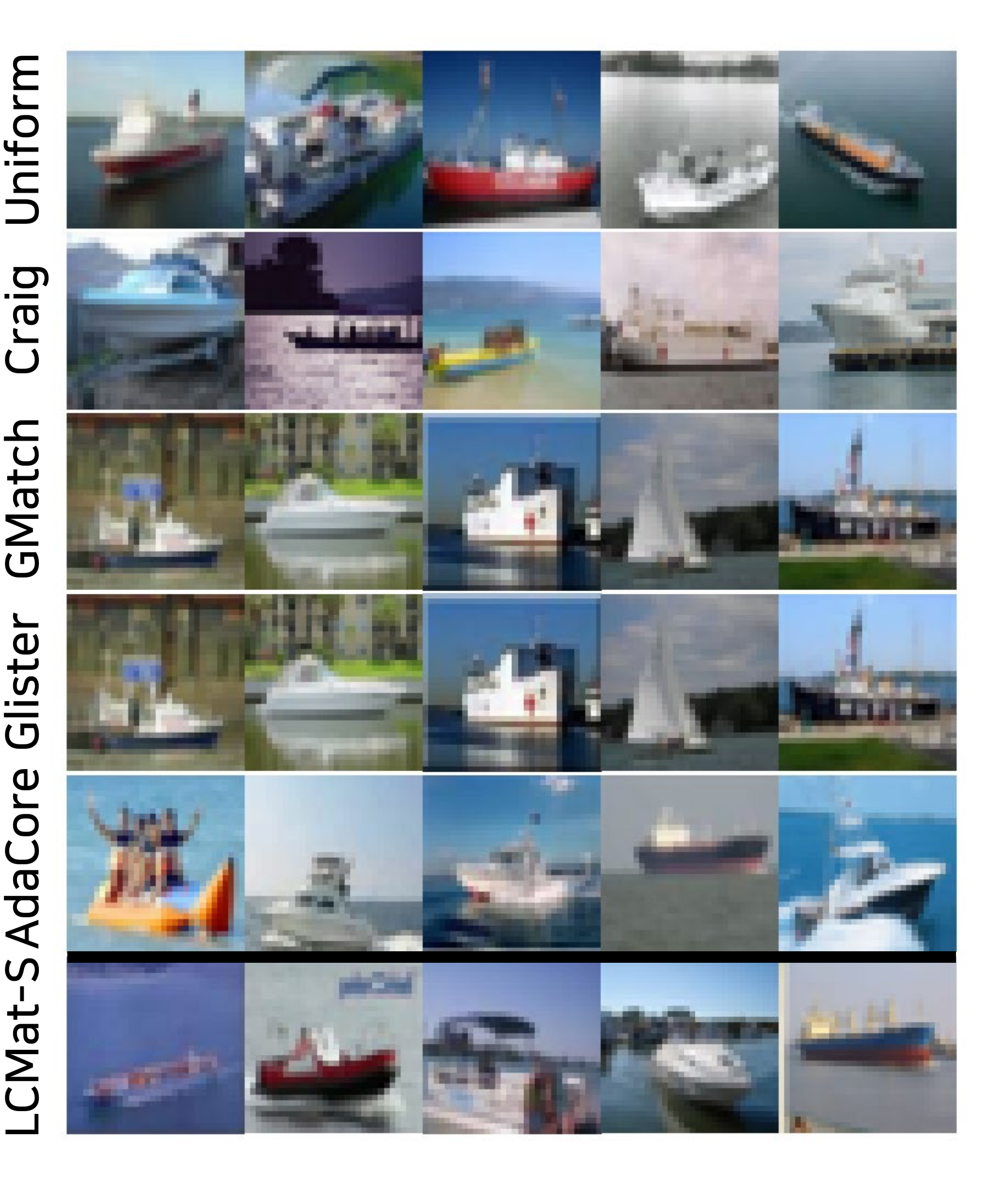}
    \caption{Ship}
\end{subfigure}\hfill
\begin{subfigure}[b]{0.195\columnwidth}
    \centering
    \includegraphics[width=\columnwidth]{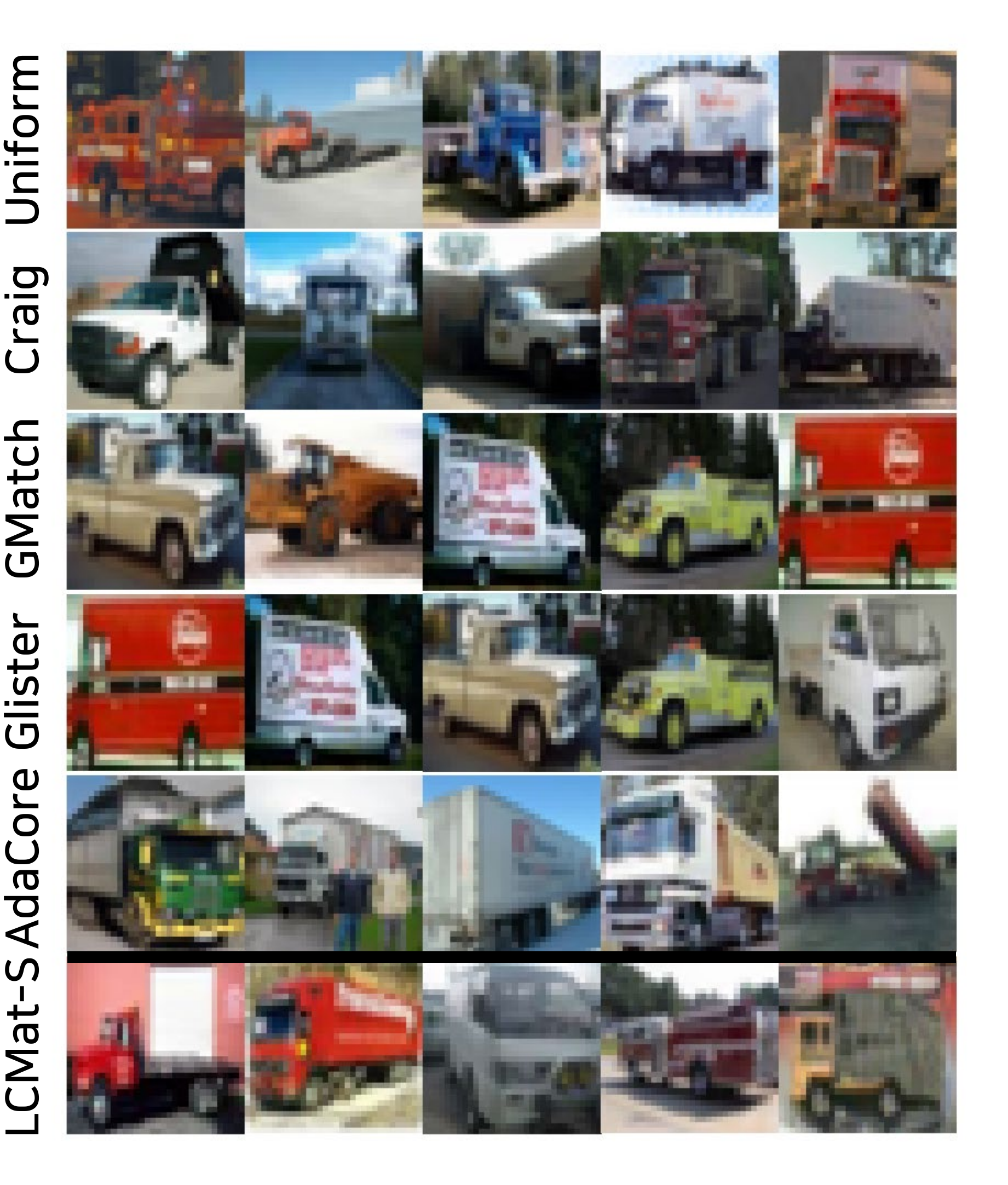}
    \caption{Truck}
\end{subfigure}\hfill
\caption{A set of images selected from each method. All samples are selected in a class-balanced way. We report selected images for 0.1$\%$ fraction here (Total of 50 images, 5 images per class).}
\label{fig:coreset-samples}
\end{figure}

Figure \ref{fig:coreset-samples} is the visualization of selected samples for CIFAR-10 dataset under ResNet-18 network structure. Whole selected images were displayed without any cherry-picking. LCMat-S selects a set of examples with diverse characteristics, e.g. the diverse shape of each object, different backgrounds without redundancy. 

\vfill

\end{document}